\documentclass{article} 
\usepackage{iclr2027_conference,times}

\usepackage{amsmath,amsfonts,bm}

\def\eqref#1{equation~\ref{#1}}

\def\1{\bm{1}}

\DeclareMathAlphabet{\mathsfit}{\encodingdefault}{\sfdefault}{m}{sl}
\SetMathAlphabet{\mathsfit}{bold}{\encodingdefault}{\sfdefault}{bx}{n}

\usepackage{hyperref}
\usepackage{url}

\usepackage[utf8]{inputenc} 
\usepackage[T1]{fontenc}    
\usepackage{booktabs}       
\usepackage{amsfonts}       
\usepackage{nicefrac}       
\usepackage{microtype}      
\usepackage{xcolor}         
\usepackage{amsmath}
\usepackage{graphicx}
\usepackage{comment}
\usepackage{float}
\usepackage[most]{tcolorbox}
\usepackage{array}
\newcolumntype{C}[1]{>{\centering\arraybackslash}p{#1}} 

\title{LLM-Based Evaluation Planning for Heterogeneous Optimization}

\author{\begin{minipage}{\textwidth}\raggedright\bfseries
Nian Ran$^{1,3}$ \quad  Sebastian Brulin$^{2}$ \quad Markus Olhofer$^{2}$ \quad
Mauricio A.~\'Alvarez$^{1}$ \quad Richard Allmendinger$^{1,\dagger}$ 
\end{minipage}  \\[6pt]
\normalfont\small $^{1}$The University of Manchester 
\normalfont\small $^{2}$Honda Research Institute Europe \\
\normalfont\small $^{3}$UKRI AI Centre for Doctoral Training in Decision Making for Complex Systems \\
\normalfont\small $^{\dagger}$Corresponding author: richard.allmendinger@manchester.ac.uk%
}

\iclrfinalcopy 
\begin{document}

\maketitle
\ificlrfinal\fancyhead[L]{LLM-Based Evaluation Planning for Heterogeneous Optimization}\fi
\begin{abstract}
Heterogeneous optimization problems require an optimizer to decide not only which candidate solutions to evaluate, but also which objectives and constraints should be evaluated under different costs and fidelities. Existing SOTA methods such as MFE-NSGA-III address this setting with manually designed selection metrics, but their behavior is largely determined by predefined heuristic rules and hyperparameters. In this work, we propose REP-HO, a reasoning-based evaluation planner that replaces the rule-based mixed-fidelity selection step with a large language model (LLM) planner. The LLM receives the same primitive optimization-state information as the heuristic planner, such as surrogate predictions, uncertainties, evaluation costs, and budget status. We first show that REP-HO achieves competitive performance compared to the human-designed metric-based method (MFE-NSGA-III). We then further enhance each decision with an adaptive nearest-neighbor summary, leading to better and more stable results. We evaluate REP-HO across five heterogeneous regimes covering cost imbalance, constraint difficulty, surrogate fidelity, budget level, and objective conflict. Across these regimes, one of the two variants obtains the best mean ranking score in every regime, REP-HO(Deepseek) is first or second throughout, and the planner shows adaptive allocation behavior, such as prioritizing constraint evaluations under tighter budgets and reallocating constraint evaluation budget when surrogate reliability changes. These results suggest that LLMs can serve as flexible evaluation planners for heterogeneous optimization, offering a general alternative to fixed hand-designed metrics.
\end{abstract}

\section{Introduction}
Many real-world optimization problems are not homogeneous. 
In scientific discovery, engineering design, simulation-based decision making, and machine learning system design, different quantities of interest may require substantially different evaluation procedures, runtimes, monetary costs, or reliability levels \citep{allmendinger2023heterogeneous}. 
For example, in molecular and materials design, cheap predictors can rapidly estimate drug-likeness or synthetic accessibility, whereas more reliable binding or toxicity assessments may require expensive physics-based simulations or wet-lab experiments \citep{eckmann2024mf-lal}. 
Similarly, in modern machine learning workflows, small-scale experiments, partial training runs, and low-fidelity approximations can provide useful but imperfect information before costly full-scale evaluations are conducted \citep{yen2025mfms-bo}. Such heterogeneity is not merely a practical detail, but a central factor in budget-limited optimization, and can change which evaluations are most valuable under a fixed budget \citep{allmendinger2023heterogeneous,cosson2024mnklandscape,cosson2022nmklandscape}. 
Classical multi-objective optimization methods typically assume that all objectives and constraints are evaluated together for each candidate solution. 
However, when evaluation costs or fidelities differ significantly, full evaluation may waste budget on quantities that are cheap but over-sampled, expensive but uninformative, or unnecessary for candidates that are already clearly infeasible or dominated. Recent work in evolutionary optimization and Bayesian optimization has therefore shown that explicitly exploiting heterogeneous evaluation structures can substantially improve optimization efficiency, especially when expensive objectives or constraints dominate the total computational cost~\citep{wang2021trsaea,xue2021dva-esto,wang2023sbp-bo,pang2023hhcbo,allmendinger2023heterogeneous,santoshkumar2025het-nsga-iii,santoshkumar2026-mfe-nsga-iii}.

Existing methods address heterogeneous evaluation through specialised modelling, acquisition, transfer, or selection mechanisms, including fidelity-aware kernels, correlation-aware surrogate models, cost-aware acquisition functions, heterogeneous selection operators, and cross-objective or cross-fidelity transfer~\citep{xing2025camo,gu2024lce-saea,wang2023sbp-bo,loka2023bi-chvei,santoshkumar2025het-nsga-iii,santoshkumar2026-mfe-nsga-iii,wang2021trsaea,tan2023T-sagp}. 
These methods provide valuable algorithmic insights, but they are usually tailored to a particular form of heterogeneity, such as asynchronous evaluation time, unequal evaluation cost, objective-wise landscape variation, or fidelity discrepancy. 
Some are further restricted to bi-objective, single-objective, or unconstrained settings, making them difficult to extend directly to general multi-objective constrained problems with multiple interacting sources of heterogeneity. A particularly relevant recent method is MFE-NSGA-III~\citep{santoshkumar2026-mfe-nsga-iii}, which integrates heterogeneous fidelity and evaluation cost into a unified mixed-fidelity selection metric for multi-objective constrained optimization. 
However, its evaluation decisions are still governed by manually predefined metrics, selection rules, and hyperparameter choices. Although effective, their behaviour is largely determined by human-designed criteria, which is complex and time-consuming, and often limit their adaptability to other problems and heterogeneities.

Multi-fidelity Bayesian optimization provides another important perspective by treating fidelity as an additional decision dimension and balancing cheap low-fidelity queries against expensive high-fidelity evaluations~\citep{do2025multifidelity-survey,xing2025camo,yen2025mfms-bo,eckmann2024mf-lal,foumani2023mfca-bo}. However, current multi-fidelity methods are mostly developed for single-objective or unconstrained settings, or for task-specific scenarios such as engineering simulation, data-mixture optimization for LLMs, molecular evaluation, and low-cost tuning of LLM judge designs~\citep{schuetz2024resum,yen2025mfms-bo,eckmann2024mf-lal,salinas2025tuning}. This specialization limits their direct transfer to general heterogeneous multi-objective constrained optimization, where objectives, constraints, fidelities, and costs may all interact under a shared budget. 

To address these limitations, this work investigates whether LLMs can serve as dynamic evaluation planners for heterogeneous optimization without additional training, hyperparameter tuning or prompt optimization. At each generation, the LLM receives the same optimization-state information available to the optimizer, without privileged information or problem-specific heuristics. Further, each prompt also lists a few nearby high-fidelity infill points chosen by an adaptive nearest-neighbor rule.
Using a fixed prompt template and only in-context reasoning, it plans the evaluation allocation across candidates, objectives, constraints, fidelities, and costs. We compare the proposed LLM-based planner against MFE-NSGA-III, a strong predefined-metric-based state-of-the-art baseline for multi-objective constrained optimization with heterogeneous fidelity and evaluation cost. 
To systematically examine its behaviour and adaptability, we design five heterogeneous regimes that emphasize fidelity, imbalanced cost, objective difficulty, and constraint difficulty respectively. 
Across these regimes, our results show that the LLM planner adapts its evaluation strategy to different heterogeneous structures and obtains competitive results and often better results with small token costs,(below \$0.11 per run (Appendix~\ref{app:llm-cost}), suggesting that LLMs can provide a flexible planner for heterogeneous optimization, offering a promising alternative to manually predefined metrics and problem-specific heuristic designs.
\section{Related Work}
\textbf{Heterogeneous Optimization}
The optimization community has studied heterogeneity mainly through objective-wise evaluation time, evaluation cost, constraint cost, and heterogeneous objective landscapes. 
For objectives with different evaluation times, Tr-SAEA~\citep{wang2021trsaea}, SBP-BO~\citep{wang2023sbp-bo}, and LCE-SAEA~\citep{gu2024lce-saea} use transfer learning, search-bias penalties, or local correlation estimation to exploit the information from cheaper or faster objectives. These methods improve the use of partially evaluated information, but they are usually designed for unconstrained or bi-objective problems and rely on specific assumptions about the relation between objectives. Other works focus on cheap and expensive objectives or constraints. CHVEI/CHVPOI~\citep{loka2023bi-chvei} and HHCBO~\citep{pang2023hhcbo} modify hypervolume-based acquisition functions for bi-objective problems where one objective is cheap and the other is expensive. IOC-SAMO-COBRA~\citep{de2024ioc-samo-cobra} trains surrogates only for expensive functions and directly evaluates inexpensive ones, while SParEAcost~\citep{rahi2024spareacost} orders expensive constraints according to predicted violation and evaluation cost. These methods are effective when the cheap--expensive structure is predefined, but they often assume a fixed cost pattern, limited objective number, or specific objective--constraint roles. Heterogeneous objective landscapes have also been studied through MNK landscapes~\citep{cosson2022nmklandscape,cosson2024mnklandscape}, where they provide a controlled benchmark and analytical lens to show that objective heterogeneity fundamentally alters the local search geometry of MOO even when global pareto fronts remain similar. ExLLM \citep{ran2025exllm} and MCCE \citep{ran2025mcce} further proves that LLM can also be used for heterogeneous problems in optimization. The most relevant line of work is HET-NSGA-III and MFE-NSGA-III by ~\citet{santoshkumar2025het-nsga-iii,santoshkumar2026-mfe-nsga-iii}.  They use manually designed selection metrics to decide which candidate--objective or candidate--constraint pairs should receive expensive evaluations. 
MFE-NSGA-III further integrates heterogeneous fidelity and evaluation cost into a unified mixed-fidelity selection metric, making it a strong baseline for heterogeneous multi-objective constrained optimization. 

\textbf{Multi-Fidelity Optimization in ML}
In the machine learning community, heterogeneity is more commonly studied as multi-fidelity optimization. Multi-fidelity Bayesian optimization treats fidelity as an additional decision dimension and balances cheap low-fidelity evaluations against expensive high-fidelity evaluations. 
Recent methods improve this process through cost-aware acquisition functions, fidelity-aware kernels, latent fidelity representations, or neural surrogate models~\citep{foumani2023mfca-bo,niu2024mfrnp,zhang2024rmfbo-dp,xing2025camo}. For example, CAMO~\citep{xing2025camo} models fidelity-dependent convergence with a non-stationary GP kernel, while MFCA-BO~\citep{foumani2023mfca-bo} uses latent-map Gaussian processes and cost-normalized acquisition. These methods provide strong modelling tools for fidelity--cost trade-offs, but they are usually designed for single-objective unconstrained optimization and often require assumptions about cross-fidelity correlation.
Multi-fidelity optimization has also been applied to specific ML and scientific discovery tasks. 
MFMS-BO~\citep{yen2025mfms-bo} optimizes data mixtures by treating model scale and training steps as fidelity dimensions. MF-LAL~\citep{eckmann2024mf-lal} uses multiple molecular evaluation oracles with different costs and accuracies. LLM judge tuning methods~\citep{salinas2025tuning} define fidelity through the number of evaluation examples or comparisons. These applications show the practical value of multi-fidelity optimization, but their fidelity structures are usually task-specific and manually constructed, making direct transfer to general heterogeneous multi-objective constrained optimization difficult. \textit{We also review related work on Bayesian optimization with LLMs and LLM-based planning, as these directions are less directly related to our problem setting and space limitations, we put it in Appendix~\ref{app:additional-related-work}.}

\section{Methodology}
\label{sec:method}
\paragraph{Problem Definition} We consider a constrained multi-objective optimization problem with heterogeneous
evaluation fidelity and heterogeneous evaluation cost. Let
\(\mathbf{x}\in\Omega\subset\mathbb{R}^{D}\) denote a candidate solution. The
problem is formulated as
\begin{equation}
    \begin{aligned}
        \min_{\mathbf{x}\in\Omega}\quad
        \mathbf{f}(\mathbf{x}) &=
        \left(f_1(\mathbf{x}), f_2(\mathbf{x}), \ldots, f_M(\mathbf{x})\right),
        \text{s.t.}\quad
        g_j(\mathbf{x}) &\leq 0,\qquad j=1,2,\ldots,J .
    \end{aligned}
    \label{eq:constrained-many-objective-problem}
\end{equation}
Here \(M\) is the number of objectives and \(J\) is the number of constraints.
The goal is to approximate the feasible Pareto front under a fixed evaluation
budget \(T_{\max}\). The budget is measured in normalized time units: one full
high-fidelity evaluation of all objectives and all constraints costs one time
unit. Thus, if \(ET^m_f\) is the high-fidelity cost of objective \(f_m\), and
\(ET^j_g\) is the high-fidelity cost of constraint \(g_j\), then$\sum_{m=1}^{M} ET^m_f + \sum_{j=1}^{J} ET^j_g = 1$. First, each objective and
constraint can be evaluated at different fidelities. A high-fidelity evaluation
uses the true evaluation function and returns the actual value of
\(f_m(\mathbf{x})\) or \(g_j(\mathbf{x})\), whereas a low-fidelity evaluation often
uses a surrogate model prediction. For a solution \(s\), the surrogate provides
the predicted objective value, predicted constraint value, and their associated
uncertainties. Second, the high-fidelity costs are imbalanced across targets:
some objectives or constraints may be much more expensive than others. The
algorithm must therefore decide not only which candidate solutions to evaluate,
but also which subset of objectives and constraints should receive
high-fidelity evaluation under the remaining budget.

\paragraph{Performance Metrics} We evaluate the quality of the final nondominated feasible approximation set \(A\) using IGD+ and hypervolume (HV). Let \(P^\star\) denote a reference set
sampled from the true Pareto front. The IGD+ metric is
\begin{equation}
    \mathrm{IGD}^{+}(A, P^\star)
    =
    \frac{1}{|P^\star|}
    \sum_{\mathbf{z}\in P^\star}
    \min_{\mathbf{a}\in A}
    d^{+}(\mathbf{z}, \mathbf{a}), \hspace{1cm}
    d^{+}(\mathbf{z}, \mathbf{a})
    =
    \sqrt{
    \sum_{m=1}^{M}
    \left(\max\{a_m-z_m,0\}\right)^2
    } .
    \label{eq:igd-plus}
\end{equation}
A smaller IGD+ value indicates that the obtained approximation set is closer to
and better covers the reference Pareto front. The HV metric measures the objective-space volume dominated by \(A\) and bounded
by a user-defined reference point \(\mathbf{r}\):
\begin{equation}
    \mathrm{HV}(A)
    =
    \lambda\left(
    \bigcup_{\mathbf{a}\in A}
    [a_1,r_1]\times [a_2,r_2]\times \cdots \times [a_M,r_M]
    \right),
    \label{eq:hypervolume}
\end{equation}
where \(\lambda(\cdot)\) denotes the Lebesgue measure. A larger HV value
indicates better convergence and diversity, assuming the same reference point is
used across methods.

\subsection{Baselines and Preliminaries}
\label{sec:baselines}
\textbf{NSGA-III} is a reference-vector-guided evolutionary multi-objective optimization
algorithm. It uses nondominated sorting and a reference direction based niching
mechanism to maintain a diverse approximation set in high-dimensional objective
spaces. In our setting, NSGA-III serves as the high-fidelity baseline: every
selected solution is evaluated by the true evaluation functions for all
objectives and all constraints.

\textbf{SA-NSGA-III} is a surrogate-assisted version of NSGA-III. It uses surrogate
models to search cheaply between high-fidelity evaluations, but the final infill
solutions are still evaluated in a uniform manner. That is, when a candidate is
selected for high-fidelity evaluation, all \(M\) objectives and all \(J\)
constraints are evaluated together, without exploiting target-wise cost
imbalance.

\textbf{MFE-NSGA-III} \citep{santoshkumar2026-mfe-nsga-iii} is based on NSGA-III for
heterogeneous constrained multi-objective optimization with an important metric: mixed-fidelity selection (MFS) which incorporates the likelihood of a solution dominating its neighbors, computational cost, surrogate model error, and the extent of constraint violation. MFE-NSGA-III first identifies near-boundary candidates by \(Z\)-scores of constraints, then computes their MFS scores to determine which objectives or constraints of which candidates should be evaluated at high fidelity. The remaining targets are then evaluated by surrogate model (low fidelity). For each solution \(s\) and constraint \(j\), the surrogate prediction and
uncertainty are denoted by \(\mu_s^j\) and \(\sigma_s^j\), respectively. The
constraint \(Z\)-score and overall probability of satisfying all constraints are defined as
\begin{equation}
    Z_s^j = \frac{-\mu_s^j}{\sigma_s^j}, 
    \hspace{0.5cm}
    PG_s
    =
    \prod_{j=1}^{J} P(g_j(s)\leq 0)
    =
    \prod_{j=1}^{J} \Phi(Z_s^j),
    \label{eq:constraint-zscore}
\end{equation}
where \(\Phi(\cdot)\) is the standard normal cumulative distribution function. Because constraints are written as \(g_j(\mathbf{x})\leq 0\), larger
\(Z_s^j\) values indicate higher predicted feasibility. Using a threshold
\(e<0\), each population member is classified by its worst constraint score:
solutions with \(\min_j Z_s^j \geq e\) are treated as feasible or near-boundary
members, while the remaining solutions are treated as largely infeasible.

 For every feasible or near-boundary solution \(s\), MFE-NSGA-III computes an
MFS metric for each objective \(m\):
\begin{equation}
    \rho_s^m
    =
    \left(1+\widehat{ET}^{m}\right)^{\alpha}
    \left(
    1+
    \left(
    \frac{\sigma_s^m}{|\Delta f_m|}
    \right)^{\frac{1}{\eta}}
    \right).
    \label{eq:mfe-selection-metric}
\end{equation}
Here \(P_s^m\), defined by \citet{santoshkumar2025het-nsga-iii}, is the probability that solution \(s\) is nondominated among its
neighboring members with respect to the reference vectors (from NSGA-III), \(\sigma_s^m\) is the
surrogate uncertainty for objective \(m\), and \(\Delta f_m\) is the range of
the \(m\)th objective in the current population. The normalized evaluation time
\(\widehat{ET}^{m}\) represents the relative high-fidelity cost of objective
\(m\). The hyperparameter \(\eta\) controls the strength of the uncertainty term, and
\(\alpha\) changes over the run to balance cheap and expensive target
evaluations as the budget is consumed.

The selection procedure uses these quantities as follows. If enough feasible or
near-boundary members are available, MFE-NSGA-III selects candidates across
reference-vector neighborhoods using \(\rho_s^m\), evaluates the objective with
the largest selected \(\rho_s^m\), and evaluates only those constraints whose
\(Z_s^j\) values lie near the boundary, \(e\leq Z_s^j\leq -e\). If feasible
members are insufficient, the remaining slots are filled by largely infeasible
solutions ranked by \(PG_s\), usually without objective evaluation. If all
members are largely infeasible, the method evaluates the least violated
constraint, i.e., the constraint with the largest \(Z_s^j\), to guide the search
toward feasibility. In this way, MFE-NSGA-III spends high-fidelity budget. The hyperparameter \(\eta\) and \(e\) are set to be 20 and -0.5 respectively, following the best settings reported in MFE-NSGA-III.

\subsection{REP-HO}
\label{sec:rep-hbo}
\begin{figure}[htbp]
    \centering
    \includegraphics[width=0.85\linewidth]{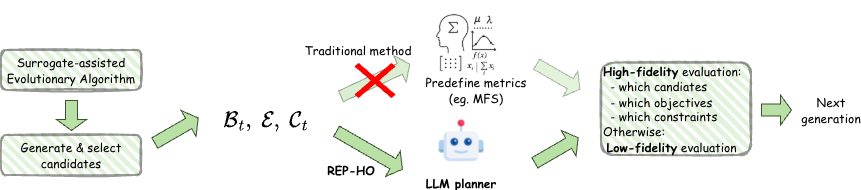}
    \caption{The workflow of REP-HO.}
    \label{fig:repho}
\end{figure}
We propose Reasoning-based Evaluation Planning for Heterogeneous
Optimization (REP-HO) as shown in Figure \ref{fig:repho}. The key
difference is the evaluation-planning step. In MFE-NSGA-III, the decision of
which objectives and constraints receive high-fidelity evaluation is made by
fixed rules based on the constraint \(Z\)-score \(Z_s^j\), the constraint
satisfaction probability \(PG_s\), and the mixed-fidelity selection metric
\(\rho_s^m\). REP-HO keeps the same surrogate-assisted optimization backbone,
candidate generation process, reference-vector neighborhood information, and
budget accounting, but replaces this rule-based target allocation with an LLM
planner. At each generation, the optimizer first constructs an infill candidate set using
the same surrogate-assisted evolutionary search as MFE-NSGA-III. Let
\(\mathcal{S}_t\) be the set of candidate solutions selected for possible
evaluation at generation \(t\). For each \(s\in\mathcal{S}_t\), the
LLM chooses which objective and constraint targets should be evaluated with
high fidelity. 

\textbf{LLM State} For each LLM call, REP-HO builds a structured state
\(\mathcal{I}_t\) containing only information available to the optimizer at
generation \(t\). The state consists of the
following components: $
    \mathcal{I}_t =
    \left(
    \mathcal{B}_t,
    \mathcal{E},
    \mathcal{C}_t,
    \mathcal{R}
    \right)$ where \(\mathcal{B}_t\) is the current budget state, \(\mathcal{E}\) is the
target-cost table, \(\mathcal{C}_t\) is the displayed candidate batch, and
\(\mathcal{R}\) is the decision rule and output schema.
The same state and action template is fixed for all problems. Only the
brief problem-specific textual introduction and the dimensions \(M\) and \(J\) change
with the benchmark; the information fields, allowed actions, output format, and
batch size are kept identical across all experiments. The budget state is
\begin{equation}
    \mathcal{B}_t =
    \left(
    T_t,\,
    T_t-\tau,\,
    T_{\max},\,
    T_{\max}-T_t,\,
    \frac{T_t-\tau}{T_{\max}-\tau}
    \right),
    \mathcal{E}
    =
    \left\{
    (f_m, ET_f^m)_{m=1}^{M},
    (g_j, ET_g^j)_{j=1}^{J}
    \right\}.
    \label{eq:rep-target-costs}
\end{equation}
where \(T_t\) is the current consumed time, \(\tau\) is the initial design of
experiments cost, and \(T_{\max}\) is the total time budget. Thus the LLM is informed that objectives and constraints have unequal
high-fidelity costs, and that only requested high-fidelity targets consume
budget. Each candidate \(s\in\mathcal{C}_t\) is shown with a stable label and the
following state:
\begin{equation}
    \psi_t(s)
    =
    \left(
    \mathbf{x}_s,\,
    \{ \hat{f}_s^m, \sigma_s^m, P_s^m, h_s^m \}_{m=1}^{M},\,
    \{ \hat{g}_s^j, \sigma_s^j, Z_s^j, h_s^j \}_{j=1}^{J},\,
    PG_s
    \right).
    \label{eq:rep-candidate-state}
\end{equation}
Here \(\hat{f}_s^m\) and \(\hat{g}_s^j\) are surrogate predictions,
\(\sigma_s^m\) and \(\sigma_s^j\) are surrogate uncertainties, \(P_s^m\) is the
surrogate-based probability that \(s\) is nondominated in the corresponding
objective view, and \(PG_s\) is the overall constraint satisfaction
probability. The binary variables \(h_s^m\) and \(h_s^j\) indicate whether the
corresponding target has already been evaluated with high fidelity, preventing
repeated high-fidelity requests for the same candidate-target pair. The
constraint \(Z\)-score is the same statistic used by MFE-NSGA-III.
The prompt also states the feasibility convention \(g_j(s)\leq 0\) and the
objective direction.

For fairness, the LLM planner is given the same amount of information that is
available to the MFS calculation at the selection stage: surrogate predictions,
surrogate uncertainties, nondomination probabilities, constraint \(Z\)-scores,
overall constraint satisfaction probability, already-evaluated target flags,
evaluation costs, and budget status. However, REP-HO does not approximate the MFS
score, and the MFS score \(\rho_s^m\) is not included in the prompt.
The LLM must therefore form its own evaluation plan from the same primitive
state variables rather than observing the hand-designed MFE score. In addition,
high-fidelity constraint selection is also decided by the LLM, replacing the
original rule-based constraint selection based on boundary \(Z\)-scores. This
design enables a direct comparison between reasoning-based planning and the
predefined MFE heuristic under matched information, candidate sets, evaluation
costs, and budget accounting. Each LLM query processes a fixed batch of 10 candidate solutions for all
problems. If the population size is \(N=50\), REP-HO therefore sends five LLM
queries to complete the target-planning step for one generation. The LLM is
told the batch index, the number of total batches, the number of shown
candidates, and the number of candidates appearing in the other batches, so that
each batch is planned as part of the same generation-level budget. In our
experiments, the same planner is instantiated with the official hosted versions of
Qwen3.5-35B-A3B and DeepSeek-V4-Flash, reported as REP-HO(Qwen) and REP-HO(Deepseek).

\textbf{LLM Action Space} For each candidate \(s\in\mathcal{C}_t\), the LLM returns an action
\begin{equation}
    a_t(s) =
    \left(
    \mathcal{O}_s,\,
    \mathcal{G}_s
    \right),
    \qquad
    \mathcal{O}_s\subseteq\{f_1,\ldots,f_M\},\quad
    \mathcal{G}_s\subseteq\{g_1,\ldots,g_J\}.
    \label{eq:rep-action}
\end{equation}
The set \(\mathcal{O}_s\) contains the objectives requested for high-fidelity
evaluation, and \(\mathcal{G}_s\) contains the constraints requested for
high-fidelity evaluation. Empty sets are allowed, meaning that the candidate is
kept with surrogate predictions for all non-requested targets. The complete LLM
decision for a batch is \(A_t = \{a_t(s): s\in\mathcal{C}_t\}.\)
The action is returned as structured JSON with one decision for every candidate
in the batch. A concrete prompt and response example is provided in
Appendix~\ref{app:prompt-example}. After parsing the response, the optimizer
evaluates only the requested candidate-target pairs with high fidelity; all
non-requested objective and constraint values remain low-fidelity surrogate
values for that generation.

\subsection{Design of regimes}
We evaluate REP-HO under five controlled experimental regimes. Unless
otherwise stated, all experiments use budget \(T_{\max}=300\), uniform evaluation costs, a fixed
surrogate family. We will briefly introduce each regime in this section, the full computing details and settings are explained in Appendix \ref{app:5regime}.

\textbf{Regime 1: cost imbalance.}
This regime tests whether the planner adapts to imbalanced evaluation costs
across objectives and constraints. We use TNK, MW3, Welded Beam, and Carside.
For TNK and MW3, we compare uniform, objective-heavy, and constraint-heavy
settings. \textbf{Regime 2: constraint difficulty.}
This regime tests whether the planner adapts when the feasible region is large
or small. We use TNK and Welded Beam, and their constraint difficulty is controlled by a
shared shift \(g'_j(\mathbf{x})=g_j(\mathbf{x})-\delta\leq 0\), where
\(\delta\) is calibrated by pilot sampling: easy settings target a feasible
ratio of about 25\%, and hard settings target about 0.1\%. \textbf{Regime 3: fidelity correlation.} This regime tests how the planner changes its behaviors when the surrogate model is high or low fidelity. We also use TNK and Welded Beam. The
high-correlation setting uses reliable surrogate predictions, while the
low-correlation setting weakens the surrogate mild noise and structural bias. The settings are calibrated on an independent reference pool using objective-wise LF-HF Spearman correlation and
constraint-wise correlation/MAE. \textbf{Regime 4: budget.}
This regime tests whether REP-HO performs genuine budget-aware planning rather
than relying on a single fixed budget. We evaluate low, mid, and high budgets,
\(T_{\max}\in\{200,300,400\}\), using uniform costs and high-correlation
surrogates. The main problems are TNK, MW3, and Carside. \textbf{Regime 5: objective conflict.}
This regime tests whether the planner responds to weak versus strong objective
conflict. We use ZDT1 as a low-conflict unconstrained case and Kursawe as a
high-conflict unconstrained case. We also use BNH as a lower-conflict
constrained case and OSY as a higher-conflict constrained case. 

\subsection{Adaptive nearest-neighbor context}
\label{sec:adaptive-knn}
We further add a simple but general component to REP-HO, a lists a few previously evaluated infill points near the candidate being planned is provided in prompt. Initial design-of-experiment points are excluded: they are fully high-fidelity evaluated, and their coverage is already reflected by the surrogate's information. Distance is the root-mean-square distance in the bound-normalized design space. Let \(d_1\) be the distance to the nearest remaining historical infill. A neighbor is kept only if its distance is at most \(1.5d_1\), and at most three neighbors are shown. If fewer points fall inside this radius the list is shorter, and if no later infill exists the list is empty. Only high-fidelity values of those neighbors are shown, low fidelity values are shown as not evaluated. The same planner is instantiated with Qwen3.5-35B-A3B and DeepSeek-V4-Flash.

\section{Results}

\label{app:no-knn}
\paragraph{Baseline results}
We first evaluate REP-HO without Adaptive Nearest-Neighbor Context (ANNC) to isolate the performance of pure LLM-based planning under the same state information. As shown in Table~\ref{tab:no-knn-ranking}, REP-HO (Qwen) already achieves competitive performance, ranking first across all regimes except Regime 5. We then incorporate ANNC into REP-HO and further investigate how it adapts to different forms of heterogeneity.
\begin{table}[h]
    \centering
    \caption{Overall ranking scores without the adaptive nearest-neighbor context. Lower is better.}
    \scriptsize
    \label{tab:no-knn-ranking}
    \begin{tabular}{lcccc}
        \toprule
        Regime & NSGA-III & SA-NSGA-III & MFE-NSGA-III & REP-HO (Qwen)\\
        \midrule
        1 & 3.625 & 3.208 & 1.667 & \textbf{1.500} \\
        2 & 2.875 & 2.750 & 1.875 & \textbf{1.375} \\
        3 & -- & 2.875 & 1.875 & \textbf{1.250} \\
        4 & 3.333 & 3.333 & 2.083 & \textbf{1.250} \\
        5 & 3.292 & 3.229 & \textbf{1.708} & 1.771 \\
        \bottomrule
    \end{tabular}
\end{table}
\begin{table*}[t]
    \centering
    \caption{Regime 1: Cost Imbalance HV performance summary (HV $\uparrow$).}
    \label{tab:regime1-performance}
    \scriptsize
    \setlength{\tabcolsep}{3pt}
    \resizebox{\textwidth}{!}{%
    \begin{tabular}{p{0.11\textwidth} p{0.12\textwidth} p{0.12\textwidth} p{0.12\textwidth} p{0.12\textwidth} p{0.13\textwidth} p{0.15\textwidth}}
        \toprule
        Problem & Setting & NSGA-III & SA-NSGA-III & MFE-NSGA-III & REP-HO(Qwen) & REP-HO(Deepseek) \\
        \midrule
        Carside & uniform & 6.3528 $\pm$ 0.2935 & 8.4753 $\pm$ 0.0745 & \textbf{8.6621 $\pm$ 0.0500} & 8.5517 $\pm$ 0.0679 & 8.5779 $\pm$ 0.0715 \\
        Carside & constraint-heavy & 6.3528 $\pm$ 0.2935 & 8.4753 $\pm$ 0.0745 & \textbf{8.6818 $\pm$ 0.0306} & 8.6276 $\pm$ 0.0467 & 8.6519 $\pm$ 0.0345 \\
        Carside & objective-heavy & 6.3528 $\pm$ 0.2935 & 8.5264 $\pm$ 0.0542 & \textbf{8.6822 $\pm$ 0.0321} & 8.6344 $\pm$ 0.0501 & 8.6235 $\pm$ 0.0641 \\
        MW3 & constraint-heavy & 1.7827 $\pm$ 0.6760 & 2.9285 $\pm$ 0.2532 & \textbf{3.2230 $\pm$ 0.2938} & 3.1430 $\pm$ 0.2630 & 3.2076 $\pm$ 0.1584 \\
        MW3 & uniform & 1.7827 $\pm$ 0.6760 & 2.9285 $\pm$ 0.2532 & 3.1713 $\pm$ 0.2775 & 3.1837 $\pm$ 0.2781 & \textbf{3.2219 $\pm$ 0.1796} \\
        MW3 & objective-heavy & 1.7827 $\pm$ 0.6760 & 2.9285 $\pm$ 0.2532 & 3.1810 $\pm$ 0.2877 & 3.1855 $\pm$ 0.1860 & \textbf{3.2880 $\pm$ 0.1751} \\
        TNK & constraint-heavy & 2.9360 $\pm$ 0.1403 & 3.1621 $\pm$ 0.1836 & 3.2193 $\pm$ 0.0909 & 3.2667 $\pm$ 0.0524 & \textbf{3.2865 $\pm$ 0.0062} \\
        TNK & uniform & 2.9360 $\pm$ 0.1403 & 3.1621 $\pm$ 0.1836 & 3.2086 $\pm$ 0.0767 & \textbf{3.2865 $\pm$ 0.0090} & 3.2716 $\pm$ 0.0521 \\
        TNK & objective-heavy & 2.9360 $\pm$ 0.1403 & 3.1621 $\pm$ 0.1836 & 3.1950 $\pm$ 0.0891 & \textbf{3.2683 $\pm$ 0.0415} & 3.2452 $\pm$ 0.0734 \\
        WeldedBeam & constraint-heavy & 3.5988 $\pm$ 0.1162 & 3.1011 $\pm$ 0.8082 & \textbf{3.7437 $\pm$ 0.1667} & 3.6128 $\pm$ 0.3698 & 3.4661 $\pm$ 0.8108 \\
        WeldedBeam & uniform & 3.5988 $\pm$ 0.1162 & 2.3793 $\pm$ 1.2787 & 3.5144 $\pm$ 0.8241 & \textbf{3.6635 $\pm$ 0.1971} & 3.4924 $\pm$ 0.8139 \\
        WeldedBeam & objective-heavy & 3.5988 $\pm$ 0.1162 & 2.3793 $\pm$ 1.2787 & 3.5809 $\pm$ 0.4645 & 3.3885 $\pm$ 0.9055 & \textbf{3.6654 $\pm$ 0.2032} \\
        \midrule
        \multicolumn{2}{l}{Mean ranking integrated with IGD+ in Table \ref{tab:regime1-performance-igd}}  & 4.562 & 4.167 & 2.188 & 2.125 & \textbf{1.958} \\
        \bottomrule
    \end{tabular}%
    }
\end{table*}
\paragraph{Regime 1} For each setting, we run each method with 20 random seeds and report the mean and standard deviation. For each task, the best-performing method receives a rank of 1, the second-best receives a rank of 2, and so on. The final ranking score is computed as the average rank across all experiments, where a lower score indicates better overall performance.
Table~\ref{tab:regime1-performance} reports the cost-imbalance results with two planner backbones. REP-HO(Deepseek) obtains the best mean ranking (1.958), followed by REP-HO(Qwen) (2.125) and MFE-NSGA-III (2.188); SA-NSGA-III (4.167) and NSGA-III (4.562) remain far behind. The two planners split the wins: REP-HO(Deepseek) is strongest on MW3 and on constraint-heavy TNK, while REP-HO(Qwen) leads IGD+ on the skewed Carside settings and on uniform and objective-heavy TNK. MFE-NSGA-III still has the best HV on all three Carside settings and the best HV and IGD+ on constraint-heavy Welded Beam. Both REP-HO variants and MFE-NSGA-III clearly outperform NSGA-III and SA-NSGA-III, so the gain comes from target-wise heterogeneous planning.
\paragraph{Regime 2}
\begin{table*}[h]
    \centering
    \caption{Regime 2: Constraint Difficulty HV performance summary (HV $\uparrow$).}
    \label{tab:regime2-performance}
    \scriptsize
    \setlength{\tabcolsep}{3pt}
    \resizebox{\textwidth}{!}{%
    \begin{tabular}{p{0.11\textwidth} p{0.15\textwidth} p{0.12\textwidth} p{0.12\textwidth} p{0.12\textwidth} p{0.13\textwidth} p{0.15\textwidth}}
        \toprule
        Problem & Constraint Difficulty & NSGA-III & SA-NSGA-III & MFE-NSGA-III & REP-HO(Qwen) & REP-HO(Deepseek) \\
        \midrule
        TNK & hard & \textbf{2.0092 $\pm$ 0.0974} & 1.8559 $\pm$ 0.1354 & 1.9155 $\pm$ 0.0842 & 1.9253 $\pm$ 0.0319 & 1.9046 $\pm$ 0.1048 \\
        TNK & easy & 0.1015 $\pm$ 0.4425 & 0.0000 $\pm$ 0.0000 & 1.7087 $\pm$ 0.5044 & \textbf{4.1710 $\pm$ 0.0000} & 1.8823 $\pm$ 0.5521 \\
        WeldedBeam & hard & 2.3750 $\pm$ 0.1480 & 2.5098 $\pm$ 0.1492 & 2.3072 $\pm$ 0.7913 & \textbf{2.5205 $\pm$ 0.2100} & 2.4801 $\pm$ 0.5828 \\
        WeldedBeam & easy & 3.5777 $\pm$ 0.1346 & 2.8006 $\pm$ 1.2375 & 3.1992 $\pm$ 1.1231 & \textbf{3.5936 $\pm$ 0.4936} & 3.4911 $\pm$ 0.8190 \\
        \midrule
        \multicolumn{2}{l}{Mean ranking integrated with IGD+ in Table \ref{tab:regime2-performance-igd}} & 3.312 & 3.938 & 2.375 & \textbf{2.000} & 2.062 \\
        \bottomrule
    \end{tabular}%
    }
\end{table*}

\begin{figure}[h]
    \centering
    \begin{minipage}{0.495\textwidth}
        \centering
        \includegraphics[width=\linewidth]{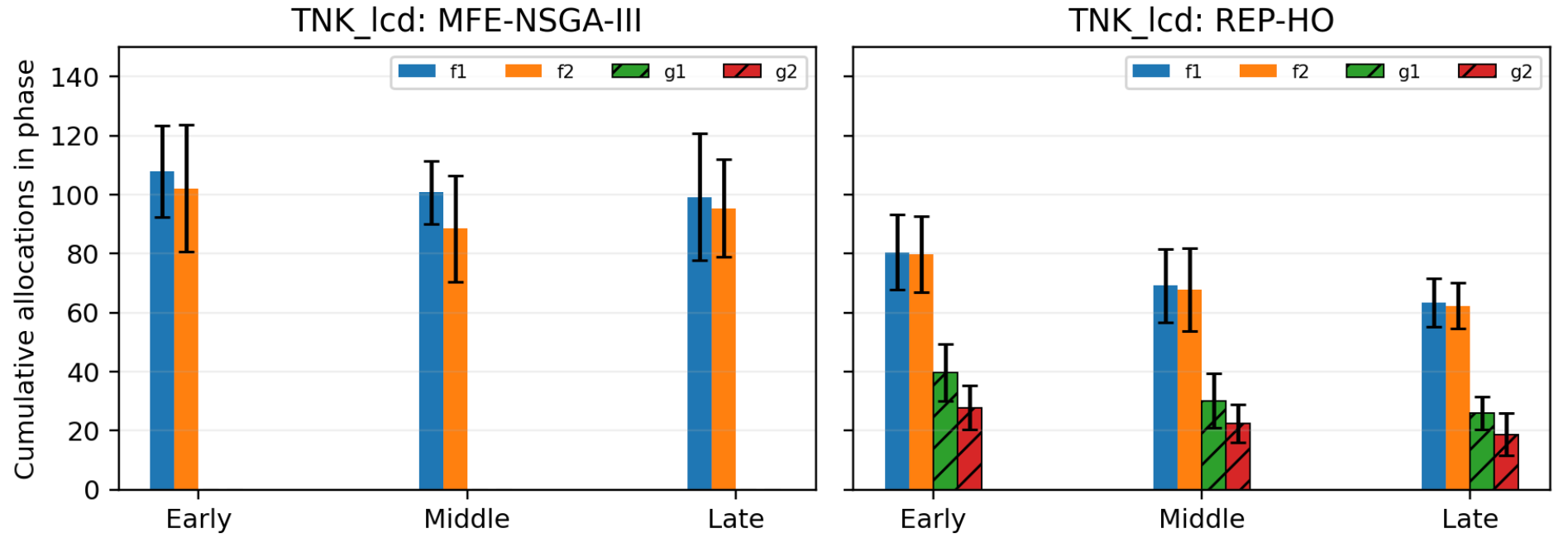}
    \end{minipage}
    \hfill
    \begin{minipage}{0.495\textwidth}
        \centering
        \includegraphics[width=\linewidth]{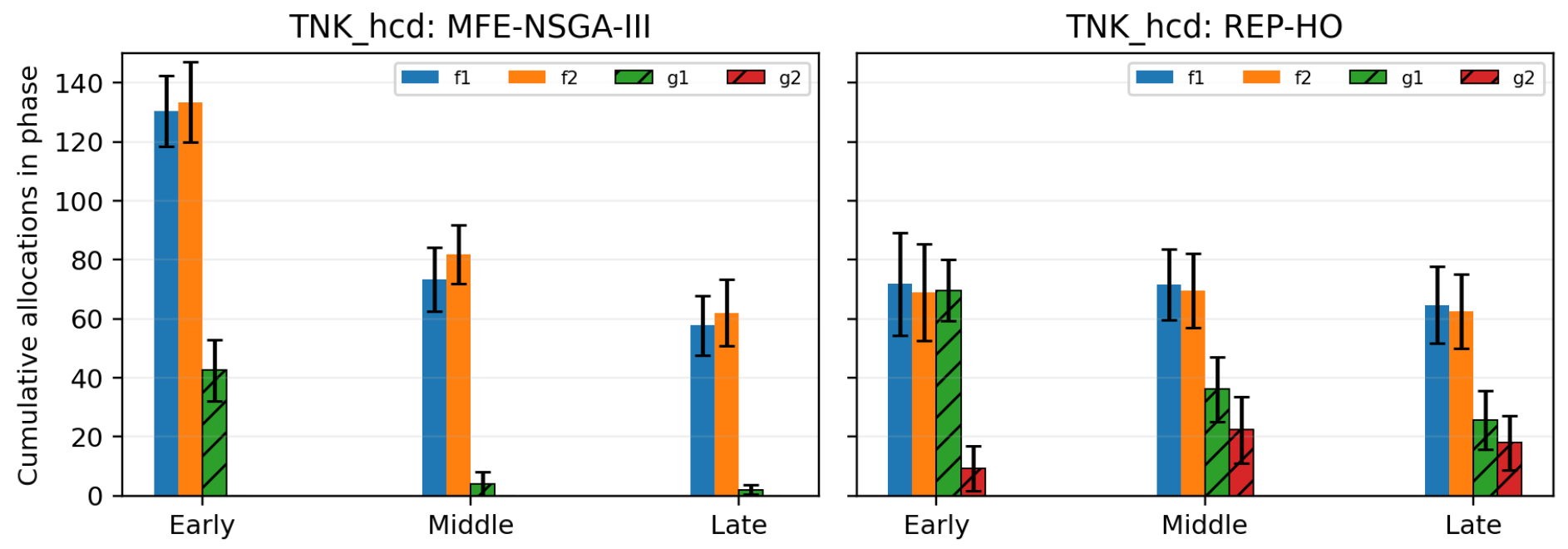}
    \end{minipage}
    \begin{minipage}{0.495\textwidth}
        \centering
        \includegraphics[width=\linewidth]{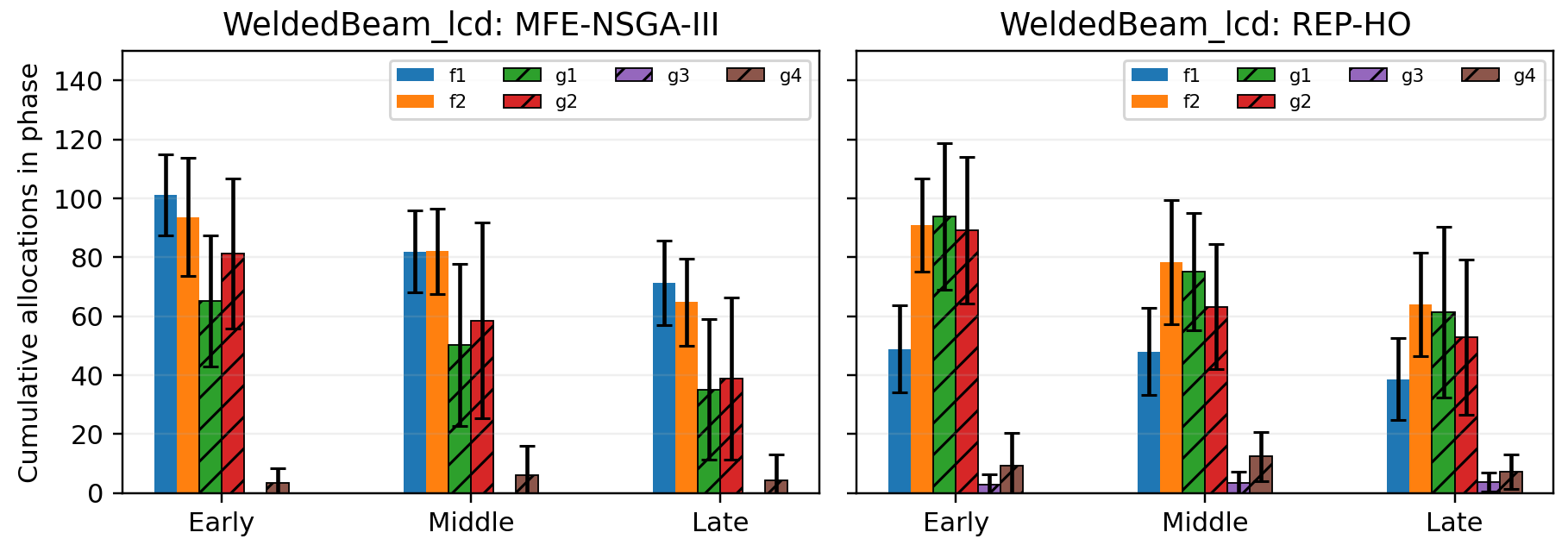}
    \end{minipage}
    \hfill
    \begin{minipage}{0.495\textwidth}
        \centering
        \includegraphics[width=\linewidth]{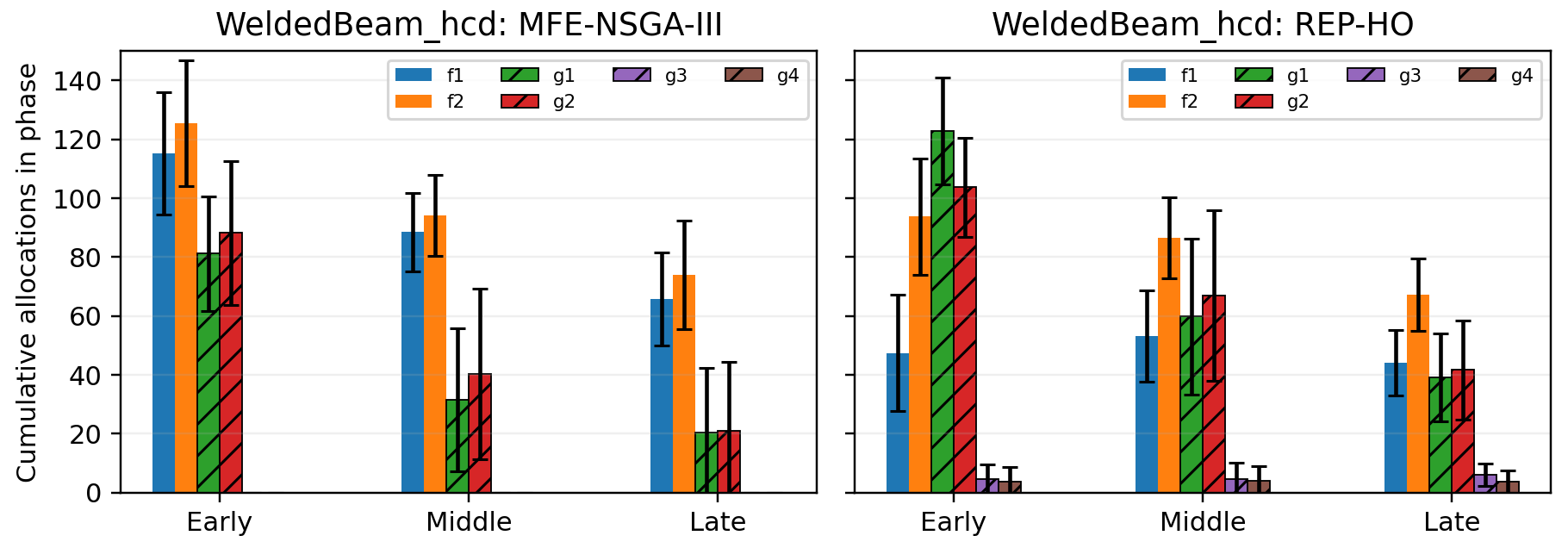}
    \end{minipage}

    \caption{Cumulative allocations on objectives and constraints, the first row is TNK and second row is WeldedBeam problem; the left images are of low constraint difficulty and the right images are of high constraint difficulty.}
    \label{fig:regime2-cumu-alloc}
\end{figure}
\begin{figure}[h]
    \centering
    \begin{minipage}{0.4\textwidth}
        \centering
        \includegraphics[width=\linewidth]{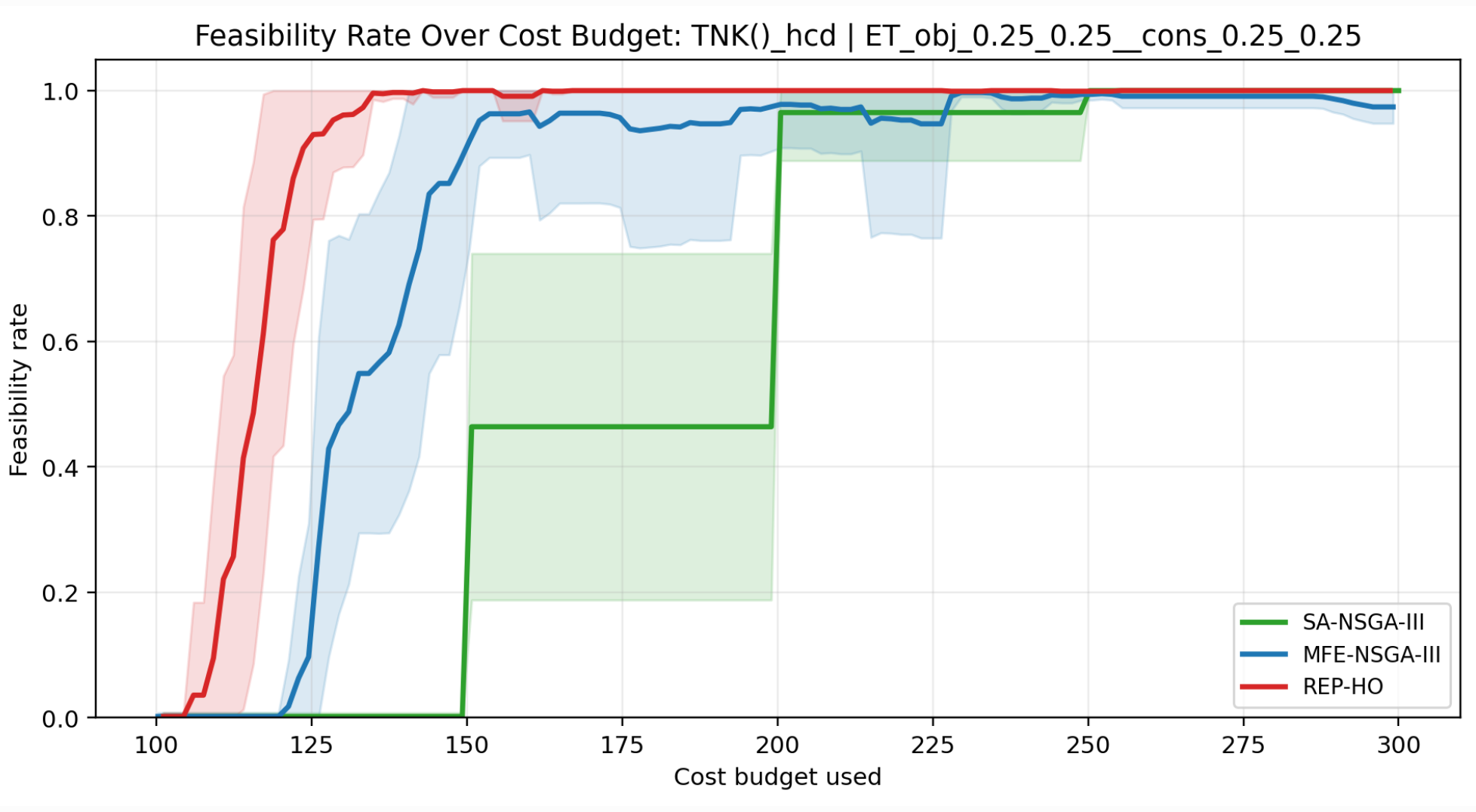}
    \end{minipage}
    \hspace{0.01\textwidth}
    \begin{minipage}{0.4\textwidth}
        \centering
        \includegraphics[width=\linewidth]{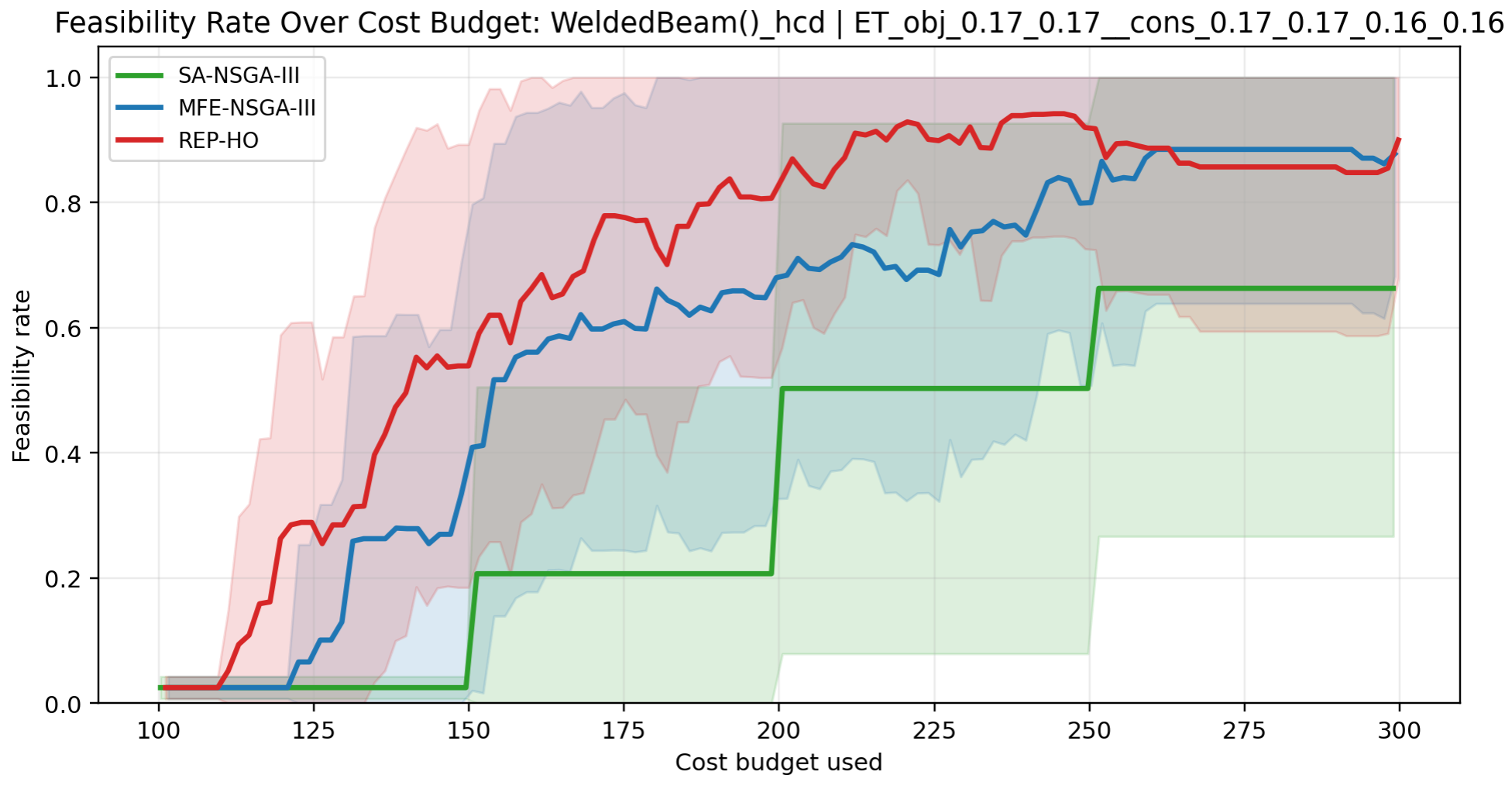}
    \end{minipage}

    \caption{Feasibility rate convergence, left: TNK with high constraint difficulty, right: WeldedBeam with high constraint difficulty}
    \label{fig:regime2-feasi-conv}
\end{figure}
Table~\ref{tab:regime2-performance} summarizes the easy and hard constraint settings. REP-HO(Qwen) obtains the best mean ranking (2.000), just ahead of REP-HO(Deepseek) (2.062), then MFE-NSGA-III (2.375), NSGA-III (3.312), and SA-NSGA-III (3.938). On easy TNK every method reaches IGD+ \(0\), but REP-HO(Qwen) is far ahead on HV. On hard TNK, REP-HO(Qwen) has the best IGD+ while NSGA-III has the best HV. On Welded Beam, REP-HO(Deepseek) has the best IGD+ at both difficulties, and REP-HO(Qwen) has the best HV.

We further analyse the allocation strategies learned by different methods. 
As shown in Figure~\ref{fig:regime2-cumu-alloc}, the methods exhibit similar allocation patterns over objectives, but differ substantially in constraint allocation. 
To quantify constraint difficulty, we report the near-boundary rate $\beta$ under different $\delta$ values in Table~\ref{tab:regime2-delta-calibration}; a larger $\beta$ indicates that a constraint is more frequently close to the feasibility boundary and thus requires more evaluation budget to resolve. For TNK under high constraint difficulty, $g_1$ and $g_2$ have $\beta$ values of 0.0769 and 0.1247, respectively. For WeldedBeam, the corresponding $\beta$ values for $g_1$--$g_4$ are 0.8425, 0.6648, 0.3616, and 0.0167. When the constraint difficulty increases, constraint allocation clearly increases in the early stage for both methods on both problems, indicating that both methods respond to constraint difficulty and allocate early budget to improve feasibility. While MFE-NSGA-III tends to ignore some constraints such as g2 in TNK and g3 g4 in WeldedBeam, REP-HO allocates a non-negligible budget to it, which is more consistent with the estimated constraint difficulty. Figure~\ref{fig:regime2-feasi-conv} further shows that REP-HO achieves better convergence of the feasibility-rate than the baselines. 

 \paragraph{Regime 3}
\begin{table*}[h]
    \centering
    \caption{Regime 3: Fidelity Correlation HV performance summary (HV $\uparrow$).}
    \label{tab:regime3-performance}
    \scriptsize
    \setlength{\tabcolsep}{3pt}
    \resizebox{\textwidth}{!}{%
    \begin{tabular}{p{0.14\textwidth} p{0.16\textwidth} p{0.15\textwidth} p{0.15\textwidth} p{0.16\textwidth} p{0.17\textwidth}}
        \toprule
        Problem & Surrogate Fidelity & SA-NSGA-III & MFE-NSGA-III & REP-HO(Qwen) & REP-HO(Deepseek) \\
        \midrule
        TNK & high & 1.1845 $\pm$ 0.8892 & 1.9192 $\pm$ 0.5321 & \textbf{2.0908 $\pm$ 0.2468} & 1.9613 $\pm$ 0.2645 \\
        TNK & low & 0.2027 $\pm$ 0.4837 & \textbf{0.6993 $\pm$ 0.7125} & 0.4579 $\pm$ 0.7140 & 0.4161 $\pm$ 0.6516 \\
        WeldedBeam & high & 2.5577 $\pm$ 1.3660 & \textbf{3.7361 $\pm$ 0.0713} & 3.4641 $\pm$ 0.8135 & 3.7129 $\pm$ 0.1501 \\
        WeldedBeam & low & 1.5813 $\pm$ 1.1336 & 1.6876 $\pm$ 0.9944 & \textbf{1.9903 $\pm$ 1.1876} & 1.8801 $\pm$ 1.2954 \\
        \midrule
        \multicolumn{2}{l}{Mean ranking integrated with IGD+ in Table \ref{tab:regime3-performance-igd}} & 3.750 & 2.188 & \textbf{1.750} & 2.125 \\
        \bottomrule
    \end{tabular}%
    }
\end{table*}
In Table \ref{tab:regime3-performance}, REP-HO(Qwen) has the best mean ranking (1.750), ahead of REP-HO(Deepseek) (2.125) and MFE-NSGA-III (2.188). It leads both metrics on high-fidelity TNK and low-fidelity Welded Beam. Under the low-fidelity surrogate setting, both MFE-NSGA-III and REP-HO
allocate less budget to constraints than under the high-fidelity
setting. This suggests that lower LF-HF correlation does not simply trigger
more constraint checking. Instead, when the surrogate becomes less reliable,
constraint-side ranking information becomes weaker, so the marginal value of
repeatedly upgrading constraint predictions to high fidelity can decrease as shown in Figure \ref{fig:regime3-cumu-alloc}. This effect is especially clear on Welded Beam. In Table \ref{tab:regime3-fidelity-calibration}, we further compute the Spearman correlation for each objective and constraint under both setting. In the high-fidelity setting,
REP-HO allocates most constraint evaluations to \(g_1\) and \(g_2\), while
\(g_3\) and \(g_4\) receive little budget. This is consistent with the
calibration results in Table \ref{tab:regime3-fidelity-calibration}: \(g_3\) and \(g_4\) have very high Spearman correlations
under high fidelity, so their surrogate predictions are already reliable. Under
low fidelity, however, \(g_1\) and \(g_2\) degrade sharply, while \(g_3\) and
\(g_4\) retain relatively stronger correlation. As a result, REP-HO shifts part
of the constraint budget away from \(g_1/g_2\) and toward \(g_3/g_4\), producing
a more distributed constraint-allocation pattern. Overall, this indicates that REP-HO is not merely increasing constraint evaluations when fidelity worsens, it also adapts allocation according to
where the surrogate still provides useful planning signal.

\paragraph{Regime 4}
\begin{table*}[h]
    \centering
    \caption{Regime 4: Budget HV performance summary (HV $\uparrow$).}
    \label{tab:regime4-performance}
    \scriptsize
    \setlength{\tabcolsep}{3pt}
    \resizebox{\textwidth}{!}{%
    \begin{tabular}{p{0.12\textwidth} p{0.08\textwidth} p{0.12\textwidth} p{0.12\textwidth} p{0.12\textwidth} p{0.13\textwidth} p{0.15\textwidth}}
        \toprule
        Problem & $T_{max}$ & NSGA-III & SA-NSGA-III & MFE-NSGA-III & REP-HO(Qwen) & REP-HO(Deepseek) \\
        \midrule
        MW3 & 200 & 0.2667 $\pm$ 0.6363 & 2.4960 $\pm$ 0.6122 & 2.9173 $\pm$ 0.2242 & 2.8853 $\pm$ 0.3005 & \textbf{2.9391 $\pm$ 0.1424} \\
        MW3 & 400 & 2.4192 $\pm$ 0.2714 & 3.0687 $\pm$ 0.3497 & \textbf{3.3314 $\pm$ 0.1688} & 3.2635 $\pm$ 0.1909 & 3.3148 $\pm$ 0.1797 \\
        MW3 & 600 & 2.6598 $\pm$ 0.2544 & \textbf{3.3510 $\pm$ 0.1495} & 3.0109 $\pm$ 0.0056 & 3.2995 $\pm$ 0.1927 & 3.3439 $\pm$ 0.1551 \\
        TNK & 200 & 2.8096 $\pm$ 0.1397 & 2.4770 $\pm$ 1.2677 & 3.1943 $\pm$ 0.1330 & \textbf{3.2577 $\pm$ 0.0345} & 3.2443 $\pm$ 0.0647 \\
        TNK & 400 & 2.9979 $\pm$ 0.1291 & 2.9807 $\pm$ 0.7573 & 3.2099 $\pm$ 0.0899 & 3.2387 $\pm$ 0.1555 & \textbf{3.2795 $\pm$ 0.0565} \\
        TNK & 600 & 3.0899 $\pm$ 0.1094 & 3.2905 $\pm$ 0.0052 & 3.2573 $\pm$ 0.0673 & \textbf{3.2964 $\pm$ 0.0074} & 3.2842 $\pm$ 0.0282 \\
        WeldedBeam & 200 & \textbf{3.5019 $\pm$ 0.1630} & 2.2884 $\pm$ 1.3171 & 3.0686 $\pm$ 1.1407 & 3.3188 $\pm$ 0.6547 & 3.4573 $\pm$ 0.3199 \\
        WeldedBeam & 400 & 3.6412 $\pm$ 0.1139 & 3.0826 $\pm$ 1.0734 & 3.6368 $\pm$ 0.8365 & \textbf{3.7537 $\pm$ 0.1385} & 3.7410 $\pm$ 0.1804 \\
        WeldedBeam & 600 & 3.7287 $\pm$ 0.1047 & 3.5583 $\pm$ 0.8216 & 3.8385 $\pm$ 0.0369 & 3.7461 $\pm$ 0.3370 & \textbf{3.8602 $\pm$ 0.0689} \\
        \midrule
        \multicolumn{2}{l}{Mean ranking integrated with IGD+ in Table \ref{tab:regime4-performance-igd}} & 4.389 & 3.917 & 2.639 & 2.167 & \textbf{1.889}  \\
        \bottomrule
    \end{tabular}%
    }
\end{table*}
In table~\ref{tab:regime4-performance}, REP-HO(Deepseek) obtains the best mean ranking (1.889), followed by REP-HO(Qwen) (2.167) and MFE-NSGA-III (2.639). REP-HO(Deepseek) leads MW3 at 200 on both metrics and TNK at 400 on both metrics, and it has the best HV on Welded Beam at 600. REP-HO(Qwen) leads TNK at 200. NSGA-III is best only on Welded Beam at 200. Since the total budget differs across settings, we compare objective/constraint allocation percentages rather than raw counts. When the budget is small, both MFE-NSGA-III and REP-HO allocate a larger fraction of their early planning evaluations to constraints. For example, under \(T_{\max}=200\), the early constraint share of REP-HO is \(58.7\%\) on MW3,
\(50.1\%\) on TNK, and \(57.8\%\) on Welded Beam. Under \(T_{\max}=400\), these
shares decrease to \(51.3\%\), \(31.7\%\), and \(56.5\%\), respectively. A
similar trend appears for MFE-NSGA-III, although its constraint share is
generally lower. This pattern indicates that, under tighter budgets, the planner prioritizes
rapid feasibility identification. With fewer opportunities for later
correction, early high-fidelity evaluations are used more aggressively to
resolve constraint uncertainty and avoid spending the limited budget on
infeasible regions. In contrast, when \(T_{\max}=400\), the planner can afford a
more gradual search: constraint allocation remains important in the early stage,
but the later stages shift more budget toward objectives and refinement of the
Pareto front. Thus, REP-HO appears to behave more selectively under low budget
and more comprehensively under high budget, consistent with a budget-aware
planning strategy rather than a fixed allocation rule.

\paragraph{Regime 5}

Table~\ref{tab:regime5-conflict-diagnostics} separates the test problems into
lower-conflict cases (ZDT1 and BNH) and higher-conflict cases (Kursawe and OSY)
using Pareto-front geometry. ZDT1 and BNH have nearly flat fronts, with very
small mean second-difference and turning-angle values, whereas Kursawe and OSY
show much stronger curvature and nonlinear trade-off structure. The results in Table~\ref{tab:regime5-performance} shows that REP-HO(Deepseek) has the best mean ranking (1.958), slightly ahead of MFE-NSGA-III (2.062), with REP-HO(Qwen) third (2.708). REP-HO(Deepseek) is best on every OSY setting. On the lower-conflict problems, MFE-NSGA-III and the two planners remain close. Overall Objective-conflict geometry is not where REP-HO gains its largest advantage, but REP-HO(Deepseek) is still the strongest method overall in this regime.

We also provide an additional ablation study to test the LLM planner's effect: a budget-matched random allocator, given the same candidate set and generation-level budget, ranks behind both REP-HO variants on Regime~1 (overall score 3.250, against 2.250 for REP-HO(Deepseek) and 2.417 for REP-HO(Qwen); Appendix~\ref{app:random-allocation}). The cost of REP-HO is less than 0.11 dollar per experiment, as reported in Appendix~\ref{app:llm-cost}, making it a cost-efficient alternative to human-designed metrics.

\section{Conclusion}
REP-HO uses an off-the-shelf LLM as a flexible planner for which objectives and constraints to evaluate under heterogeneous cost and fidelity. It keeps the surrogate-assisted backbone of MFE-NSGA-III, replaces the hand-designed mixed-fidelity rule with in-context planning, and conditions each decision on a short adaptive nearest-neighbor summary of nearby high-fidelity infill points. Across five regimes and two open-weight backbones, REP-HO obtains the best mean ranking in every regime, while the added API cost stays small. These results show that an LLM can serve as a flexible alternative to fixed, heterogeneity-specific selection metrics.

\section{Acknowledgements}
This work is supported by the Engineering and Physical Sciences Research Council [EP/Y030826/1]. The supervisory team of Nian Ran is: Richard Allmendinger, Mauricio  Alvarez, Markus Olhofer, Sebastian Brulin. 

\section{AI Use Statement}
In this work, we did not use generative AI tools for any tasks requiring mandatory disclosure. In particular, generative AI was not used to develop theoretical or conceptual frameworks, formulate mathematical claims, propose or refine hypotheses, design the research methodology or experiments, implement the proposed methods, analyze data, or interpret results.

We used generative AI tools only for tasks for which disclosure is recommended: editing the manuscript to improve grammar, clarity, and readability, and assisting with code organization and cleanup. The AI-assisted code changes were limited to restructuring, formatting, and improving code readability, and did not involve the design or implementation of the proposed methods. All AI-assisted text and code were reviewed by the authors.

We take full responsibility for the final content of this work, including all text, claims, code, and other artifacts produced with the aid of generative AI.
\bibliography{neurips_2026}

@article{santoshkumar2026-mfe-nsga-iii,
  title={Handling of objectives and constraints with heterogeneous evaluation times for surrogate-assisted evolutionary multi-and many-objective optimization},
  author={Santoshkumar, Balija and Deb, Kalyanmoy},
  journal={Swarm and Evolutionary Computation},
  volume={100},
  pages={102260},
  year={2026},
  publisher={Elsevier}
}

@article{loka2023bi-chvei,
  title={Bi-objective Bayesian optimization of engineering problems with cheap and expensive cost functions},
  author={Loka, Nasrulloh and Couckuyt, Ivo and Garbuglia, Federico and Spina, Domenico and Van Nieuwenhuyse, Inneke and Dhaene, Tom},
  journal={Engineering with Computers},
  volume={39},
  number={3},
  pages={1923--1933},
  year={2023},
  publisher={Springer}
}

@article{de2024ioc-samo-cobra,
  title={Parallel multi-objective optimization for expensive and inexpensive objectives and constraints},
  author={De Winter, Roy and Milatz, Bas and Blank, Julian and Van Stein, Niki and B{\"a}ck, Thomas and Deb, Kalyanmoy},
  journal={Swarm and Evolutionary Computation},
  volume={86},
  pages={101508},
  year={2024},
  publisher={Elsevier}
}

@article{xue2021dva-esto,
  title={Evolutionary sequential transfer optimization for objective-heterogeneous problems},
  author={Xue, Xiaoming and Yang, Cuie and Hu, Yao and Zhang, Kai and Cheung, Yiu-Ming and Song, Linqi and Tan, Kay Chen},
  journal={IEEE Transactions on Evolutionary Computation},
  volume={26},
  number={6},
  pages={1424--1438},
  year={2021},
  publisher={IEEE}
}

@article{cosson2024mnklandscape,
  title={On bi-objective combinatorial optimization with heterogeneous objectives},
  author={Cosson, Rapha{\"e}l and Santana, Roberto and Derbel, Bilel and Liefooghe, Arnaud},
  journal={European Journal of Operational Research},
  volume={319},
  number={1},
  pages={89--101},
  year={2024},
  publisher={Elsevier}
}

@article{wang2023sbp-bo,
  title={Alleviating search bias in Bayesian evolutionary optimization with many heterogeneous objectives},
  author={Wang, Xilu and Jin, Yaochu and Schmitt, Sebastian and Olhofer, Markus},
  journal={IEEE Transactions on Systems, Man, and Cybernetics: Systems},
  volume={54},
  number={1},
  pages={143--155},
  year={2023},
  publisher={IEEE}
}

@article{pang2023hhcbo,
  title={Surrogate-assisted expensive constrained Bi-objective optimization with highly heterogeneous evaluations},
  author={Pang, Yong and Lai, Xiaonan and Wang, Yitang and He, Xiwang and Zhang, Shuai and Song, Xueguan},
  journal={Swarm and Evolutionary Computation},
  volume={83},
  pages={101401},
  year={2023},
  publisher={Elsevier}
}

@inproceedings{rahi2024spareacost,
  title={Towards solving expensive optimization problems with heterogeneous constraint costs},
  author={Rahi, Kamrul and Singh, Hemant and Ray, Tapabrata},
  booktitle={Proceedings of the Genetic and Evolutionary Computation Conference Companion},
  pages={2032--2040},
  year={2024}
}

@article{gu2024lce-saea,
  title={A local correlation estimation surrogate-assisted bi-objective evolutionary algorithm for heterogeneous objectives},
  author={Gu, Chenyan and Wang, Handing},
  journal={Applied Soft Computing},
  volume={151},
  pages={111175},
  year={2024},
  publisher={Elsevier}
}

@inproceedings{cosson2022nmklandscape,
  title={Multi-objective NK landscapes with heterogeneous objectives},
  author={Cosson, Rapha{\"e}l and Santana, Roberto and Derbel, Bilel and Liefooghe, Arnaud},
  booktitle={Proceedings of the Genetic and Evolutionary Computation Conference},
  pages={502--510},
  year={2022}
}

@article{santoshkumar2025het-nsga-iii,
  title={Handling objectives with heterogeneous evaluation times in surrogate-assisted evolutionary multi-objective optimization},
  author={Santoshkumar, Balija and Deb, Kalyanmoy},
  journal={IEEE Transactions on Evolutionary Computation},
  year={2025},
  publisher={IEEE}
}

@inproceedings{tan2023T-sagp,
  title={Expensive Multi-Objective of Pre-Oxidation Process Parameter Optimization Considering Heterogeneity},
  author={Tan, Muyao and Jin, Liang and Li, Kunlun},
  booktitle={2023 28th International Conference on Automation and Computing (ICAC)},
  pages={1--6},
  year={2023},
  organization={IEEE}
}

@article{wang2021trsaea,
  title={Transfer learning based surrogate assisted evolutionary bi-objective optimization for objectives with different evaluation times},
  author={Wang, Xilu and Jin, Yaochu and Schmitt, Sebastian and Olhofer, Markus and Allmendinger, Richard},
  journal={Knowledge-Based Systems},
  volume={227},
  pages={107190},
  year={2021},
  publisher={Elsevier}
}

@incollection{allmendinger2023heterogeneous,
  title={Heterogeneous objectives: state-of-the-art and future research},
  author={Allmendinger, Richard and Knowles, Joshua},
  booktitle={Many-Criteria Optimization and Decision Analysis: State-of-the-Art, Present Challenges, and Future Perspectives},
  pages={317--335},
  year={2023},
  publisher={Springer}
}

@inproceedings{xing2025camo,
  title={CAMO: Convergence-Aware Multi-Fidelity Bayesian Optimization},
  author={Xing, Wei W and Zhenjie, Lu and Shah, Akeel},
  booktitle={The Thirty-ninth Annual Conference on Neural Information Processing Systems},
  year={2025}
}

@inproceedings{yen2025mfms-bo,
  title={Data mixture optimization: A multi-fidelity multi-scale bayesian framework},
  author={Yen, Thomson and Siah, Andrew Wei Tung and Chen, Haozhe and Peng, Tianyi and Guetta, Daniel and Namkoong, Hongseok},
  booktitle={The Thirty-ninth Annual Conference on Neural Information Processing Systems},
  year={2025}
}

@article{foumani2023mfca-bo,
  title={Multi-fidelity cost-aware Bayesian optimization},
  author={Foumani, Zahra Zanjani and Shishehbor, Mehdi and Yousefpour, Amin and Bostanabad, Ramin},
  journal={Computer Methods in Applied Mechanics and Engineering},
  volume={407},
  pages={115937},
  year={2023},
  publisher={Elsevier}
}

@inproceedings{eckmann2024mf-lal,
  title={MF-LAL: Drug Compound Generation Using Multi-Fidelity Latent Space Active Learning},
  author={Eckmann, Peter and Wu, Dongxia and Heinzelmann, Germano and Gilson, Michael K and Yu, Rose},
  booktitle={Forty-second International Conference on Machine Learning},
  year={2025}
}

@inproceedings{niu2024mfrnp,
  title={Multi-fidelity residual neural processes for scalable surrogate modeling},
  author={Niu, Ruijia and Wu, Dongxia and Kim, Kai and Ma, Yi-An and Watson-Parris, Duncan and Yu, Rose},
  booktitle={The Forty-First International Conference on Machine Learning},
  year={2024}
}

@inproceedings{zhang2024rmfbo-dp,
  title={Robust multi-fidelity bayesian optimization with deep kernel and partition},
  author={Zhang, Fengxue and Desautels, Thomas and Chen, Yuxin},
  booktitle={NeurIPS 2024 Workshop on Bayesian Decision-making and Uncertainty},
  year={2024}
}

@inproceedings{salinas2025tuning,
  title={Tuning LLM judge design decisions for 1/1000 of the cost},
  author={Salinas, David and Swelam, Omar and Hutter, Frank},
  booktitle={Forty-second International Conference on Machine Learning},
  year={2025}
}

@article{do2025multifidelity-survey,
  title={Multifidelity bayesian optimization: a review},
  author={Do, Bach and Zhang, Ruda},
  journal={AIAA Journal},
  volume={63},
  number={6},
  pages={2286--2322},
  year={2025},
  publisher={American Institute of Aeronautics and Astronautics}
}

@inproceedings{schuetz2024resum,
  title={RESuM: Rare Event Surrogate Model for Physics Detector Design},
  author={Schuetz, Ann-Kathrin and Poon, Alan WP and Li, Aobo},
  booktitle={The Thirteenth International Conference on Learning Representations},
  year={2025}
}

@inproceedings{ngo2026lmabo,
  title={Adaptive Acquisition Selection for Bayesian Optimization with Large Language Models},
  author={Ngo, Giang and Trong, Dat Phan and Nguyen, Dang and Gupta, Sunil and Venkatesh, Svetha},
  booktitle={The Fourteenth International Conference on Learning Representations},
  year={2026}
}

@inproceedings{liu2024llambo,
  title={Large language models to enhance bayesian optimization},
  author={Liu, Tennison and Astorga, Nicol{\'a}s and Seedat, Nabeel and van der Schaar, Mihaela},
  booktitle={The Twelfth International Conference on Learning Representations},
  year={2024}
}

@article{ramos2023bayesian,
  title={Bayesian Optimization of Catalysis with In-Context Learning},
  author={Caldas Ramos, Mayk and Michtavy, Shane S and White, Andrew D and Porosoff, Marc D},
  journal={ACS Central Science},
  year={2023},
  publisher={ACS Publications}
}

@inproceedings{zeng2026bolt,
  title={Scaling Multi-Task Bayesian Optimization with Large Language Models},
  author={Zeng, Yimeng and Maus, Natalie and Jones, Haydn Thomas and Tao, Jeffrey and Wan, Fangping and Torres, Marcelo Der Torossian and de la Fuente-Nunez, Cesar and Marcus, Ryan and Bastani, Osbert and Gardner, Jacob R},
  booktitle={The Fourteenth International Conference on Learning Representations},
  year={2026}
}

@article{huang2025autonomous,
  title={Autonomous multi-objective optimization using large language model},
  author={Huang, Yuxiao and Wu, Shenghao and Zhang, Wenjie and Wu, Jibin and Feng, Liang and Tan, Kay Chen},
  journal={IEEE Transactions on Evolutionary Computation},
  year={2025},
  publisher={IEEE}
}

@inproceedings{yao2025meoh,
  title={Multi-objective evolution of heuristic using large language model},
  author={Yao, Shunyu and Liu, Fei and Lin, Xi and Lu, Zhichao and Wang, Zhenkun and Zhang, Qingfu},
  booktitle={Proceedings of the AAAI Conference on Artificial Intelligence},
  volume={39},
  number={25},
  pages={27144--27152},
  year={2025}
}

@inproceedings{ye2025llm-lns,
  title={Large language model-driven large neighborhood search for large-scale milp problems},
  author={Ye, Huigen and Xu, Hua and Yan, An and Cheng, Yaoyang},
  booktitle={Forty-second International Conference on Machine Learning},
  year={2025}
}

@inproceedings{chen2025lecmo,
  title={A Large Language Model-assisted Evolutionary Algorithm for Expensive Constrained Multi-objective Optimization},
  author={Chen, Siyu and Tang, Haoran and Zhang, Jinyuan and Tang, Ke},
  booktitle={Proceedings of the Genetic and Evolutionary Computation Conference Companion},
  pages={347--350},
  year={2025}
}

@article{aglietti2024funbo,
  title={Funbo: Discovering acquisition functions for bayesian optimization with funsearch},
  author={Aglietti, Virginia and Ktena, Ira and Schrouff, Jessica and Sgouritsa, Eleni and Ruiz, Francisco JR and Malek, Alan and Bellot, Alexis and Chiappa, Silvia},
  journal={arXiv preprint arXiv:2406.04824},
  year={2024}
}

@article{ran2025exllm,
  title={ExLLM: Experience-Enhanced LLM Optimization for Molecular Design and Beyond},
  author={Ran, Nian and Wang, Yue and Zhang, Xiaoyuan and Li, Zhongzheng and Ran, Qingsong and Li, Wenhao and Allmendinger, Richard},
  journal={arXiv preprint arXiv:2502.12845},
  year={2025}
}

@article{ran2025mcce,
  title={MCCE: A Framework for Multi-LLM Collaborative Co-Evolution},
  author={Ran, Nian and Li, Zhongzheng and Wang, Yue and Ran, Qingsong and Zhang, Xiaoyuan and Feng, Shikun and Allmendinger, Richard and Zhao, Xiaoguang},
  journal={arXiv preprint arXiv:2510.06270},
  year={2025}
}
\bibliographystyle{iclr2027_conference}

\appendix
\section{Reproducibility}
We release our anonymous code at
\url{https://anonymous.4open.science/r/REP-HO-A870/README.md}.
The repository contains the implementation of REP-HO, baseline methods,
experimental regimes, configuration files, and scripts for generating the main
tables and figures. The LLM backbones used in our experiments are the official
hosted versions of Qwen3.5-35B-A3B and DeepSeek-V4-Flash.
All query parameters are kept at their
official default values, except that we set \texttt{reasoning-effort=none} to
avoid unnecessary overthinking and to keep the planner behavior consistent
across runs. And for all problems the prompt template, parameters are all fixed and recorded in code, demonstrating the generalizability and flexibility of our method. Token use, estimated API cost, and wall-clock time are reported in Appendix~\ref{app:llm-cost}.

\begin{table*}[h]
    \centering
    \caption{Regime 5: Objective Conflict performance summary. }
    \label{tab:regime5-performance}
    \scriptsize
    \setlength{\tabcolsep}{3pt}
    \resizebox{\textwidth}{!}{%
    \begin{tabular}{p{0.10\textwidth} p{0.13\textwidth} p{0.12\textwidth} p{0.12\textwidth} p{0.12\textwidth} p{0.13\textwidth} p{0.15\textwidth}}
        \toprule
        Problem & Setting & NSGA-III & SA-NSGA-III & MFE-NSGA-III & REP-HO(Qwen) & REP-HO(Deepseek) \\
        \midrule
        BNH & constraint-heavy & 3.7813 $\pm$ 0.0121 & 3.8042 $\pm$ 0.0002 & 3.8044 $\pm$ 0.0001 & 3.8044 $\pm$ 0.0001 & \textbf{3.8044 $\pm$ 0.0001} \\
        BNH & uniform & 3.7813 $\pm$ 0.0121 & 3.8042 $\pm$ 0.0002 & \textbf{3.8045 $\pm$ 0.0000} & 3.8044 $\pm$ 0.0001 & 3.8044 $\pm$ 0.0001 \\
        BNH & objective-heavy & 3.7813 $\pm$ 0.0121 & 3.8042 $\pm$ 0.0002 & 3.8044 $\pm$ 0.0001 & 3.8043 $\pm$ 0.0002 & \textbf{3.8044 $\pm$ 0.0001} \\
        Kursawe & $f_2$-heavy & \textbf{2.6705 $\pm$ 0.1693} & 2.3337 $\pm$ 0.6176 & 2.4251 $\pm$ 0.3712 & 2.6131 $\pm$ 0.2896 & 2.6175 $\pm$ 0.3333 \\
        Kursawe & uniform & \textbf{2.6705 $\pm$ 0.1693} & 2.3337 $\pm$ 0.6176 & 2.5635 $\pm$ 0.4923 & 2.4755 $\pm$ 0.5329 & 2.4127 $\pm$ 0.7118 \\
        Kursawe & $f_1$-heavy & \textbf{2.6705 $\pm$ 0.1693} & 2.3337 $\pm$ 0.6176 & 2.6025 $\pm$ 0.3051 & 2.5619 $\pm$ 0.3585 & 2.4714 $\pm$ 0.2334 \\
        OSY & constraint-heavy & 2.1089 $\pm$ 0.3122 & 3.5004 $\pm$ 0.2007 & 3.6209 $\pm$ 0.2162 & 3.6073 $\pm$ 0.0814 & \textbf{3.6257 $\pm$ 0.0757} \\
        OSY & uniform & 2.1089 $\pm$ 0.3122 & 3.5004 $\pm$ 0.2007 & 3.6098 $\pm$ 0.2154 & 3.5980 $\pm$ 0.0718 & \textbf{3.6239 $\pm$ 0.0745} \\
        OSY & objective-heavy & 2.1089 $\pm$ 0.3122 & 3.5004 $\pm$ 0.2007 & 3.5602 $\pm$ 0.2110 & 3.5967 $\pm$ 0.0703 & \textbf{3.6015 $\pm$ 0.0730} \\
        ZDT1 & $f_2$-heavy & 0.0403 $\pm$ 0.0999 & 3.4211 $\pm$ 0.1563 & 3.5321 $\pm$ 0.1595 & 3.5132 $\pm$ 0.1515 & \textbf{3.5360 $\pm$ 0.1676} \\
        ZDT1 & uniform & 0.0403 $\pm$ 0.0999 & 3.4211 $\pm$ 0.1563 & 3.5181 $\pm$ 0.1361 & \textbf{3.5466 $\pm$ 0.1040} & 3.5377 $\pm$ 0.1511 \\
        ZDT1 & $f_1$-heavy & 0.0403 $\pm$ 0.0999 & 3.4211 $\pm$ 0.1563 & \textbf{3.5829 $\pm$ 0.0371} & 3.5196 $\pm$ 0.1724 & 3.5660 $\pm$ 0.1038 \\
        \hline
        \multicolumn{2}{l}{Ranking score} & 4.042 & 4.229 & 2.062 & 2.708 & \textbf{1.958} \\
        \bottomrule
    \end{tabular}%
    }
\end{table*}

\section{Additional Related work}
\label{app:additional-related-work}
\paragraph{LLMs for Optimization Planning}
Recent work has begun to explore the use of large language models as decision-making components within optimization loops. 
In Bayesian optimization, LLMs have been used to improve warm-starting, surrogate modelling, candidate generation, and representation learning from textual or structured problem descriptions~\citep{liu2024llambo,ramos2023bayesian,zeng2026bolt}. 
These methods show that pretrained language models can provide useful prior knowledge for expensive black-box optimization, but the LLM is usually used as a surrogate, representation model, initializer, or candidate proposer rather than as an online planner for evaluation decisions. 
More closely related to our motivation, recent work has considered LLMs for controlling parts of the optimization process. 
For example, LMABO~\citep{ngo2026lmabo} uses an LLM to select acquisition functions online from the current Bayesian optimization state, while FunBO~\citep{aglietti2024funbo} uses LLMs to discover acquisition functions that can replace manually designed rules. 
Similarly, in evolutionary and heuristic optimization, LLMs have been used to generate search operators, heuristics, or neighborhood-selection policies~\citep{huang2025autonomous,yao2025meoh,ye2025llm-lns}. 
These studies suggest that LLMs can act not only as black-box optimizers, but also as reasoning modules that replace hand-crafted search or control rules.

However, existing LLM-based optimization methods are mostly studied in single-objective, unconstrained, or task-specific settings, and rarely address heterogeneous evaluation allocation. 
A closely related recent work is LECMO~\citep{chen2025lecmo}, which applies an LLM to expensive constrained multi-objective optimization. 
However, its LLM is mainly used to approximate objective and constraint values within a fixed two-stage evolutionary framework, rather than to decide which heterogeneous evaluations should be performed. 
In contrast, our work investigates whether an off-the-shelf LLM can serve as an online evaluation planner for heterogeneous multi-objective constrained optimization, reasoning over candidates, objectives, constraints, fidelities, costs, and remaining budget without additional training or problem-specific tuning.

\section{Detail of designing five regimes}
\label{app:5regime}

This appendix gives the full design of the five experimental regimes. Unless a
regime explicitly changes a factor, the default configuration is:
\begin{itemize}
    \item total budget \(T_{\max}=300\);
    \item uniform evaluation-cost setting;
    \item fixed surrogate family across all compared methods;
\end{itemize}
All cost vectors are normalized so that evaluating all objectives and
constraints once costs one time unit as shown in Section \ref{sec:method}.

\subsection{Regime 1: Cost Imbalance}
\label{app:regime-cost-imbalance}

This regime varies the distribution of high-fidelity evaluation cost across
objectives and constraints. The purpose is to test whether the planner adapts
to objective-heavy, constraint-heavy, and uniform cost structures without
depending on a fixed hand-designed metric preference.

\paragraph{TNK and MW3.}
TNK and MW3 have two objectives and two constraints in this regime. We use:
\[
\begin{aligned}
    s_2 &: (0.25, 0.25 \mid 0.25, 0.25)\\
    s_1 &: (0.70, 0.10 \mid 0.15, 0.05)\\
    s_3 &: (0.15, 0.05 \mid 0.70, 0.10)
\end{aligned}
\]
Here \(s_2\) is uniform, \(s_1\) is objective-heavy, and \(s_3\) is
constraint-heavy. 

\paragraph{Welded Beam.}
The intended design pattern is two objectives and several constraints. Welded Beam has two objectives and four constraints,
so the cost vector is normalized over six evaluation channels. The main-text
settings are:
\[
\begin{aligned}
    \text{WB-uniform} &: (0.17, 0.17 \mid
    0.17, 0.17, 0.16, 0.16)\\
    \text{WB-objective-heavy} &: (0.4211, 0.2632 \mid
    0.1053, 0.0842, 0.0631, 0.0631)\\
    \text{WB-constraint-heavy} &: (0.0941, 0.0824 \mid
    0.2353, 0.2118, 0.1882, 0.1882)
\end{aligned}
\]
These are normalized versions of the desired uniform, objective-heavy, and
constraint-heavy patterns.

\paragraph{Carside.}
Carside is evaluated as a three-objective, ten-constraint problem. The main
settings are:
\[
\begin{aligned}
    \text{Carside-uniform} &: (0.077, 0.077, 0.077 \mid
    0.077, 0.077, 0.077, 0.077, 0.077, 0.077, 0.077, \\0.077, 0.077, 0.077)\\
    \text{Carside-objective-heavy} &: (0.35, 0.22, 0.13 \mid
    0.05, 0.04, 0.04, 0.035, 0.03, 0.03, 0.025, 0.02, 0.015, \\0.015)\\
    \text{Carside-constraint-heavy} &: (0.14, 0.10, 0.06 \mid
    0.13, 0.11, 0.09, 0.08, 0.07, 0.06, 0.05, 0.04, 0.04, 0.03)
\end{aligned}
\] 

\subsubsection{Additional Results}
\begin{table*}[t]
    \centering
    \caption{Regime 1: Cost Imbalance IGD+ performance summary (IGD+ $\downarrow$).}
    \label{tab:regime1-performance-igd}
    \scriptsize
    \setlength{\tabcolsep}{3pt}
    \resizebox{\textwidth}{!}{%
    \begin{tabular}{p{0.12\textwidth} p{0.14\textwidth} p{0.12\textwidth} p{0.12\textwidth} p{0.12\textwidth} p{0.13\textwidth} p{0.15\textwidth}}
        \toprule
        Problem & Setting & NSGA-III & SA-NSGA-III & MFE-NSGA-III & REP-HO(Qwen) & REP-HO(Deepseek) \\
        \midrule
        Carside & uniform & 0.0532 $\pm$ 0.0167 & 0.0031 $\pm$ 0.0015 & \textbf{0.0010 $\pm$ 0.0008} & 0.0013 $\pm$ 0.0009 & 0.0015 $\pm$ 0.0009 \\
        Carside & constraint-heavy & 0.0532 $\pm$ 0.0167 & 0.0031 $\pm$ 0.0015 & 0.0010 $\pm$ 0.0005 & \textbf{0.0010 $\pm$ 0.0008} & 0.0013 $\pm$ 0.0006 \\
        Carside & objective-heavy & 0.0532 $\pm$ 0.0167 & 0.0026 $\pm$ 0.0012 & 0.0011 $\pm$ 0.0006 & \textbf{0.0010 $\pm$ 0.0005} & 0.0013 $\pm$ 0.0007 \\
        MW3 & constraint-heavy & 0.4049 $\pm$ 0.1509 & 0.1058 $\pm$ 0.0857 & 0.0486 $\pm$ 0.0748 & 0.0584 $\pm$ 0.0718 & \textbf{0.0407 $\pm$ 0.0246} \\
        MW3 & uniform & 0.4049 $\pm$ 0.1509 & 0.1058 $\pm$ 0.0857 & 0.0568 $\pm$ 0.0725 & 0.0516 $\pm$ 0.0736 & \textbf{0.0372 $\pm$ 0.0277} \\
        MW3 & objective-heavy & 0.4049 $\pm$ 0.1509 & 0.1058 $\pm$ 0.0857 & 0.0550 $\pm$ 0.0745 & 0.0447 $\pm$ 0.0292 & \textbf{0.0295 $\pm$ 0.0278} \\
        TNK & constraint-heavy & 0.0841 $\pm$ 0.0275 & 0.0333 $\pm$ 0.0408 & 0.0190 $\pm$ 0.0158 & 0.0100 $\pm$ 0.0096 & \textbf{0.0068 $\pm$ 0.0007} \\
        TNK & uniform & 0.0841 $\pm$ 0.0275 & 0.0333 $\pm$ 0.0408 & 0.0224 $\pm$ 0.0166 & \textbf{0.0072 $\pm$ 0.0007} & 0.0100 $\pm$ 0.0091 \\
        TNK & objective-heavy & 0.0841 $\pm$ 0.0275 & 0.0333 $\pm$ 0.0408 & 0.0236 $\pm$ 0.0169 & \textbf{0.0100 $\pm$ 0.0088} & 0.0133 $\pm$ 0.0136 \\
        WeldedBeam & constraint-heavy & 0.0916 $\pm$ 0.0319 & 0.2065 $\pm$ 0.1348 & \textbf{0.0605 $\pm$ 0.0526} & 0.1049 $\pm$ 0.1320 & 0.1758 $\pm$ 0.3441 \\
        WeldedBeam & uniform & 0.0916 $\pm$ 0.0319 & 0.4889 $\pm$ 0.4804 & \textbf{0.0743 $\pm$ 0.0638} & 0.0870 $\pm$ 0.0676 & 0.0822 $\pm$ 0.0486 \\
        WeldedBeam & objective-heavy & 0.0916 $\pm$ 0.0319 & 0.4889 $\pm$ 0.4804 & 0.1196 $\pm$ 0.1783 & 0.2241 $\pm$ 0.4696 & \textbf{0.0885 $\pm$ 0.0645} \\
        \midrule
        \multicolumn{2}{l}{Mean ranking integrated with HV in Table \ref{tab:regime1-performance}} & 4.562 & 4.167 & 2.188 & 2.125 & \textbf{1.958} \\
        \bottomrule
    \end{tabular}%
    }
\end{table*}
\subsection{Regime 2: Constraint Difficulty}
\label{app:regime-constraint-difficulty}

This regime varies the size of the feasible region while holding cost,
surrogate fidelity, and budget fixed. The main problems are TNK and Welded
Beam, with uniform costs:
\[
\begin{aligned}
    \text{TNK} &: (0.25,0.25 \mid 0.25,0.25)\\
    \text{Welded Beam} &: (0.17,0.17 \mid 0.17,0.17,0.16,0.16)
\end{aligned}
\]
Each original inequality constraint \(g_j(\mathbf{x})\leq 0\) is shifted by a
shared slack parameter \(\delta\):
\begin{equation}
    g'_j(\mathbf{x}) = g_j(\mathbf{x}) - \delta \leq 0 
    \label{eq:appendix-constraint-shift}
\end{equation}
Larger \(\delta\) relaxes the constraints and increases the feasible region,
whereas smaller \(\delta\) tightens them. The value of \(\delta\) is chosen
by scanning. For each problem, we draw 30,000 Latin hypercube samples and scan
a problem-specific grid of \(\delta\) values. For each candidate \(\delta\), the
pilot feasible ratio is
\begin{equation}
    r_{\mathrm{feas}}(\delta)
    =
    \frac{1}{N_{\mathrm{pilot}}}
    \sum_{i=1}^{N_{\mathrm{pilot}}}
    \mathbb{I}\left[
    g_j(\mathbf{x}_i)\leq \delta,\ \forall j
    \right]
    \label{eq:appendix-feasible-ratio}
\end{equation}
The easy setting selects the scanned \(\delta\) whose feasible ratio lies in
25\% and is closest to the midpoint of that band. The hard setting selects
the scanned \(\delta\) whose feasible ratio lies in 0.1\% or 1\% and is closest to
the midpoint of that band. The reason we choose that small ratio for hard settings is that we found that these ratios can effectively increase the difficulty of increasing feasibility, while a larger value such as 5\% or 3\% is not effective enough. 
We additionally near-boundary rate of each constraint to classify the constraints into hard and easy constraints, here, “hard” means that a constraint is frequently near the boundary and therefore requires more frequent high-fidelity evaluation. The selected \(\delta\) and its information are recorded in Table \ref{tab:regime2-delta-calibration}.
\begin{table}[t]
    \centering
    \caption{Constraint-difficulty calibration for Regime 2. Feasible ratios
    and near-boundary rates are estimated from the 30,000-point LHS pilot scan.}
    \label{tab:regime2-delta-calibration}
    \footnotesize
    \begin{tabular}{llrcp{0.46\linewidth}}
        \toprule
        Problem & Level & \(\delta\) & Feasible ratio & Near-boundary rate by constraint \\
        \midrule
        TNK & Easy & 1.80 & 0.2546 & \(g_1:0.0000,\ g_2:0.0624\) \\
        TNK & Normal & 0.00 & 0.0499 & \(g_1:0.0766,\ g_2:0.0860\) \\
        TNK & Hard & -0.45 & 0.0111 & \(g_1:0.0769,\ g_2:0.1247\) \\
        Welded Beam & Easy & 0.05 & 0.3560 &
        \(g_1:0.1340,\ g_2:0.1639,\ g_3:0.7463,\ g_4:0.0239\) \\
        Welded Beam & Normal & 0.00 & 0.3190 &
        \(g_1:0.1473,\ g_2:0.1793,\ g_3:0.7500,\ g_4:0.0224\) \\
        Welded Beam & Hard & -0.65 & 0.0011 &
        \(g_1:0.8425,\ g_2:0.6648,\ g_3:0.3616,\ g_4:0.0167\) \\
        \bottomrule
    \end{tabular}
\end{table}

\subsubsection{Additional Results}
\begin{table*}[h]
    \centering
    \caption{Regime 2: Constraint Difficulty IGD+ performance summary (IGD+ $\downarrow$).}
    \label{tab:regime2-performance-igd}
    \scriptsize
    \setlength{\tabcolsep}{3pt}
    \resizebox{\textwidth}{!}{%
    \begin{tabular}{p{0.11\textwidth} p{0.15\textwidth} p{0.12\textwidth} p{0.12\textwidth} p{0.12\textwidth} p{0.13\textwidth} p{0.15\textwidth}}
        \toprule
        Problem & Constraint Difficulty & NSGA-III & SA-NSGA-III & MFE-NSGA-III & REP-HO(Qwen) & REP-HO(Deepseek) \\
        \midrule
        TNK & hard & 0.4224 $\pm$ 0.0548 & 0.3613 $\pm$ 0.0476 & 0.3437 $\pm$ 0.0295 & \textbf{0.3376 $\pm$ 0.0100} & 0.3457 $\pm$ 0.0372 \\
        TNK & easy & \textbf{0.0000 $\pm$ 0.0000} & \textbf{0.0000 $\pm$ 0.0000} & \textbf{0.0000 $\pm$ 0.0000} & \textbf{0.0000 $\pm$ 0.0000} & \textbf{0.0000 $\pm$ 0.0000} \\
        WeldedBeam & hard & 0.5541 $\pm$ 0.0572 & 0.4981 $\pm$ 0.0413 & 0.4789 $\pm$ 0.0582 & 0.4965 $\pm$ 0.0809 & \textbf{0.4756 $\pm$ 0.0390} \\
        WeldedBeam & easy & 0.0943 $\pm$ 0.0445 & 0.3056 $\pm$ 0.6120 & 0.1770 $\pm$ 0.4074 & 0.1130 $\pm$ 0.1630 & \textbf{0.0666 $\pm$ 0.0529} \\
        \midrule
        \multicolumn{2}{l}{Mean ranking integrated with IGD+ in Table \ref{tab:regime2-performance}} & 3.312 & 3.938 & 2.375 & \textbf{2.000} & 2.062 \\
        \bottomrule
    \end{tabular}%
    }
\end{table*}

\subsection{Regime 3: Fidelity Correlation}
\label{app:regime-fidelity-correlation}

This regime controls how reliable the low-fidelity surrogate is. The main
problems are TNK and Welded Beam, with uniform costs and \(T_{\max}=300\). The
high-correlation setting uses a stronger low-fidelity model, while the
low-correlation setting weakens the model through model choice and mild label degradation with Gaussian noise. Calibration is performed with a fixed surrogate family and an independent
reference pool. Let \(y_k^{\mathrm{HF}}\) denote true high-fidelity values for
target \(k\) on the reference pool, and let \(y_k^{\mathrm{LF}}\) denote the
corresponding surrogate predictions. For each objective, we compute
\begin{equation}
    \rho_k^{F}
    =
    \mathrm{Spearman}
    \left(
    y_k^{\mathrm{HF}}, y_k^{\mathrm{LF}}
    \right).
    \label{eq:appendix-objective-Spearman}
\end{equation}
For each constraint, we compute the analogous Spearman correlation and the mean
absolute error:
\begin{equation}
    \mathrm{MAE}_j^{G}
    =
    \frac{1}{N_{\mathrm{ref}}}
    \sum_{i=1}^{N_{\mathrm{ref}}}
    \left|
    g_j^{\mathrm{HF}}(\mathbf{x}_i) -
    g_j^{\mathrm{LF}}(\mathbf{x}_i)
    \right|.
    \label{eq:appendix-constraint-mae}
\end{equation}
The calibration target is mean objective Spearman correlation above 0.8 for
high-correlation settings and approximately 0.3 for low-correlation
settings. Constraint Spearman and constraint MAE are reported to verify that
the constraint-side low-fidelity quality changes consistently. In implementation, candidate fidelity configurations are generated by varying
the model choice, the label-noise level, and a mild structural
bias term. The selected settings and the resulting per-target correlations are
shown together in Table \ref{tab:regime3-fidelity-calibration}. 

In this regime, NSGA-III is removed from comparison because it is unfair to put it here because it is not affected by surrogate fidelity.
\begin{table}[h]
    \centering
    \caption{Surrogate-fidelity calibration for Regime 3. Each setting uses
    \(n_{\mathrm{DOE}}=100\). The per-target entries report LF-HF Spearman
    correlations for individual objectives and constraints, with the
    corresponding mean shown in parentheses.}
    \label{tab:regime3-fidelity-calibration}
    \scriptsize
    \begin{tabular}{p{0.05\linewidth}p{0.05\linewidth}p{0.05\linewidth}p{0.07\linewidth}p{0.07\linewidth}p{0.2\linewidth}p{0.3\linewidth}}
        \toprule
        Problem & Fidelity & Model & Noise std. & Bias strength &
        Objective Spearman & Constraint Spearman \\
        \midrule
        TNK & High & kriging-const & 0.2 & 1.0 &
        \(f_1:0.7667,\ f_2:0.7837\) (mean \(0.7752\)) &
        \(g_1:0.7839,\ g_2:0.7897\) (mean \(0.7868\)) \\
        TNK & Low & kriging-const & 1.0 & 5.0 &
        \(f_1:0.2782,\ f_2:0.2491\) (mean \(0.2637\)) &
        \(g_1:0.3134,\ g_2:0.2905\) (mean \(0.3020\)) \\
        Welded Beam & High & kriging-lin & 0.0 & 0.0 &
        \(f_1:1.0000,\ f_2:0.4932\) (mean \(0.7466\)) &
        \(g_1:0.6545,\ g_2:0.5898,\ g_3:1.0000,\ g_4:0.9967\)
        (mean \(0.8103\)) \\
        Welded Beam & Low & kriging-const & 1.0 & 1.0 &
        \(f_1:0.5688,\ f_2:0.0361\) (mean \(0.3025\)) &
        \(g_1:0.2252,\ g_2:0.1377,\ g_3:0.6434,\ g_4:0.4828\)
        (mean \(0.3723\)) \\
        \bottomrule
    \end{tabular}
\end{table}

\subsubsection{Additional analysis}
\begin{table*}[h]
    \centering
    \caption{Regime 3: Fidelity Correlation IGD+ performance summary (IGD+ $\downarrow$).}
    \label{tab:regime3-performance-igd}
    \scriptsize
    \setlength{\tabcolsep}{3pt}
    \resizebox{\textwidth}{!}{%
    \begin{tabular}{p{0.14\textwidth} p{0.16\textwidth} p{0.15\textwidth} p{0.15\textwidth} p{0.16\textwidth} p{0.17\textwidth}}
        \toprule
        Problem & Surrogate Fidelity & SA-NSGA-III & MFE-NSGA-III & REP-HO(Qwen) & REP-HO(Deepseek) \\
        \midrule
        TNK & high & 0.4083 $\pm$ 0.0937 & 0.3396 $\pm$ 0.0926 & \textbf{0.3044 $\pm$ 0.0806} & 0.3469 $\pm$ 0.1026 \\
        TNK & low & 0.5767 $\pm$ 0.0292 & 0.5631 $\pm$ 0.0825 & \textbf{0.4813 $\pm$ 0.1105} & 0.5544 $\pm$ 0.1119 \\
        WeldedBeam & high & 0.3335 $\pm$ 0.4246 & \textbf{0.0697 $\pm$ 0.0297} & 0.0919 $\pm$ 0.0621 & 0.0761 $\pm$ 0.0524 \\
        WeldedBeam & low & 0.8589 $\pm$ 0.6880 & 0.7401 $\pm$ 0.5475 & \textbf{0.6253 $\pm$ 0.5352} & 0.7342 $\pm$ 0.8379 \\
        \midrule
        \multicolumn{2}{l}{Mean ranking integrated with IGD+ in Table \ref{tab:regime3-performance}} & 3.750 & 2.188 & \textbf{1.750} & 2.125 \\
        \bottomrule
    \end{tabular}%
    }
\end{table*}

\begin{figure}[h]
    \centering
    \begin{minipage}{0.495\textwidth}
        \centering
        \includegraphics[width=\linewidth]{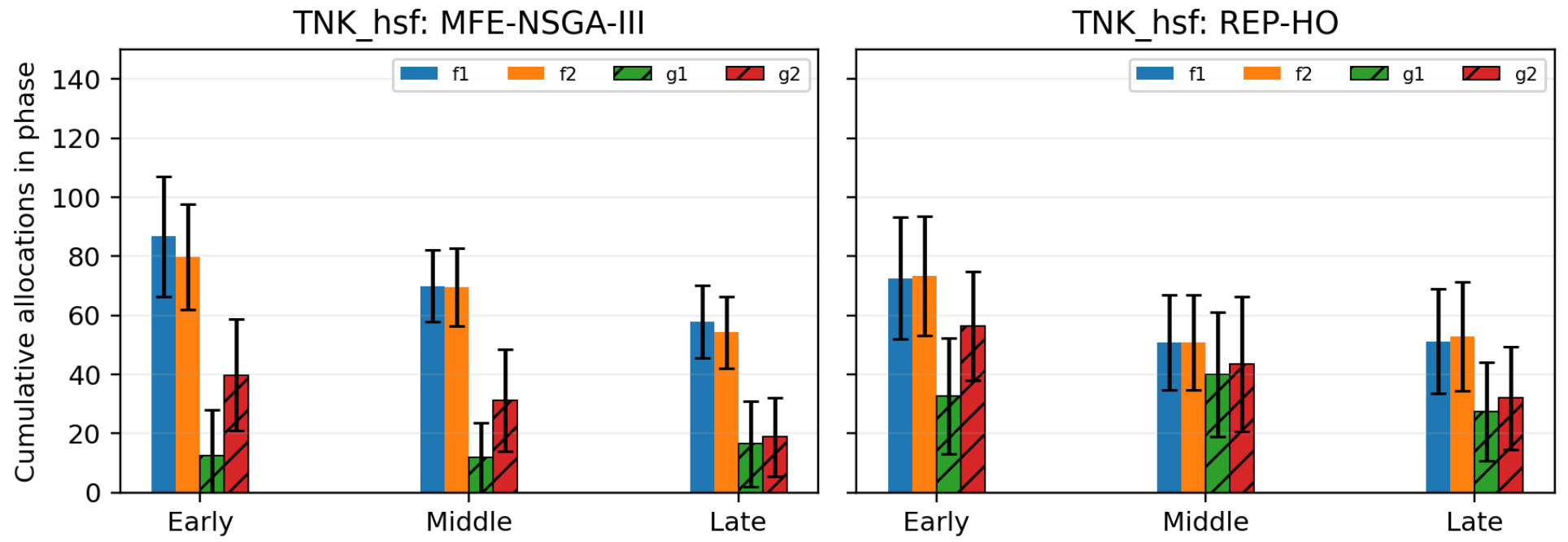}
    \end{minipage}
    \hfill
    \begin{minipage}{0.495\textwidth}
        \centering
        \includegraphics[width=\linewidth]{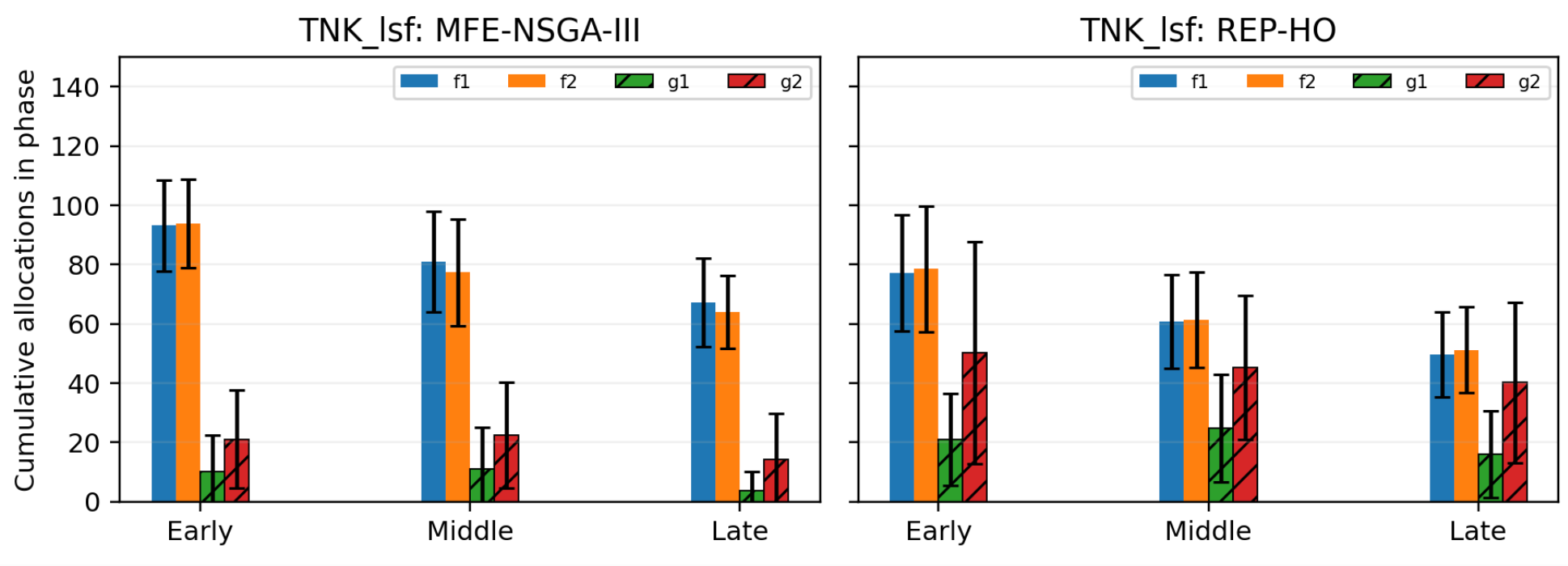}
    \end{minipage}
    \begin{minipage}{0.495\textwidth}
        \centering
        \includegraphics[width=\linewidth]{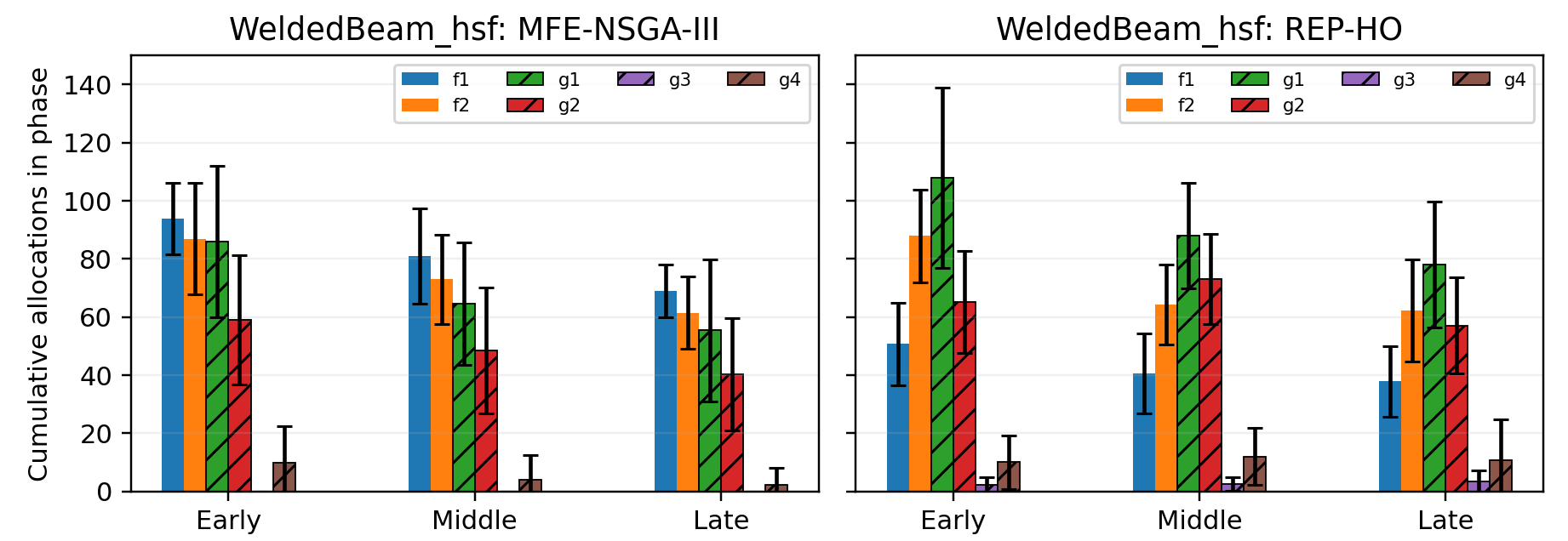}
    \end{minipage}
    \hfill
    \begin{minipage}{0.495\textwidth}
        \centering
        \includegraphics[width=\linewidth]{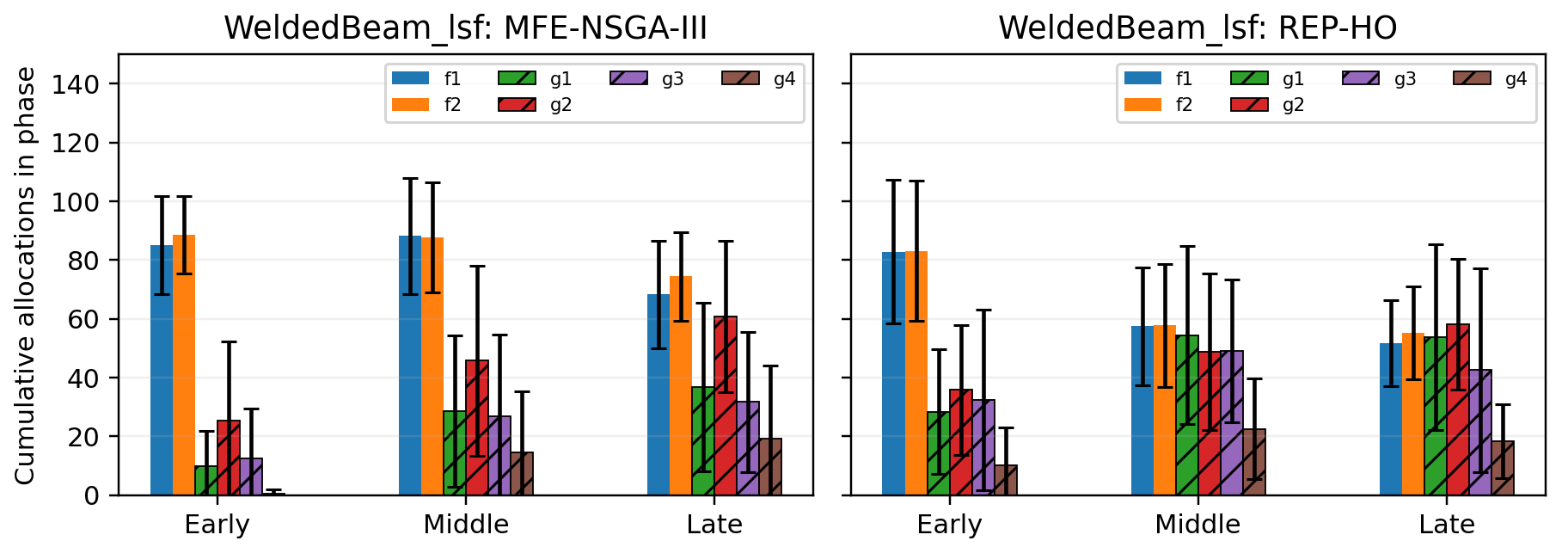}
    \end{minipage}

    \caption{Cumulative allocations on objectives and constraints, the first row is TNK and second row is WeldedBeam problem; the left images are of high surrogate fidelity (easy) and the right images are of low surrogate fidelity (hard).}
    \label{fig:regime3-cumu-alloc}
\end{figure}

\subsection{Regime 4: Budget}
\label{app:regime-budget}
This regime varies the total time budget while holding cost and surrogate
fidelity fixed. The main design uses TNK, MW3, and Carside with:
\begin{equation}
    T_{\max}\in\{200,300,400\}.
    \label{eq:appendix-budget-levels}
\end{equation}
The mid-budget case \(T_{\max}=300\) is with the global default. The
low-budget and high-budget cases test whether the planner becomes more
selective when evaluations are scarce and more comprehensive when evaluations
are plentiful. Costs are set to the uniform setting and fidelity correlation is
set to high-correlation.

We report final IGD+, HV, feasible rate, objective/constraint budget share, and
allocation entropy. For a generation-level allocation distribution
\(\mathbf{p}_t\) over targets, entropy is computed as
\begin{equation}
    H_t = -\sum_k p_{t,k}\log p_{t,k}.
    \label{eq:appendix-allocation-entropy}
\end{equation}
Lower entropy indicates a more concentrated plan, while higher entropy
indicates more even target coverage. We also report sparsity, defined as the
fraction of candidate-target pairs not selected for high-fidelity evaluation.
\subsubsection{Additional Analysis}
\begin{table*}[h]
    \centering
    \caption{Regime 4: Budget IGD+ performance summary (IGD+ $\downarrow$).}
    \label{tab:regime4-performance-igd}
    \scriptsize
    \setlength{\tabcolsep}{3pt}
    \resizebox{\textwidth}{!}{%
    \begin{tabular}{p{0.12\textwidth} p{0.08\textwidth} p{0.12\textwidth} p{0.12\textwidth} p{0.12\textwidth} p{0.13\textwidth} p{0.15\textwidth}}
        \toprule
        Problem & $T_{max}$ & NSGA-III & SA-NSGA-III & MFE-NSGA-III & REP-HO(Qwen) & REP-HO(Deepseek) \\
        \midrule
        MW3 & 200 & 0.4343 $\pm$ 0.0482 & 0.1943 $\pm$ 0.0901 & 0.1011 $\pm$ 0.0707 & 0.1090 $\pm$ 0.0826 & \textbf{0.0962 $\pm$ 0.0326} \\
        MW3 & 400 & 0.2574 $\pm$ 0.0856 & 0.0798 $\pm$ 0.0990 & 0.0278 $\pm$ 0.0301 & 0.0329 $\pm$ 0.0292 & \textbf{0.0276 $\pm$ 0.0277} \\
        MW3 & 600 & 0.1840 $\pm$ 0.0744 & 0.0216 $\pm$ 0.0233 & 0.0729 $\pm$ 0.0019 & 0.0295 $\pm$ 0.0294 & \textbf{0.0214 $\pm$ 0.0232} \\
        TNK & 200 & 0.1135 $\pm$ 0.0271 & 0.0500 $\pm$ 0.0878 & 0.0238 $\pm$ 0.0276 & \textbf{0.0102 $\pm$ 0.0066} & 0.0151 $\pm$ 0.0134 \\
        TNK & 400 & 0.0636 $\pm$ 0.0233 & 0.0466 $\pm$ 0.0991 & 0.0212 $\pm$ 0.0158 & 0.0168 $\pm$ 0.0278 & \textbf{0.0089 $\pm$ 0.0086} \\
        TNK & 600 & 0.0399 $\pm$ 0.0174 & \textbf{0.0066 $\pm$ 0.0005} & 0.0121 $\pm$ 0.0107 & 0.0069 $\pm$ 0.0012 & 0.0086 $\pm$ 0.0033 \\
        WeldedBeam & 200 & \textbf{0.1251 $\pm$ 0.0525} & 0.5967 $\pm$ 0.6835 & 0.2661 $\pm$ 0.4119 & 0.2056 $\pm$ 0.2596 & 0.1639 $\pm$ 0.1197 \\
        WeldedBeam & 400 & 0.0750 $\pm$ 0.0294 & 0.1512 $\pm$ 0.1082 & \textbf{0.0292 $\pm$ 0.0233} & 0.0531 $\pm$ 0.0382 & 0.0557 $\pm$ 0.0524 \\
        WeldedBeam & 600 & 0.0485 $\pm$ 0.0248 & 0.0591 $\pm$ 0.0430 & \textbf{0.0178 $\pm$ 0.0083} & 0.0583 $\pm$ 0.1192 & 0.0206 $\pm$ 0.0192 \\
        \midrule
        \multicolumn{2}{l}{Mean ranking integrated with IGD+ in Table \ref{tab:regime4-performance}} & 4.389 & 3.917 & 2.639 & 2.167 & \textbf{1.889} \\
        \bottomrule
    \end{tabular}%
    }
\end{table*}
\begin{figure}[h]
    \centering
    \begin{minipage}{0.495\textwidth}
        \centering
        \includegraphics[width=\linewidth]{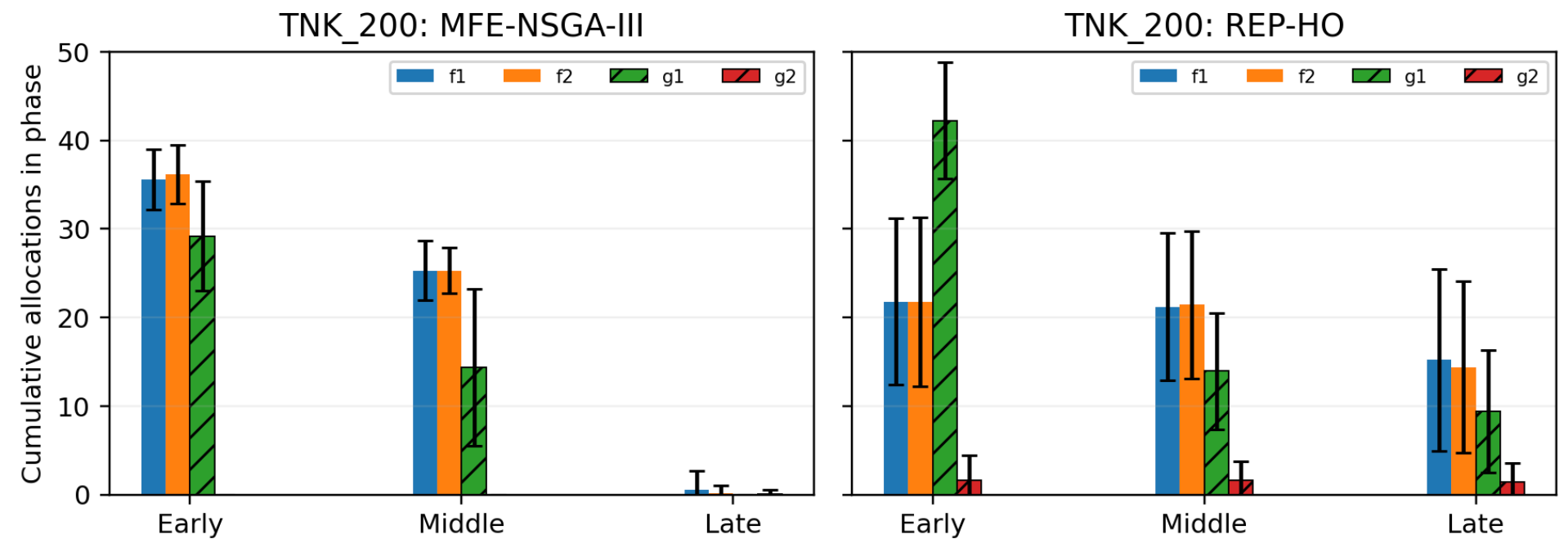}
    \end{minipage}
    \hfill
    \begin{minipage}{0.495\textwidth}
        \centering
        \includegraphics[width=\linewidth]{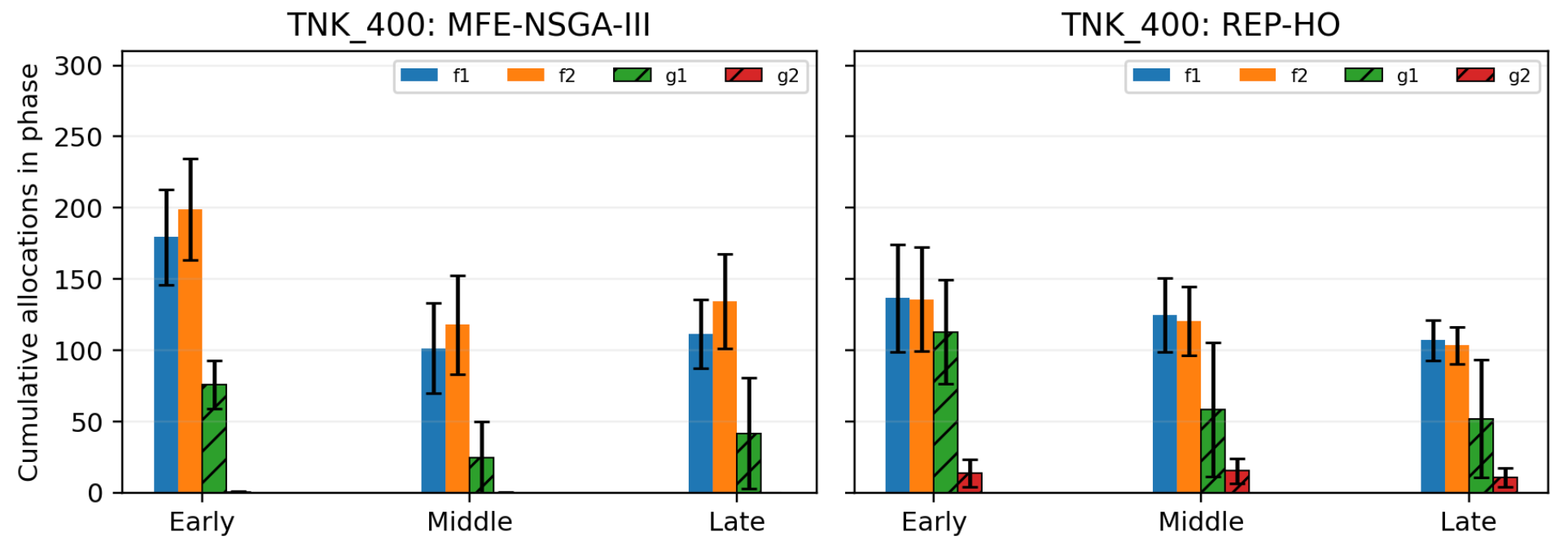}
    \end{minipage}
    \begin{minipage}{0.495\textwidth}
        \centering
        \includegraphics[width=\linewidth]{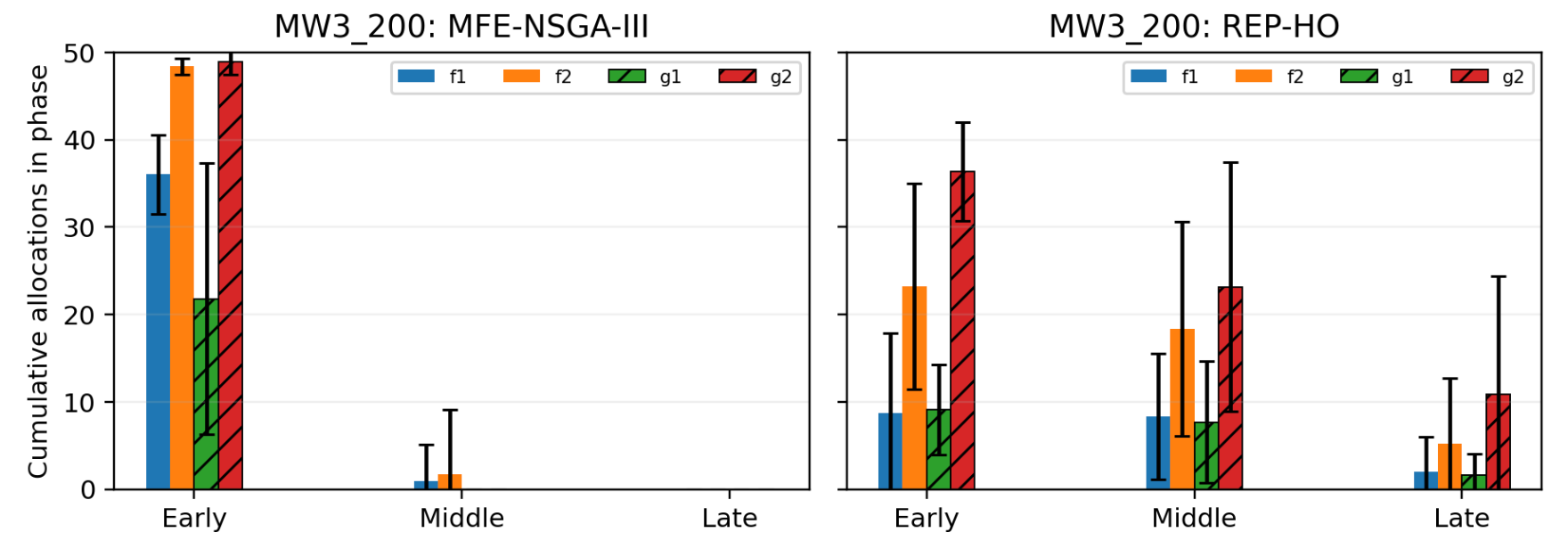}
    \end{minipage}
    \hfill
    \begin{minipage}{0.495\textwidth}
        \centering
        \includegraphics[width=\linewidth]{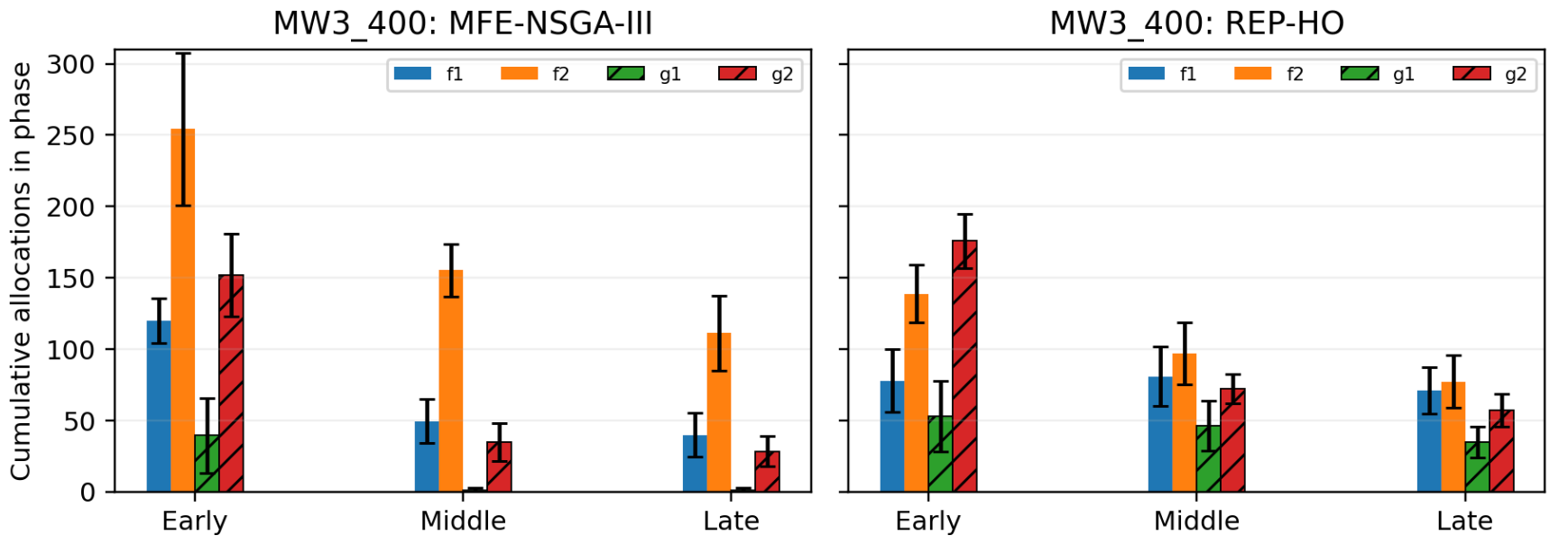}
    \end{minipage}
    \begin{minipage}{0.495\textwidth}
        \centering
        \includegraphics[width=\linewidth]{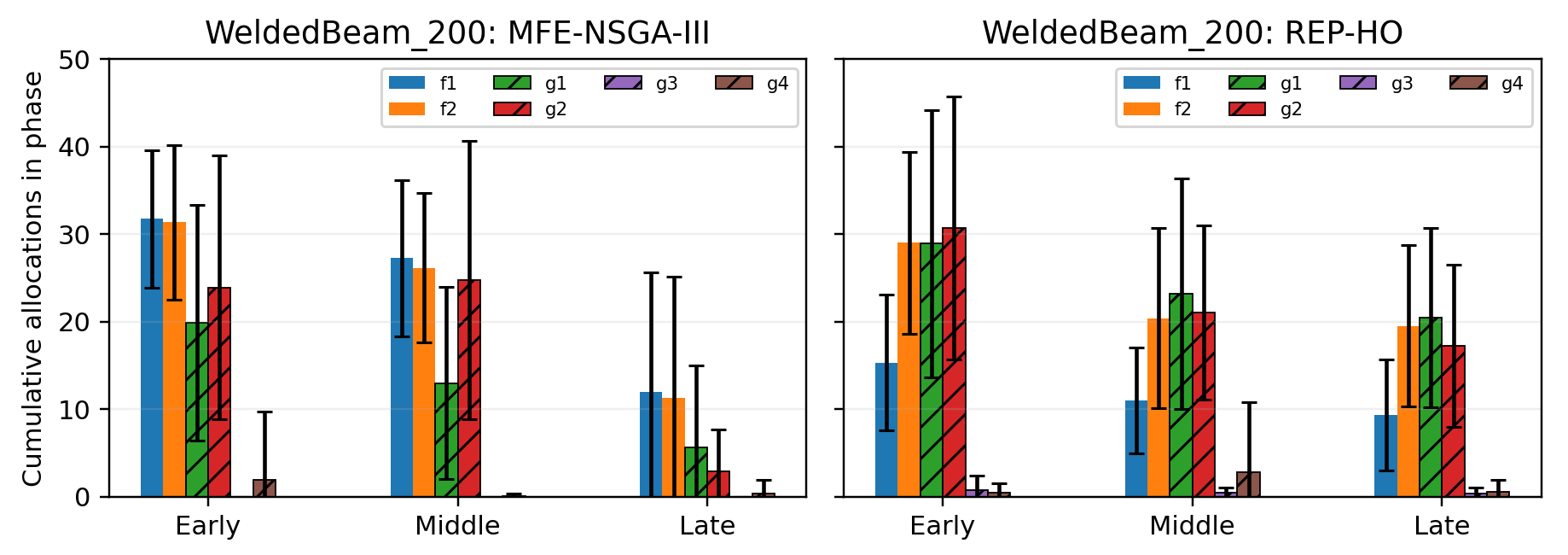}
    \end{minipage}
    \hfill
    \begin{minipage}{0.495\textwidth}
        \centering
        \includegraphics[width=\linewidth]{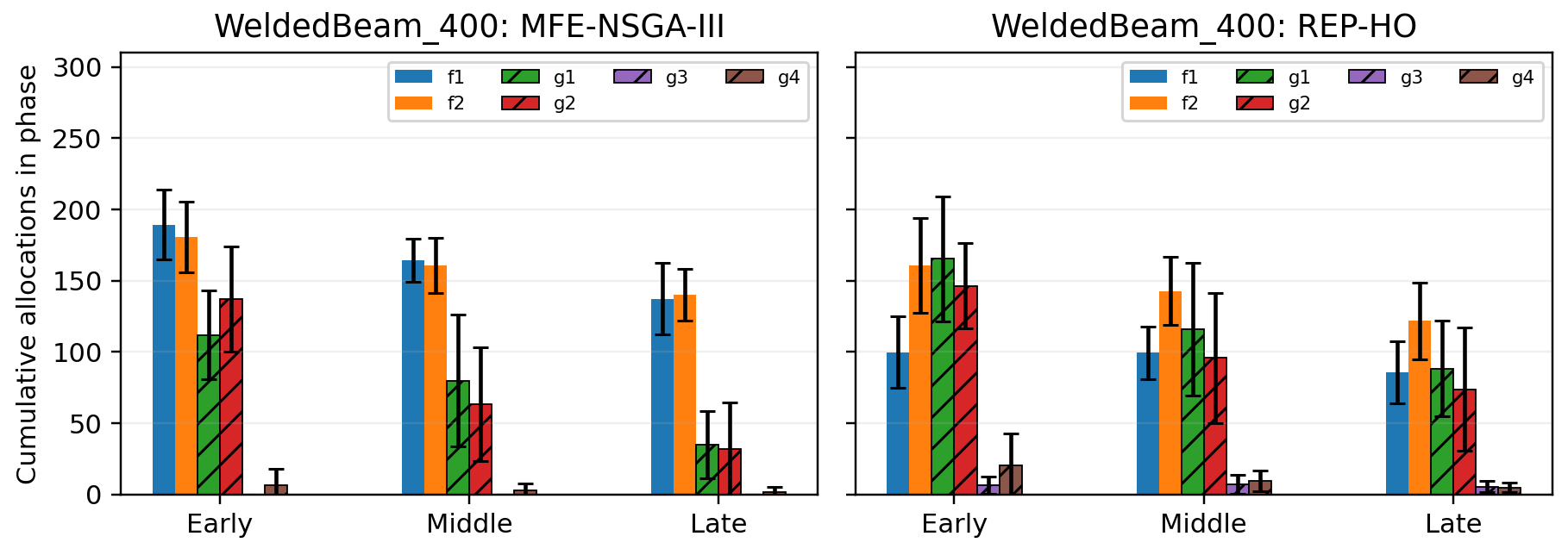}
    \end{minipage}
    \caption{Cumulative allocations on objectives and constraints, the first row is TNK, the second row is MW3, and third row is WeldedBeam problem; the left images are with 200 time unit budget and the right images are with 400 time unit.}
    \label{fig:regime4-cumu-alloc}
\end{figure}
\subsection{Regime 5: Objective Conflict}
\label{app:regime-objective-conflict}

This regime varies the conflict structure among objectives. The unconstrained
main comparison uses ZDT1 as a low-conflict representative and Kursawe as a
high-conflict representative. The constrained comparison uses BNH as a
lower-conflict representative and OSY as a higher-conflict representative. The
default uniform-cost settings are:
\[
\begin{aligned}
    \text{ZDT1} &: (0.50,0.50),\\
    \text{Kursawe} &: (0.50,0.50),\\
    \text{BNH} &: (0.25,0.25 \mid 0.25,0.25),\\
    \text{OSY} &: (0.125,0.125 \mid
    0.125,0.125,0.125,0.125,0.125,0.125).
\end{aligned}
\]
Additional cost-skewed experiments are:
\[
\begin{aligned}
    \text{ZDT1} &: (0.90,0.10),\quad (0.10,0.90),\\
    \text{Kursawe} &: (0.90,0.10),\quad (0.10,0.90),\\
    \text{OSY objective-heavy} &: (0.35,0.35 \mid
    0.05,0.05,0.05,0.05,0.05,0.05),\\
    \text{BNH objective-heavy} &: (0.45,0.45 \mid 0.10,0.10),\\
    \text{OSY constraint-heavy} &: (0.05,0.05 \mid
    0.15,0.15,0.15,0.15,0.15,0.15),\\
    \text{BNH constraint-heavy} &: (0.05,0.05 \mid 0.45,0.45).
\end{aligned}
\]
Objective conflict is characterized using the geometry of the Pareto front.
This is the criterion used for this regime, rather than Spearman correlation or
rank disagreement between objectives because they are often insufficient in identifying the complex conflicting nature of the objectives. The front is first sorted by
\(f_1\) and normalized objective-wise. Let the normalized front points be
\(\mathbf{p}_1,\ldots,\mathbf{p}_L\). We compute the mean second-difference
norm,
\begin{equation}
    C_2 =
    \frac{1}{L-2}
    \sum_{i=2}^{L-1}
    \left\|
    \mathbf{p}_{i+1} - 2\mathbf{p}_i + \mathbf{p}_{i-1}
    \right\|_2,
    \label{eq:appendix-front-second-diff}
\end{equation}
and the mean turning angle between consecutive front segments,
\begin{equation}
    C_{\theta}
    =
    \frac{1}{L-2}
    \sum_{i=2}^{L-1}
    \arccos
    \left(
    \frac{
    (\mathbf{p}_i-\mathbf{p}_{i-1})^\top
    (\mathbf{p}_{i+1}-\mathbf{p}_i)}
    {\|\mathbf{p}_i-\mathbf{p}_{i-1}\|_2
    \|\mathbf{p}_{i+1}-\mathbf{p}_i\|_2}
    \right).
    \label{eq:appendix-front-turning-angle}
\end{equation}
Larger values indicate a more nonlinear and conflict-sensitive front, which
requires finer discrimination between objective trade-offs.

\begin{table}[t]
    \centering
    \caption{Objective-conflict diagnostics for Regime 5. Larger front
    curvature and turning angle indicate stronger conflict/nonlinearity in the
    trade-off structure.}
    \label{tab:regime5-conflict-diagnostics}
    \footnotesize
    \begin{tabular}{lcc}
        \toprule
        Problem & Mean 2nd diff & Mean turning angle (deg) \\
        \midrule
        ZDT1 & 0.000011 & 0.031107 \\
        Kursawe & 0.007836 & 4.232093 \\
        OSY & 0.005388 & 4.605905 \\
        BNH & 0.000392 & 0.912159 \\
        \bottomrule
    \end{tabular}
\end{table}

Based on Table~\ref{tab:regime5-conflict-diagnostics}, ZDT1 and BNH are used
as lower-conflict cases because their Pareto fronts are close to smooth and
weakly curved. ZDT1 has an almost zero mean second-difference value
(\(0.000011\)) and a very small mean turning angle (\(0.031107^\circ\)),
indicating a nearly linear local front geometry. BNH is also relatively mild,
with a mean second difference of \(0.000392\) and a mean turning angle of
\(0.912159^\circ\). In contrast, Kursawe and OSY show much stronger front
nonlinearity. Kursawe has a mean second difference of \(0.007836\) and a mean
turning angle of \(4.232093^\circ\), while OSY has a mean second difference of
\(0.005388\) and the largest mean turning angle, \(4.605905^\circ\). We
therefore treat Kursawe and OSY as higher-conflict cases.

\subsubsection{Additional Results}
\begin{table*}[h]
    \centering
    \caption{Regime 5: Objective Conflict IGD+ performance summary (IGD+ $\downarrow$).}
    \label{tab:regime5-performance-igd}
    \scriptsize
    \setlength{\tabcolsep}{3pt}
    \resizebox{\textwidth}{!}{%
    \begin{tabular}{p{0.10\textwidth} p{0.13\textwidth} p{0.12\textwidth} p{0.12\textwidth} p{0.12\textwidth} p{0.13\textwidth} p{0.15\textwidth}}
        \toprule
        Problem & Setting & NSGA-III & SA-NSGA-III & MFE-NSGA-III & REP-HO(Qwen) & REP-HO(Deepseek) \\
        \midrule
        BNH & constraint-heavy & 0.0098 $\pm$ 0.0019 & 0.0059 $\pm$ 0.0001 & \textbf{0.0058 $\pm$ 0.0000} & 0.0058 $\pm$ 0.0001 & 0.0058 $\pm$ 0.0001 \\
        BNH & uniform & 0.0098 $\pm$ 0.0019 & 0.0059 $\pm$ 0.0001 & \textbf{0.0057 $\pm$ 0.0000} & 0.0058 $\pm$ 0.0000 & 0.0058 $\pm$ 0.0001 \\
        BNH & objective-heavy & 0.0098 $\pm$ 0.0019 & 0.0059 $\pm$ 0.0001 & 0.0058 $\pm$ 0.0001 & 0.0058 $\pm$ 0.0001 & \textbf{0.0058 $\pm$ 0.0000} \\
        Kursawe & $f_2$-heavy & \textbf{0.1804 $\pm$ 0.0453} & 0.3471 $\pm$ 0.2232 & 0.2792 $\pm$ 0.1203 & 0.2300 $\pm$ 0.0864 & 0.2204 $\pm$ 0.0883 \\
        Kursawe & uniform & \textbf{0.1804 $\pm$ 0.0453} & 0.3471 $\pm$ 0.2232 & 0.2622 $\pm$ 0.1750 & 0.2975 $\pm$ 0.2353 & 0.3121 $\pm$ 0.3094 \\
        Kursawe & $f_1$-heavy & \textbf{0.1804 $\pm$ 0.0453} & 0.3471 $\pm$ 0.2232 & 0.2292 $\pm$ 0.0967 & 0.2631 $\pm$ 0.1310 & 0.2827 $\pm$ 0.0846 \\
        OSY & constraint-heavy & 0.5655 $\pm$ 0.1651 & 0.0659 $\pm$ 0.0787 & 0.0305 $\pm$ 0.0849 & 0.0304 $\pm$ 0.0213 & \textbf{0.0246 $\pm$ 0.0168} \\
        OSY & uniform & 0.5655 $\pm$ 0.1651 & 0.0659 $\pm$ 0.0787 & 0.0334 $\pm$ 0.0845 & 0.0322 $\pm$ 0.0193 & \textbf{0.0259 $\pm$ 0.0172} \\
        OSY & objective-heavy & 0.5655 $\pm$ 0.1651 & 0.0659 $\pm$ 0.0787 & 0.0468 $\pm$ 0.0824 & 0.0332 $\pm$ 0.0179 & \textbf{0.0324 $\pm$ 0.0170} \\
        ZDT1 & $f_2$-heavy & 1.8220 $\pm$ 0.1522 & 0.0736 $\pm$ 0.0494 & \textbf{0.0387 $\pm$ 0.0506} & 0.0419 $\pm$ 0.0463 & 0.0390 $\pm$ 0.0538 \\
        ZDT1 & uniform & 1.8220 $\pm$ 0.1522 & 0.0736 $\pm$ 0.0494 & \textbf{0.0355 $\pm$ 0.0438} & 0.0357 $\pm$ 0.0311 & 0.0360 $\pm$ 0.0538 \\
        ZDT1 & $f_1$-heavy & 1.8220 $\pm$ 0.1522 & 0.0736 $\pm$ 0.0494 & \textbf{0.0203 $\pm$ 0.0090} & 0.0419 $\pm$ 0.0588 & 0.0248 $\pm$ 0.0326 \\
        \midrule
        \multicolumn{2}{l}{Mean ranking integrated with IGD+ in Table \ref{tab:regime5-performance}} & 4.042 & 4.229 & 2.062 & 2.708 & \textbf{1.958} \\
        \bottomrule
    \end{tabular}%
    }
\end{table*}

\section{Budget-matched random allocation}
\label{app:random-allocation}
The random allocator uses the same candidate set and the same generation-level high-fidelity budget as REP-HO, but selects unevaluated candidate--objective and candidate--constraint pairs uniformly at random until that budget is exhausted. Table~\ref{tab:random-allocation} reports its Regime~1 results under the same HV and IGD+ definitions as the main tables. Ranking scores below are recomputed on these 12 settings with the random allocator included, so they are not the five-method scores in Table~\ref{tab:regime1-performance}. Lower is better. Both REP-HO variants and MFE-NSGA-III remain ahead of random allocation, while random allocation itself remains ahead of NSGA-III and SA-NSGA-III.

\begin{table*}[h]
    \centering
    \caption{Regime 1 budget-matched random allocation (HV $\uparrow$, IGD+ $\downarrow$).}
    \label{tab:random-allocation}
    \scriptsize
    \setlength{\tabcolsep}{4pt}
    \begin{tabular}{llcc}
        \toprule
        Problem & Setting & HV & IGD+ \\
        \midrule
        Carside & uniform & 8.5781 $\pm$ 0.0615 & 0.0018 $\pm$ 0.0016 \\
        Carside & constraint-heavy & 8.6521 $\pm$ 0.0523 & 0.0012 $\pm$ 0.0008 \\
        Carside & objective-heavy & 8.6366 $\pm$ 0.0483 & 0.0009 $\pm$ 0.0008 \\
        MW3 & constraint-heavy & 3.1401 $\pm$ 0.2353 & 0.0579 $\pm$ 0.0516 \\
        MW3 & uniform & 3.1514 $\pm$ 0.1915 & 0.0510 $\pm$ 0.0360 \\
        MW3 & objective-heavy & 3.1327 $\pm$ 0.2020 & 0.0521 $\pm$ 0.0338 \\
        TNK & constraint-heavy & 3.2528 $\pm$ 0.0882 & 0.0125 $\pm$ 0.0142 \\
        TNK & uniform & 3.2370 $\pm$ 0.0792 & 0.0167 $\pm$ 0.0145 \\
        TNK & objective-heavy & 3.1807 $\pm$ 0.2292 & 0.0288 $\pm$ 0.0529 \\
        WeldedBeam & constraint-heavy & 3.2279 $\pm$ 1.1061 & 0.1199 $\pm$ 0.1100 \\
        WeldedBeam & uniform & 3.4880 $\pm$ 0.3572 & 0.1514 $\pm$ 0.1210 \\
        WeldedBeam & objective-heavy & 3.6152 $\pm$ 0.2149 & 0.0997 $\pm$ 0.0622 \\
        \bottomrule
    \end{tabular}
\end{table*}

\begin{table}[h]
    \centering
    \caption{Regime 1 mean ranking score with the random allocator included. Lower is better.}
    \label{tab:random-allocation-rank}
    \begin{tabular}{lc}
        \toprule
        Method & Mean ranking score \\
        \midrule
        REP-HO(Deepseek) & \textbf{2.250} \\
        REP-HO(Qwen) & 2.417 \\
        MFE-NSGA-III & 2.521 \\
        Random & 3.250 \\
        SA-NSGA-III & 5.167 \\
        NSGA-III & 5.396 \\
        \bottomrule
    \end{tabular}
\end{table}

\section{LLM call cost}
\label{app:llm-cost}
Table~\ref{tab:llm-cost} summarizes every completed agg run of REP-HO(Qwen) and REP-HO(Deepseek) across the five regimes (41 settings, 20 seeds each). Dollar cost is not stored in the logs. It is estimated from the recorded token counts using published list prices: DeepSeek-V4-Flash at the official off-peak cache-miss rate of \$0.15 / \$0.60 per million input/output tokens, and Qwen3.5-35B-A3B at \$0.14 / \$1.00 per million input/output tokens. 

\begin{table}[h]
    \centering
    \caption{Per-run LLM usage, mean $\pm$ standard deviation.}
    \label{tab:llm-cost}
    \scriptsize
    \setlength{\tabcolsep}{3pt}
    \resizebox{\textwidth}{!}{%
    \begin{tabular}{lccccc}
        \toprule
        Backbone & Calls & Input tokens & Output tokens & Est.\ cost (USD) & Wall-clock (s) \\
        \midrule
        REP-HO(Qwen) & 52.8 $\pm$ 30.6 & 550662 $\pm$ 299085 & 24592 $\pm$ 12694 & 0.102 $\pm$ 0.054 & 293.0 $\pm$ 211.0 \\
        REP-HO(Deepseek) & 56.3 $\pm$ 31.0 & 513342 $\pm$ 310864 & 19768 $\pm$ 10458 & 0.089 $\pm$ 0.052 & 202.7 $\pm$ 139.0 \\
        \bottomrule
    \end{tabular}%
    }
\end{table}

\section{Pareto Fronts}
\label{app:pareto-front}
\begin{figure}[H]
    \centering
    \begin{minipage}{0.33\textwidth}
        \centering
        \includegraphics[width=\linewidth]{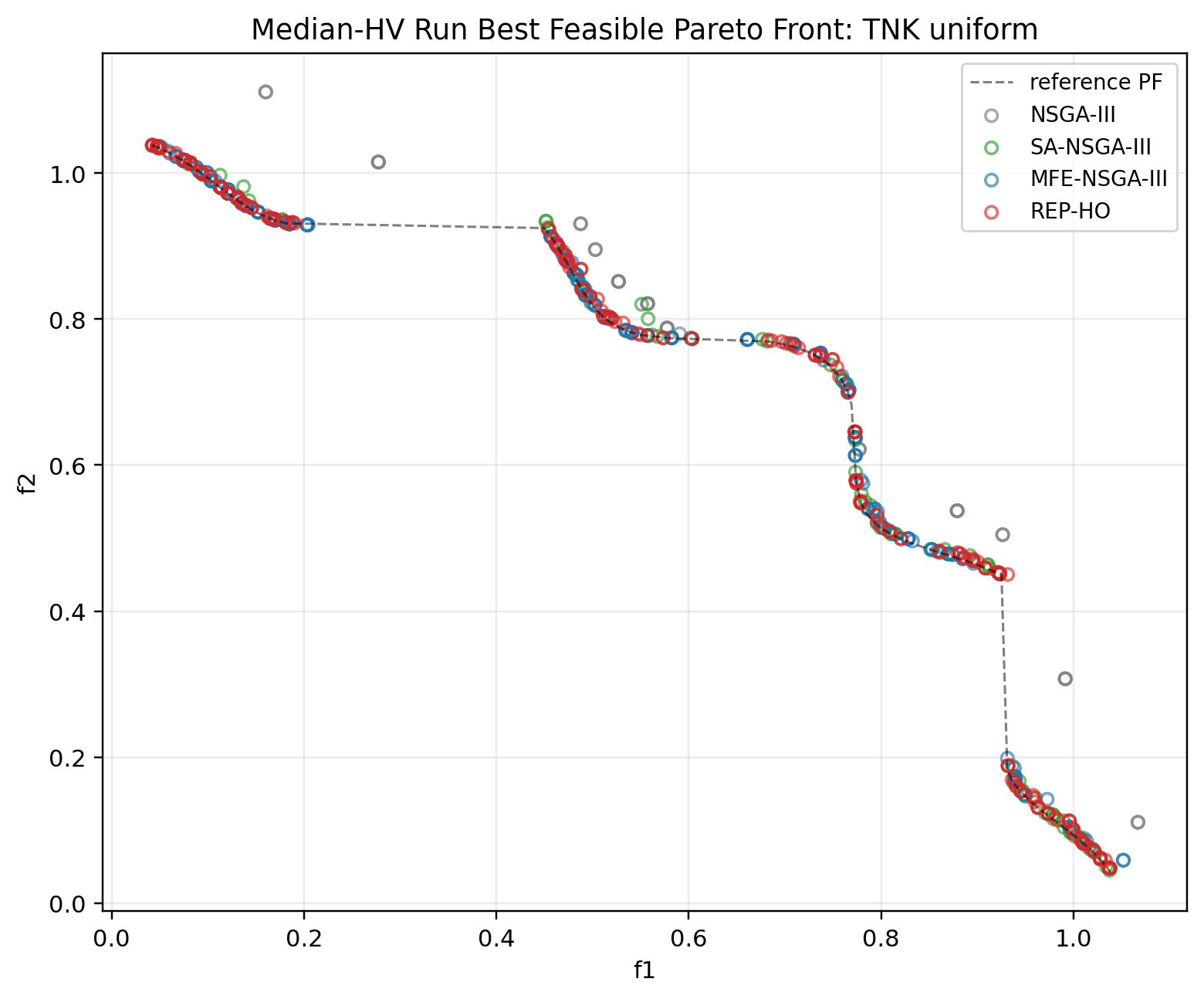}
    \end{minipage}\hfill
    \begin{minipage}{0.33\textwidth}
        \centering
        \includegraphics[width=\linewidth]{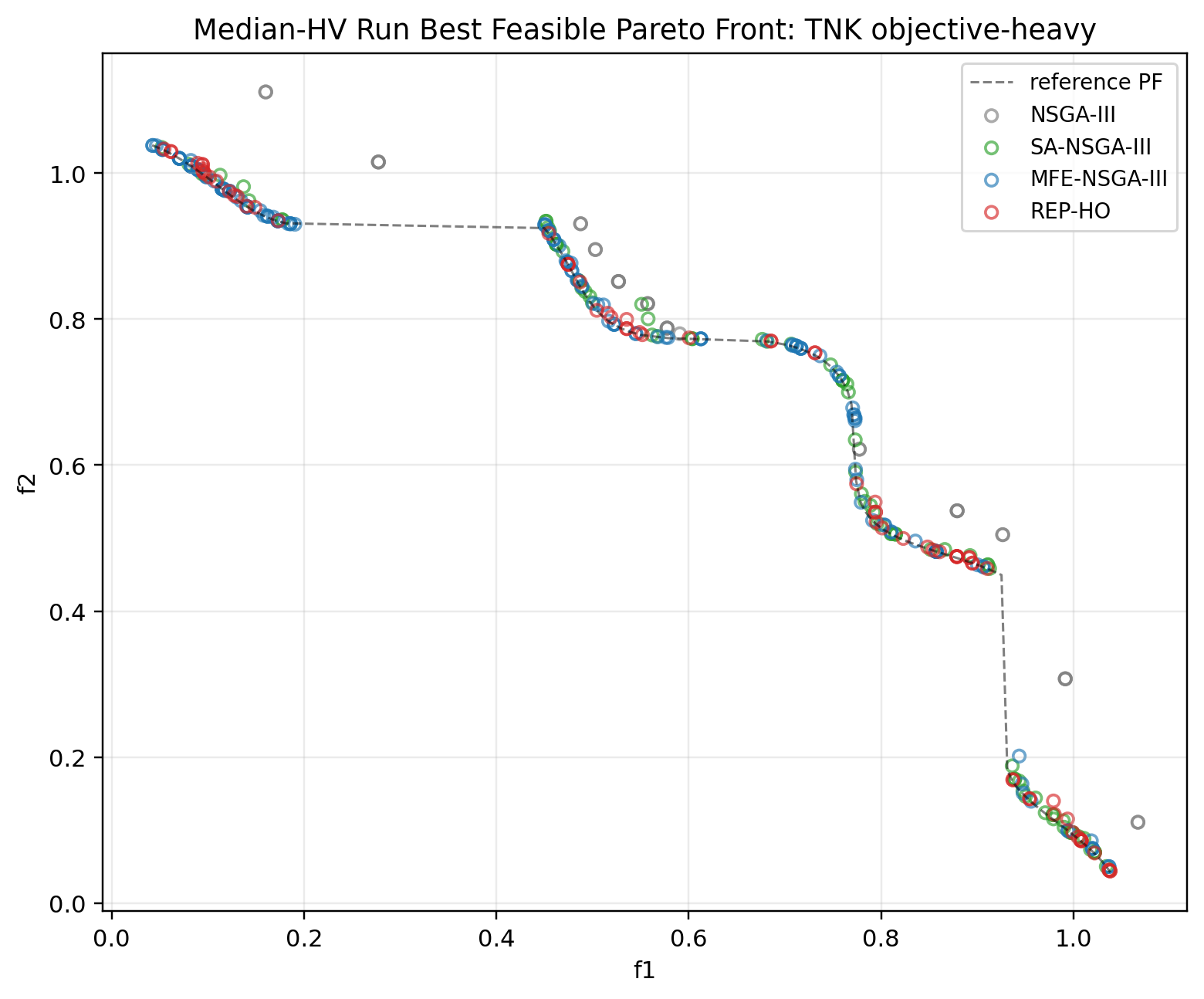}
    \end{minipage}\hfill
    \begin{minipage}{0.33\textwidth}
        \centering
        \includegraphics[width=\linewidth]{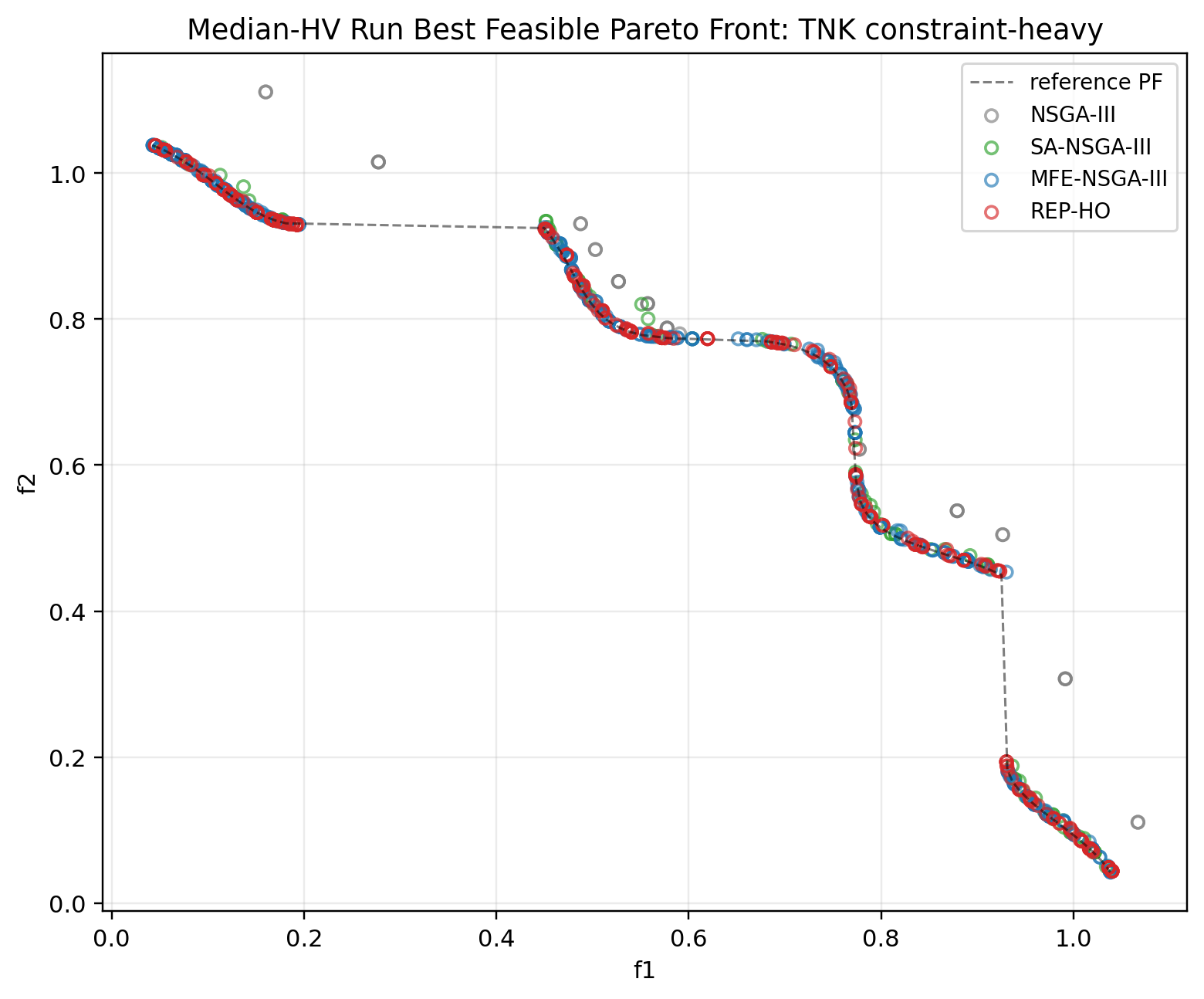}
    \end{minipage}
    \centering
    \begin{minipage}{0.33\textwidth}
        \centering
        \includegraphics[width=\linewidth]{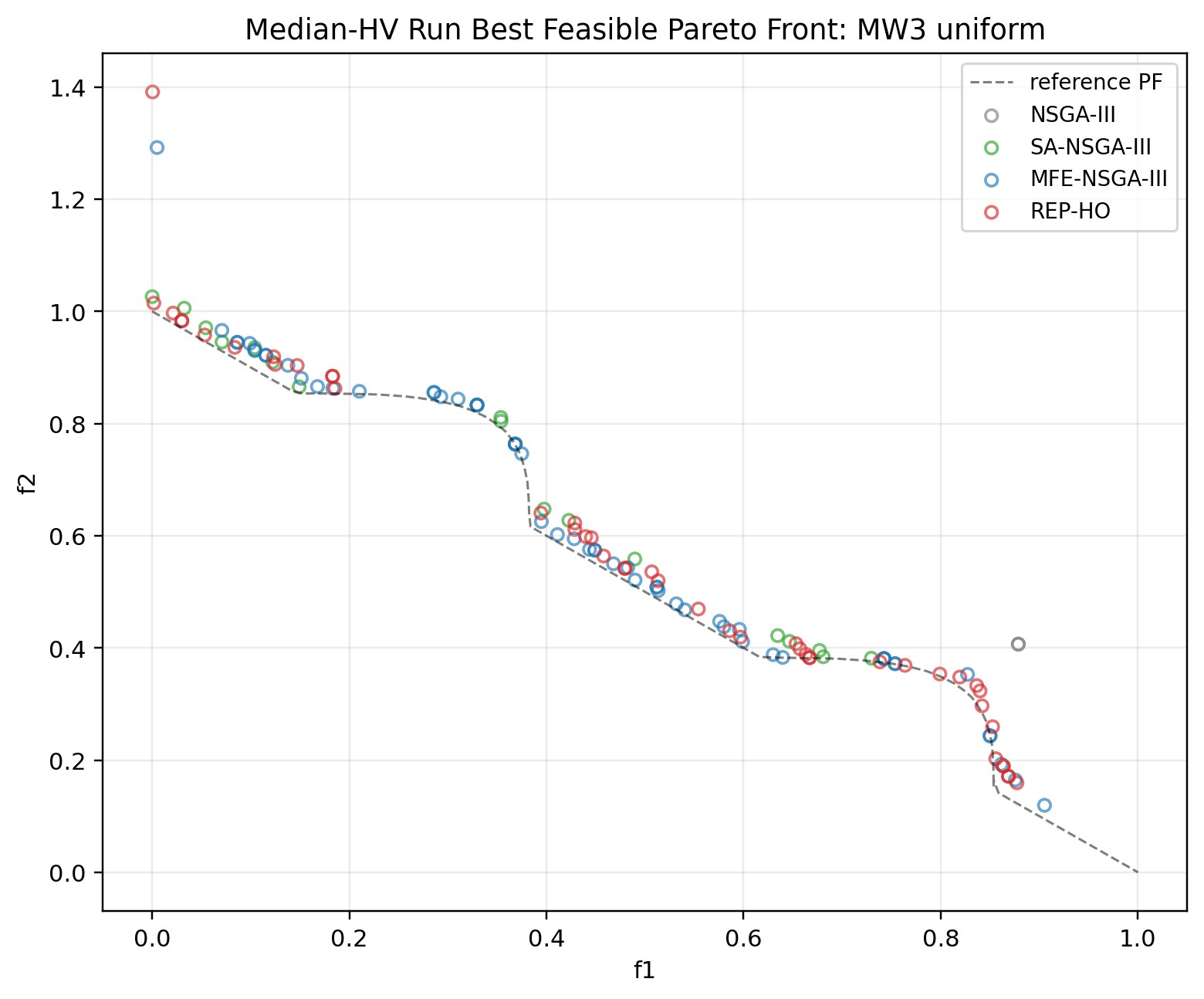}
    \end{minipage}\hfill
    \begin{minipage}{0.33\textwidth}
        \centering
        \includegraphics[width=\linewidth]{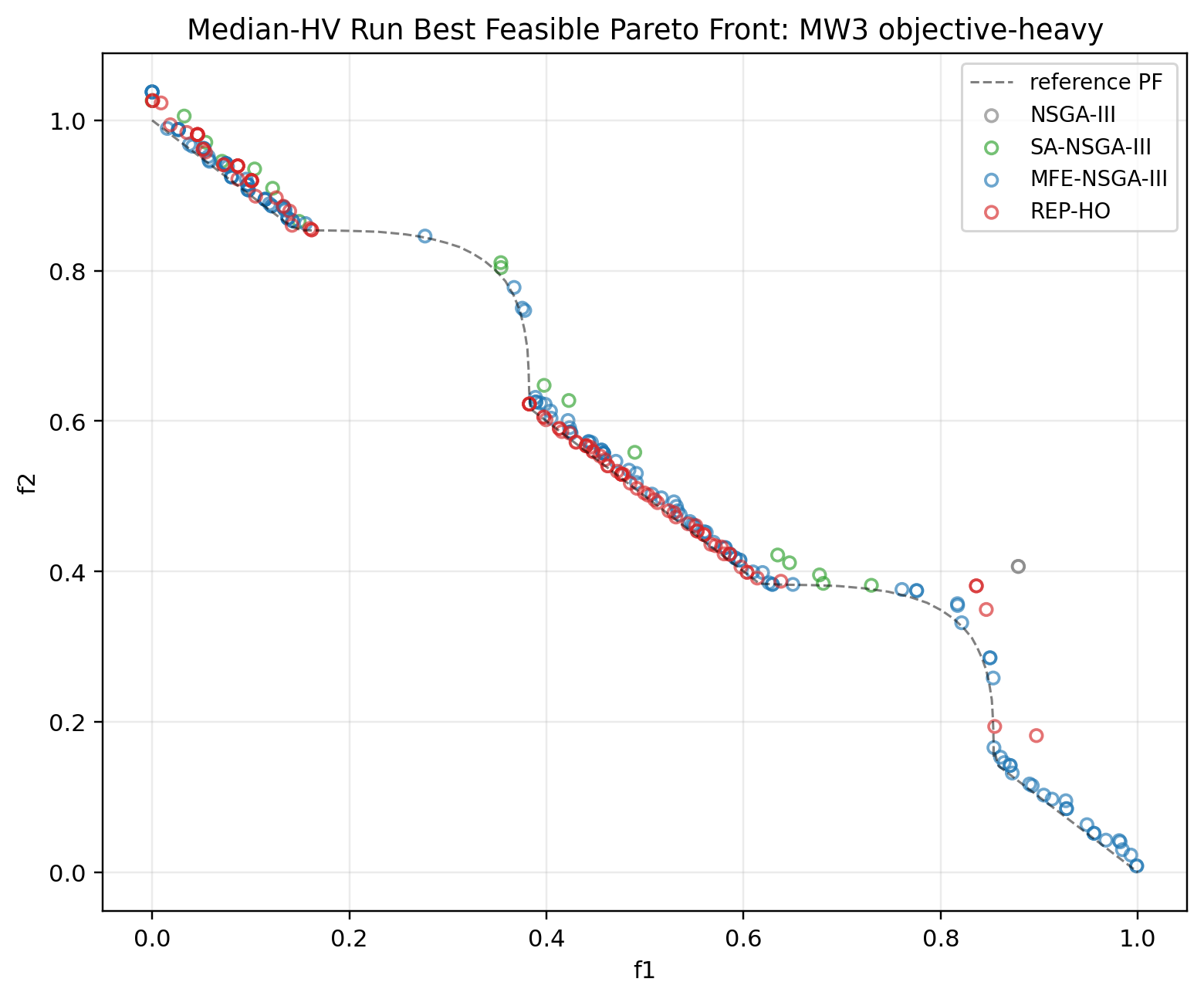}
    \end{minipage}\hfill
    \begin{minipage}{0.33\textwidth}
        \centering
        \includegraphics[width=\linewidth]{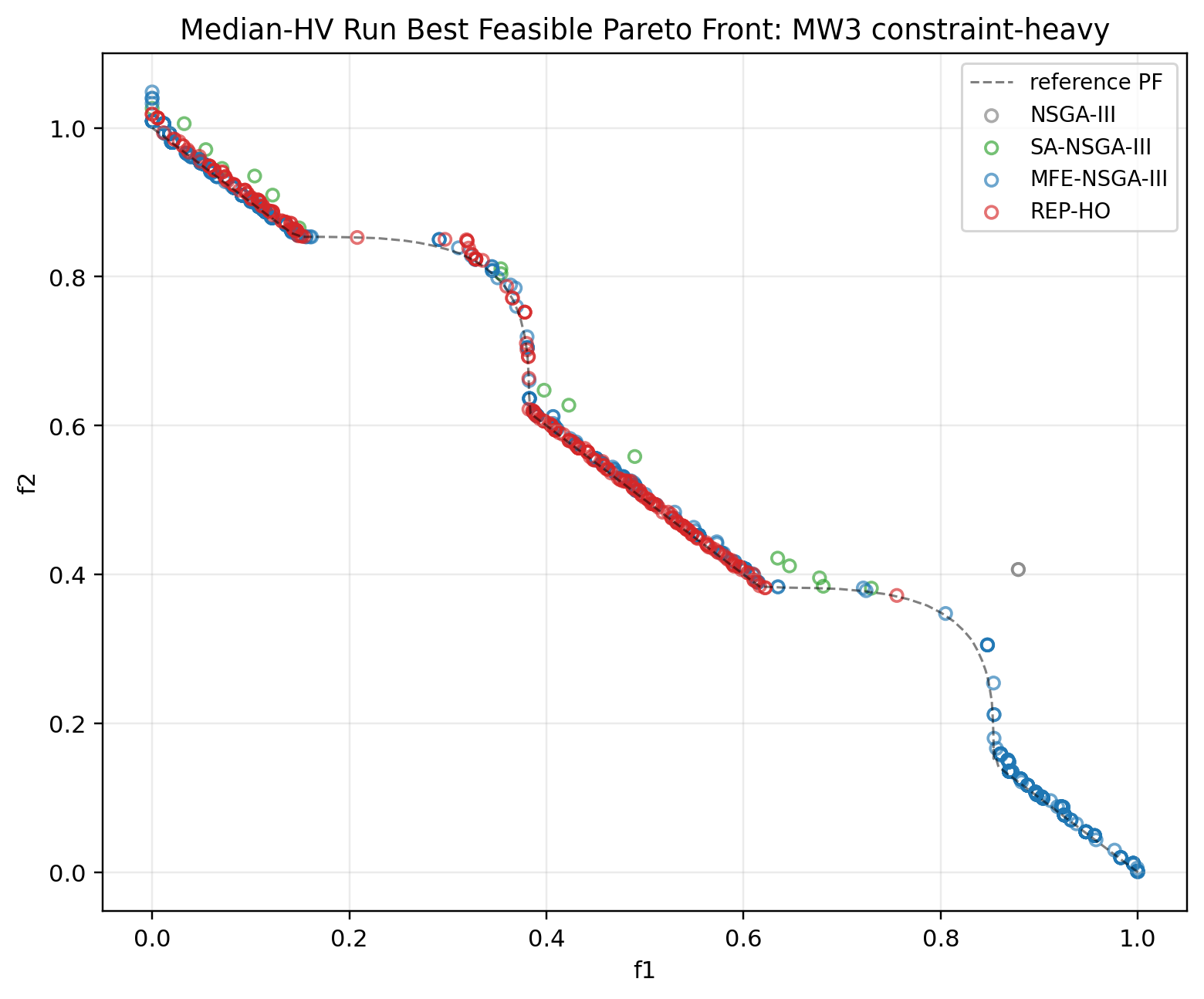}
    \end{minipage}
    \centering
    \begin{minipage}{0.33\textwidth}
        \centering
        \includegraphics[width=\linewidth]{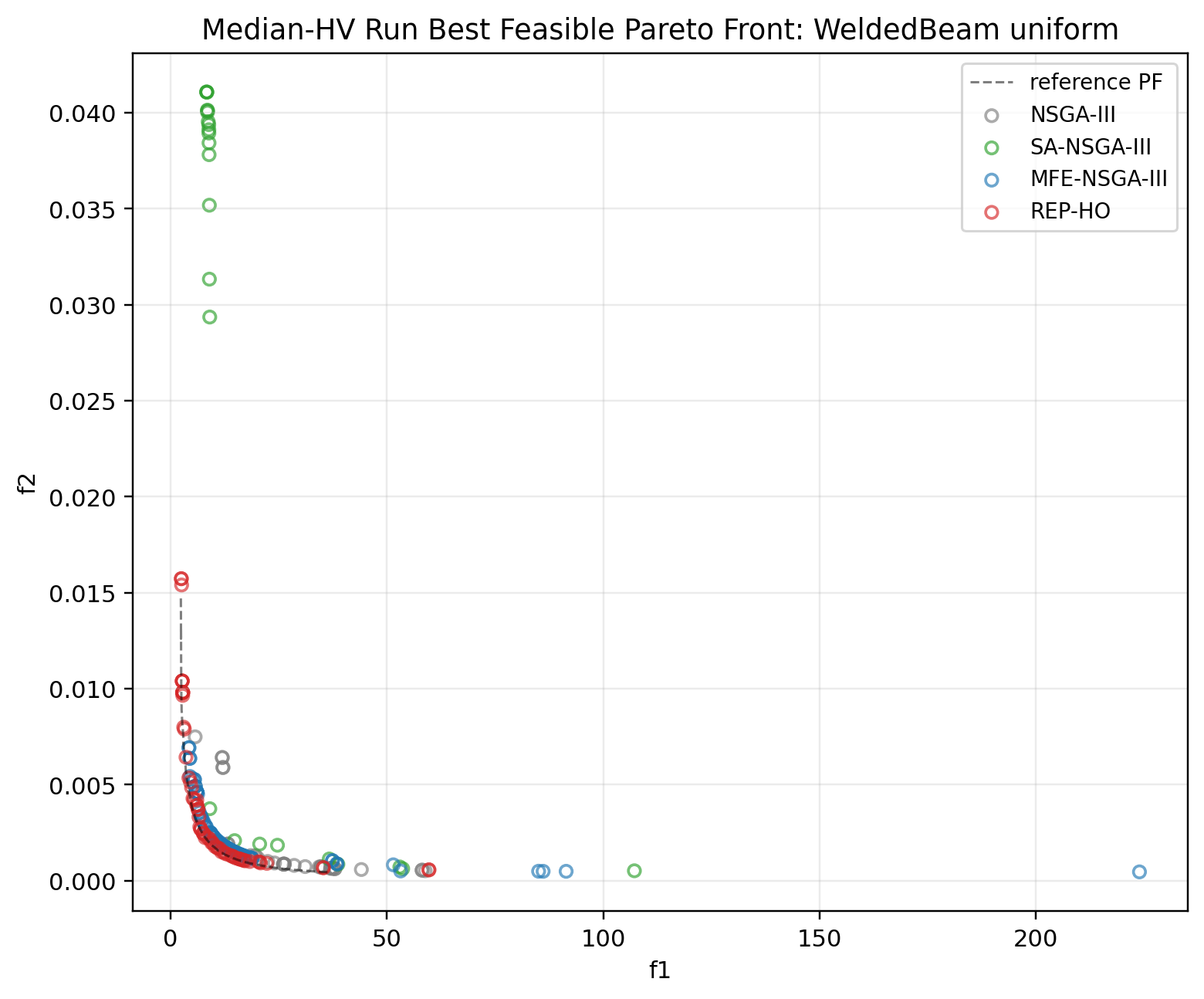}
    \end{minipage}\hfill
    \begin{minipage}{0.33\textwidth}
        \centering
        \includegraphics[width=\linewidth]{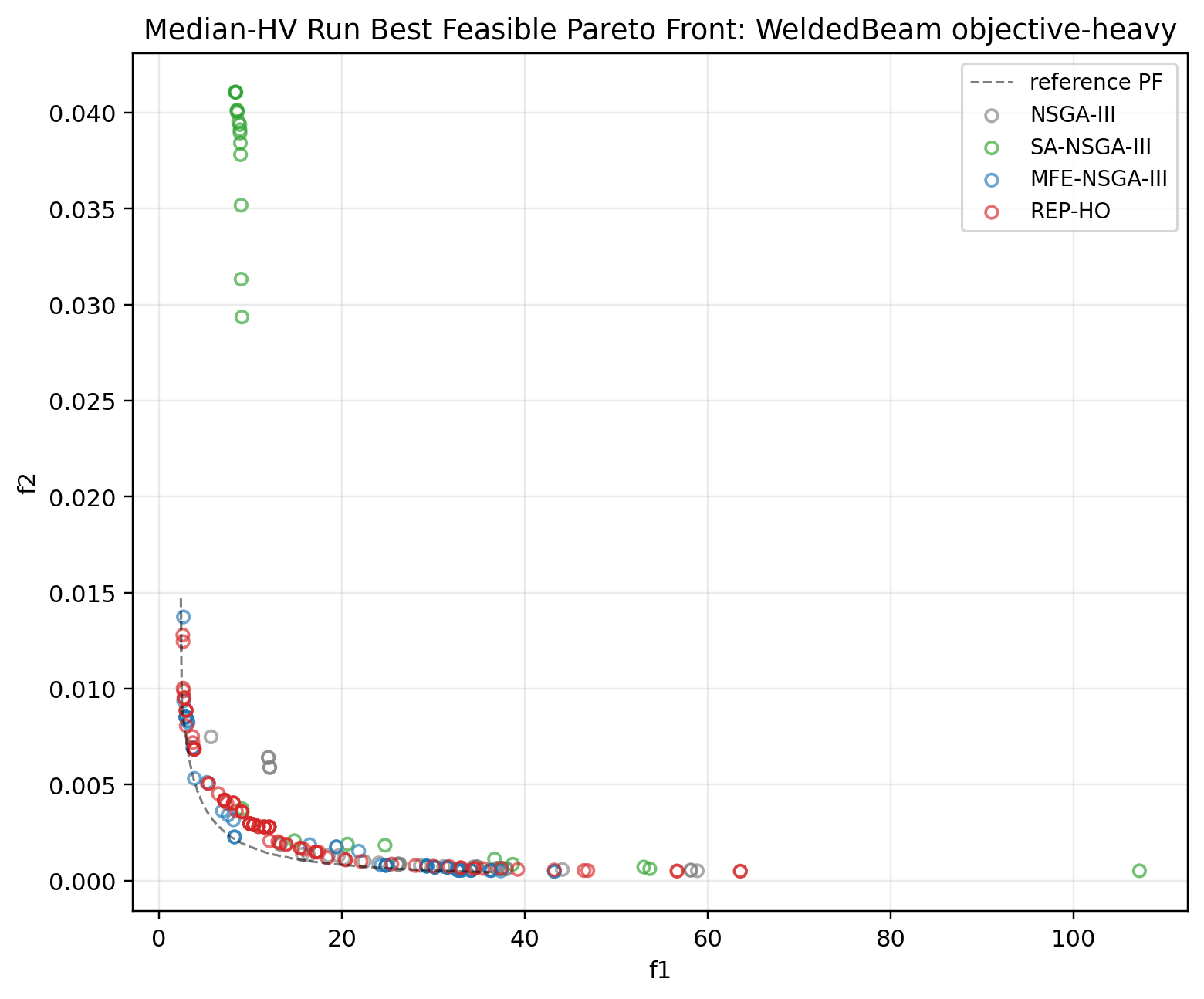}
    \end{minipage}\hfill
    \begin{minipage}{0.33\textwidth}
        \centering
        \includegraphics[width=\linewidth]{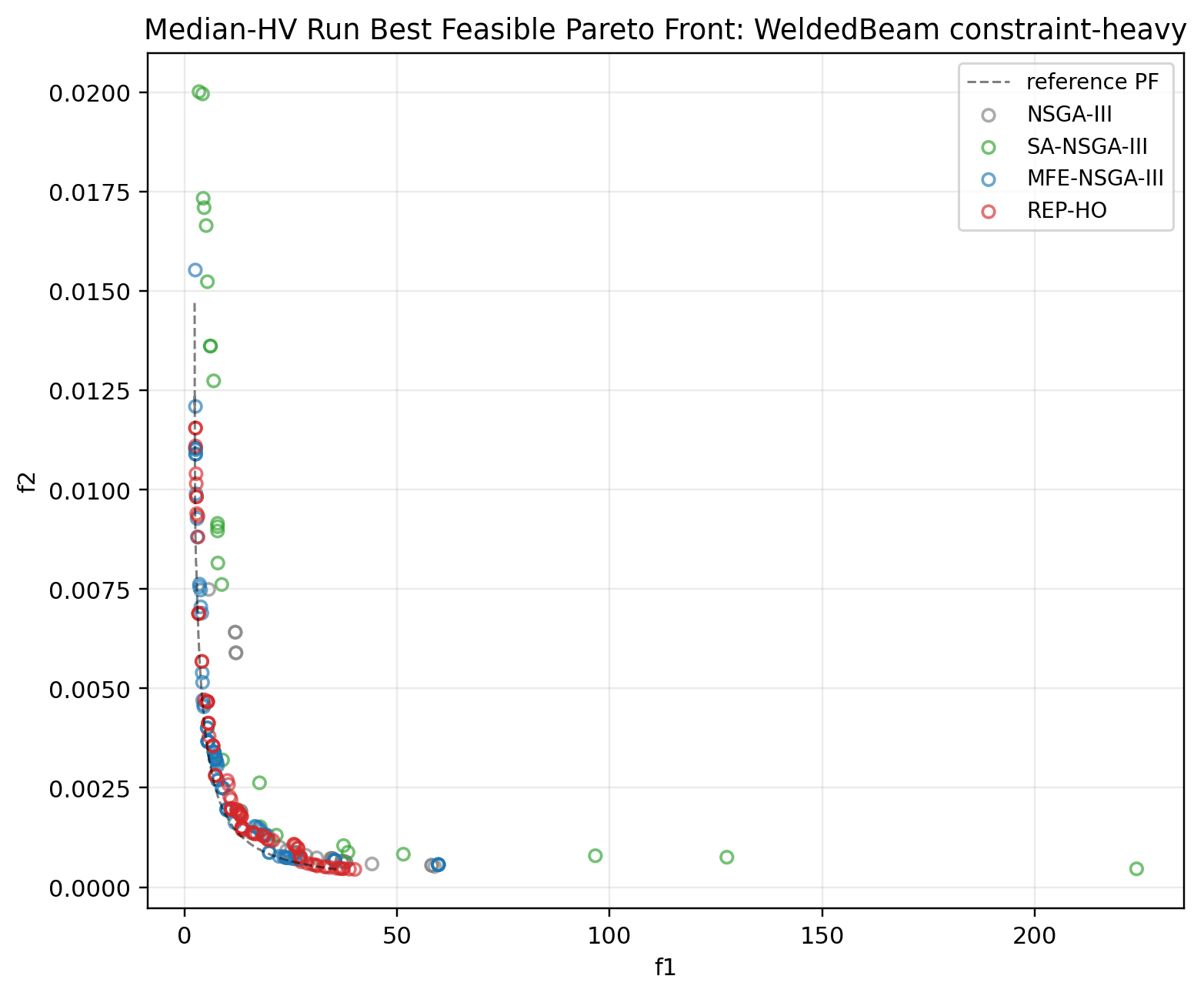}
    \end{minipage}
    \caption{Pareto front of the problems in regime 1.}
    \label{fig:regime1-convergence}
\end{figure}

\section{Broader impacts}
This work studies whether large language models can be used as flexible evaluation planners for heterogeneous optimization problems. A positive societal impact is that such planners may reduce the design complexity and increase the efficiency of expensive optimization tasks in engineering design, scientific discovery, and machine learning system development. By allocating high-fidelity evaluations more selectively, the method may help lower energy consumption but achieve better performance. The negative side is that LLM-based planners may introduce opaque decision behavior, and is hard to comprehensively evaluate all the behaviors of it. 

\section{Convergence}
\label{app:convergence}
\begin{figure}[H]
    \centering
    \begin{minipage}{0.33\textwidth}
        \centering
        \includegraphics[width=\linewidth]{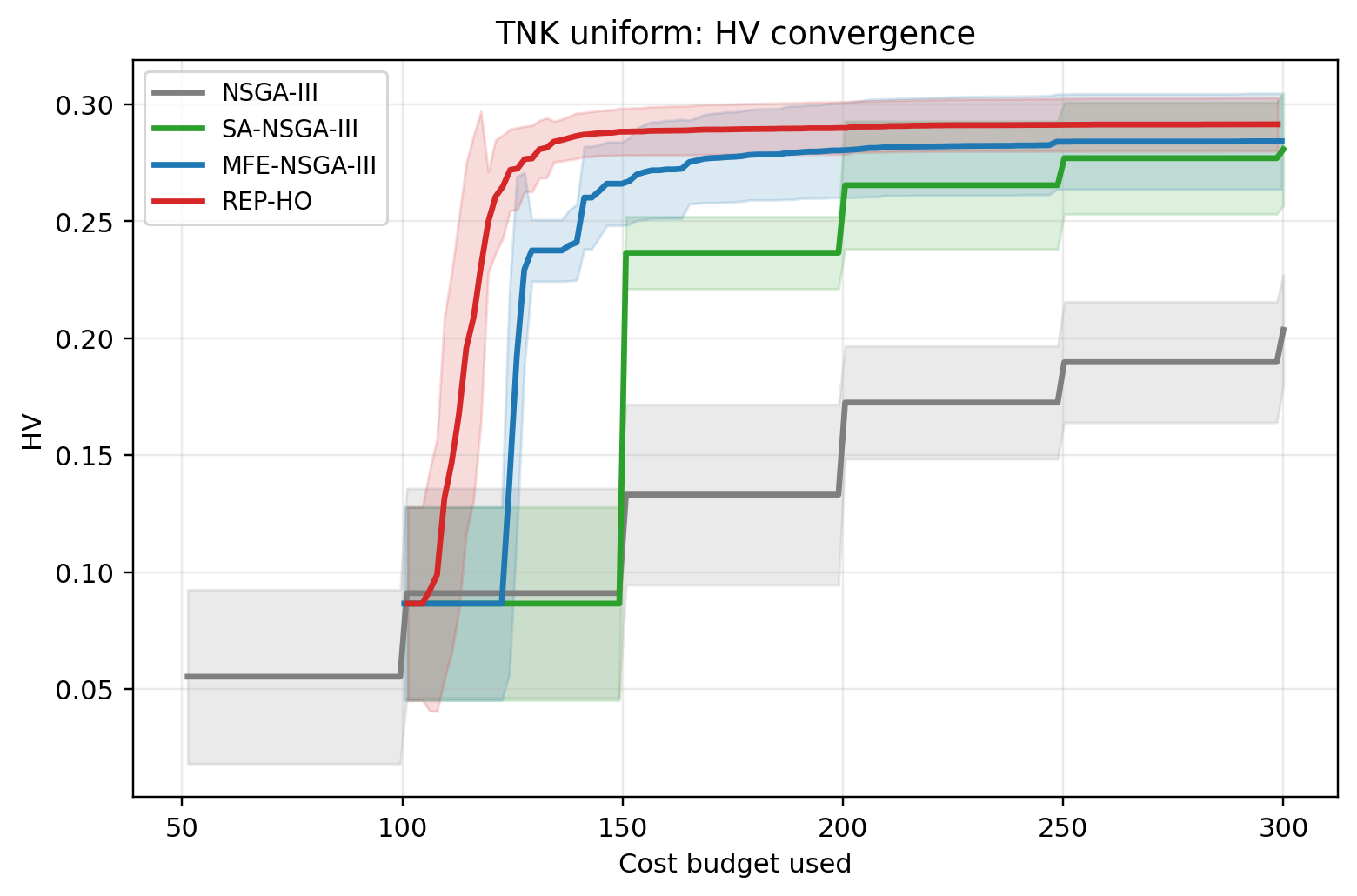}
    \end{minipage}\hfill
    \begin{minipage}{0.33\textwidth}
        \centering
        \includegraphics[width=\linewidth]{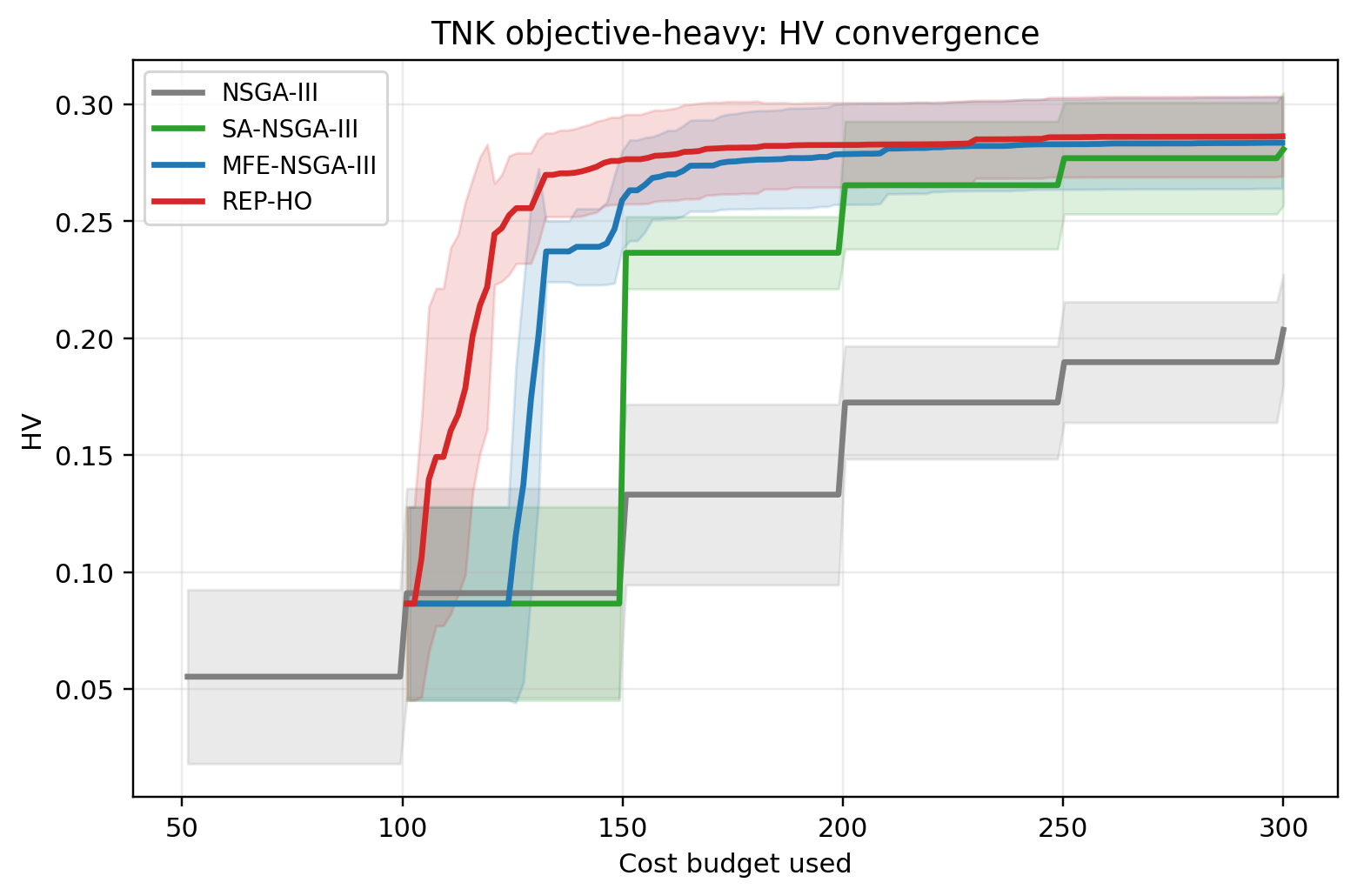}
    \end{minipage}\hfill
    \begin{minipage}{0.33\textwidth}
        \centering
        \includegraphics[width=\linewidth]{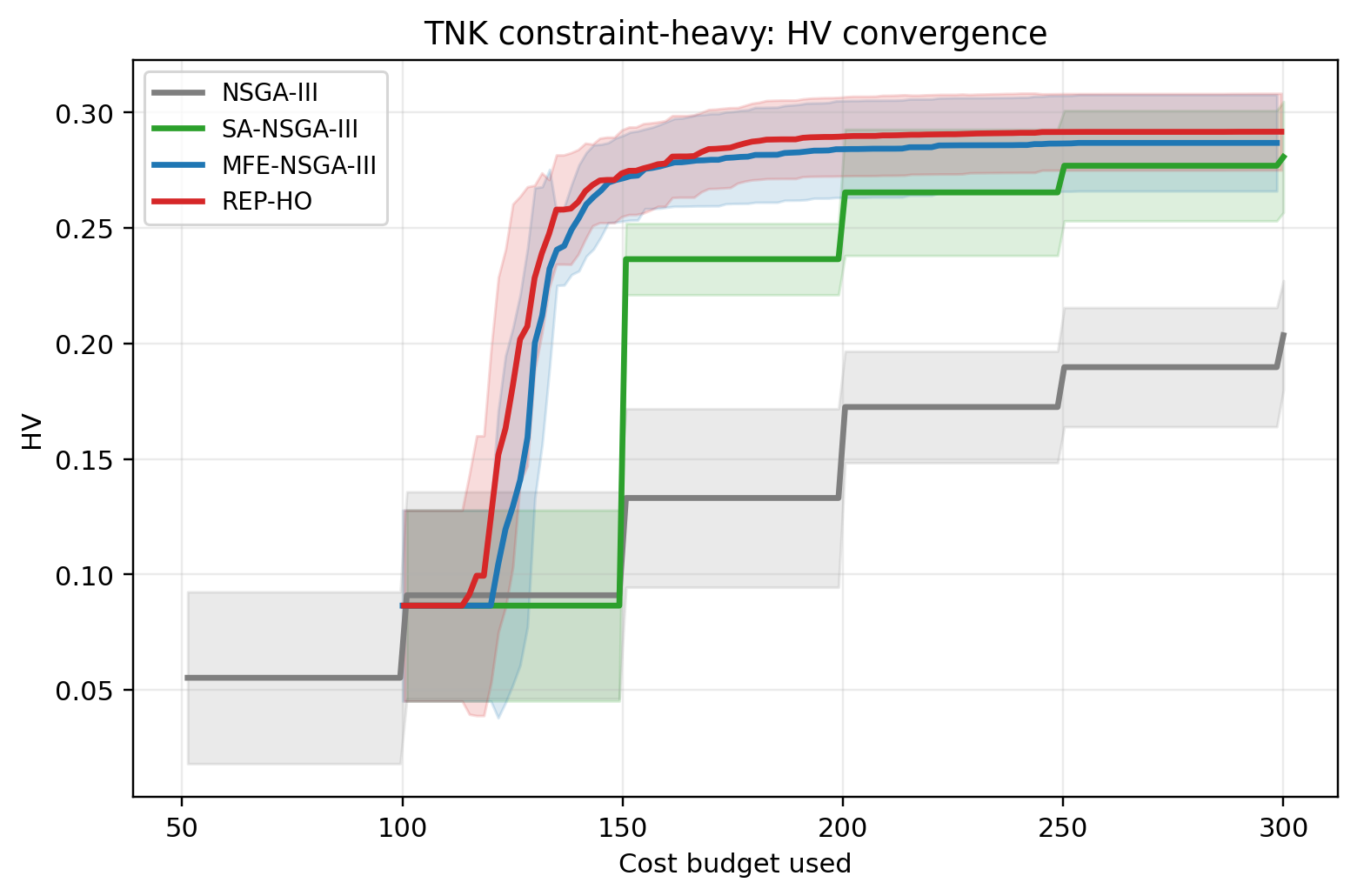}
    \end{minipage}
    \centering
    \begin{minipage}{0.33\textwidth}
        \centering
        \includegraphics[width=\linewidth]{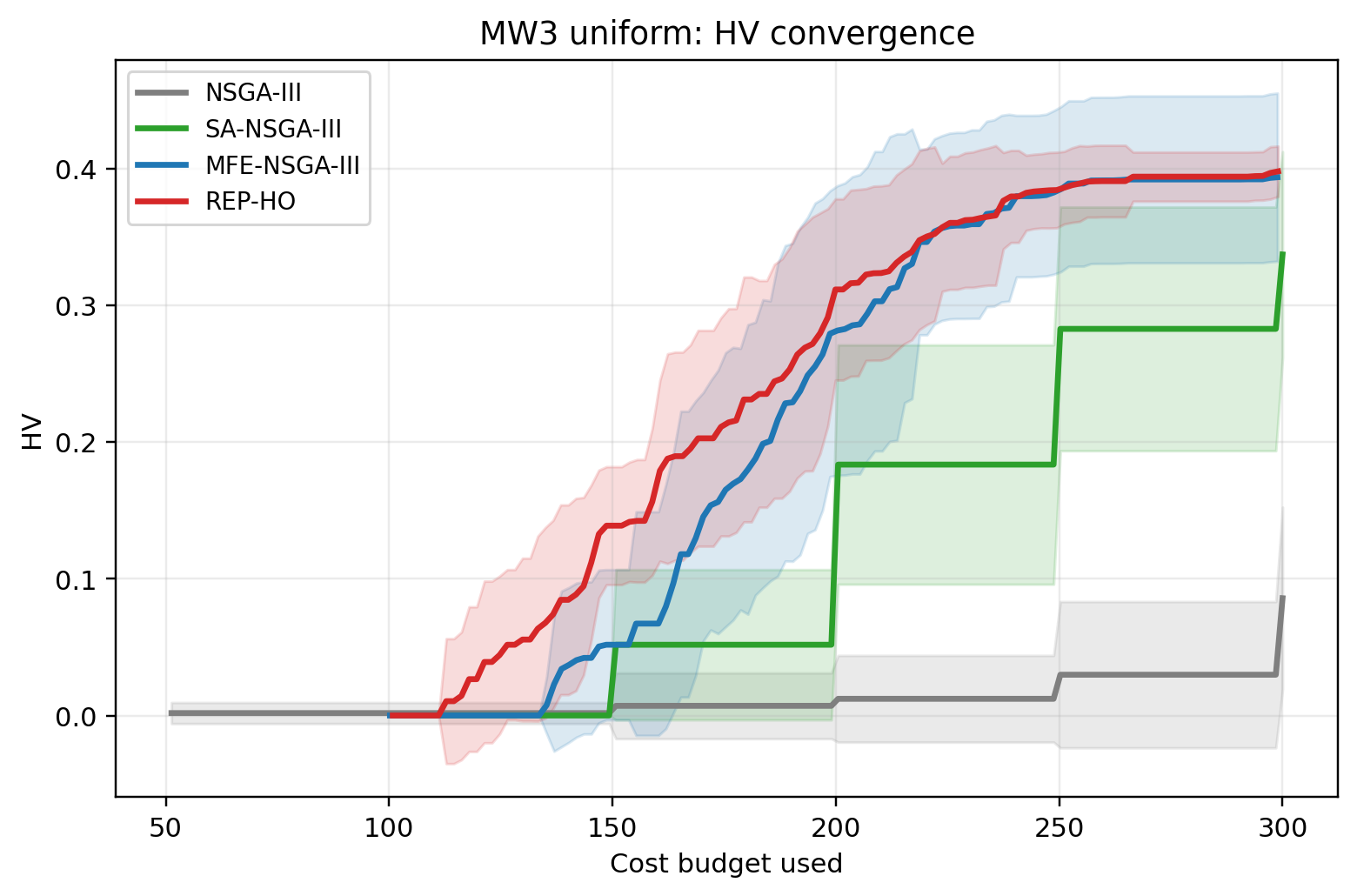}
    \end{minipage}\hfill
    \begin{minipage}{0.33\textwidth}
        \centering
        \includegraphics[width=\linewidth]{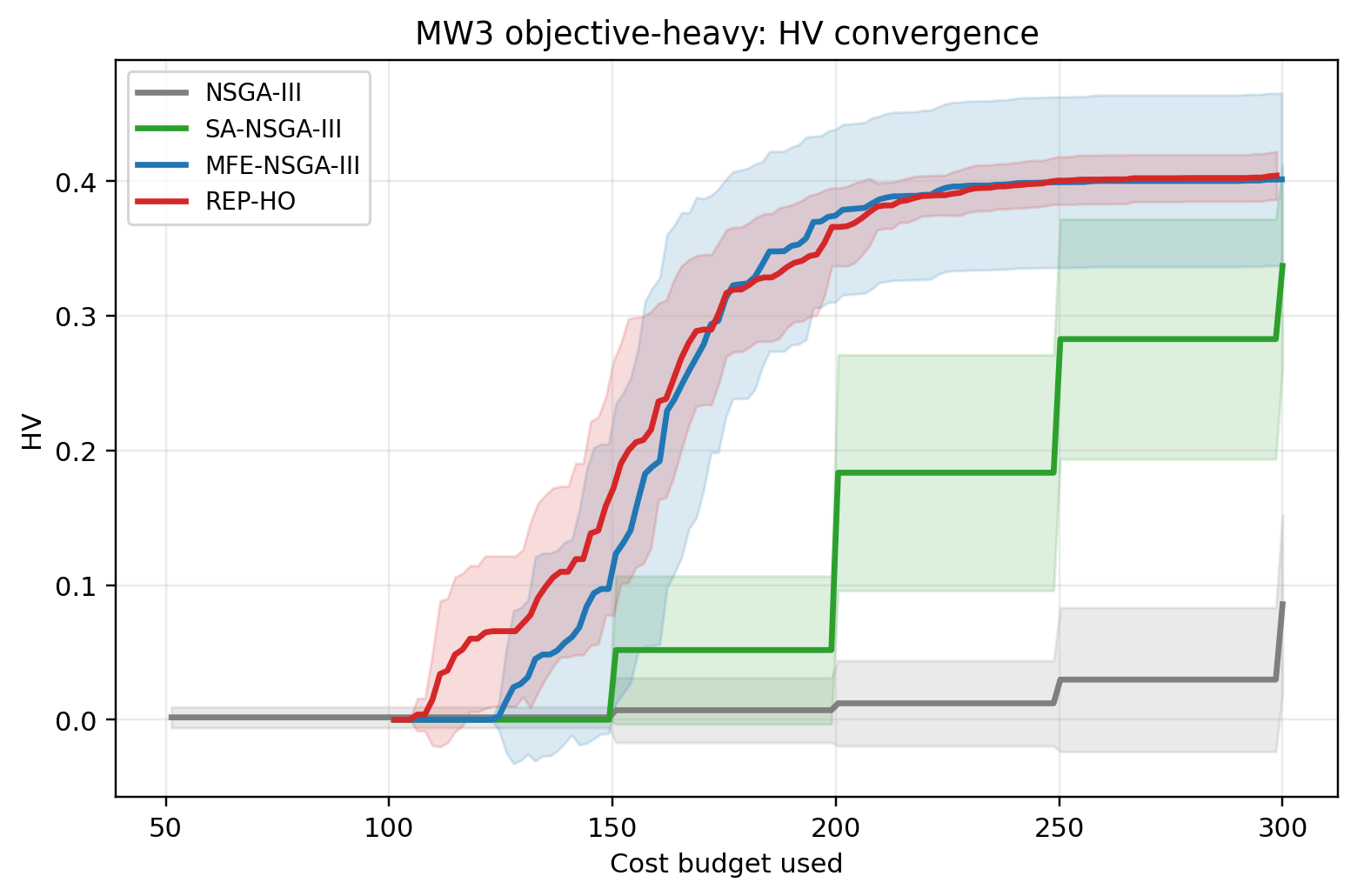}
    \end{minipage}\hfill
    \begin{minipage}{0.33\textwidth}
        \centering
        \includegraphics[width=\linewidth]{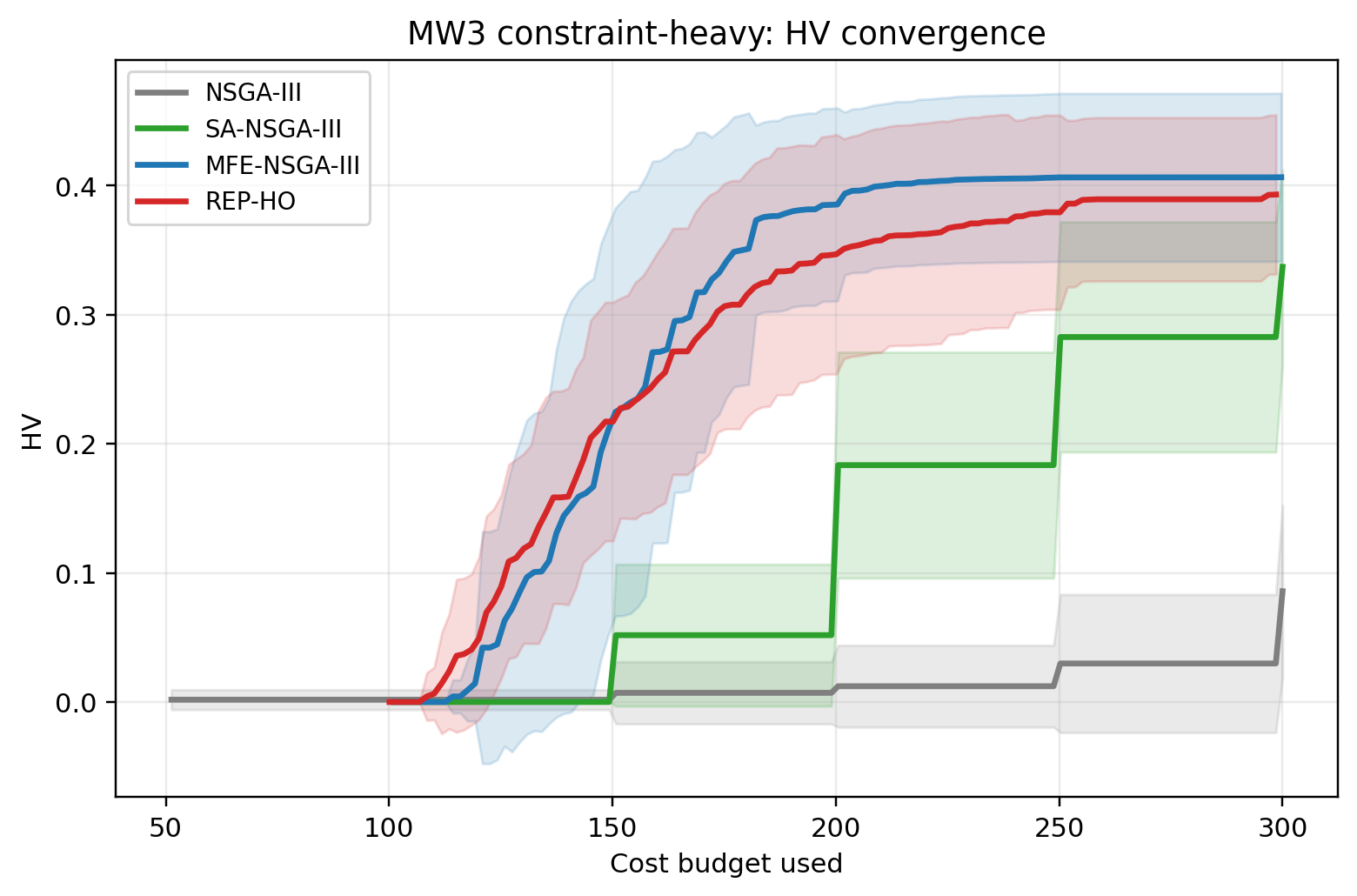}
    \end{minipage}
    \centering
    \begin{minipage}{0.33\textwidth}
        \centering
        \includegraphics[width=\linewidth]{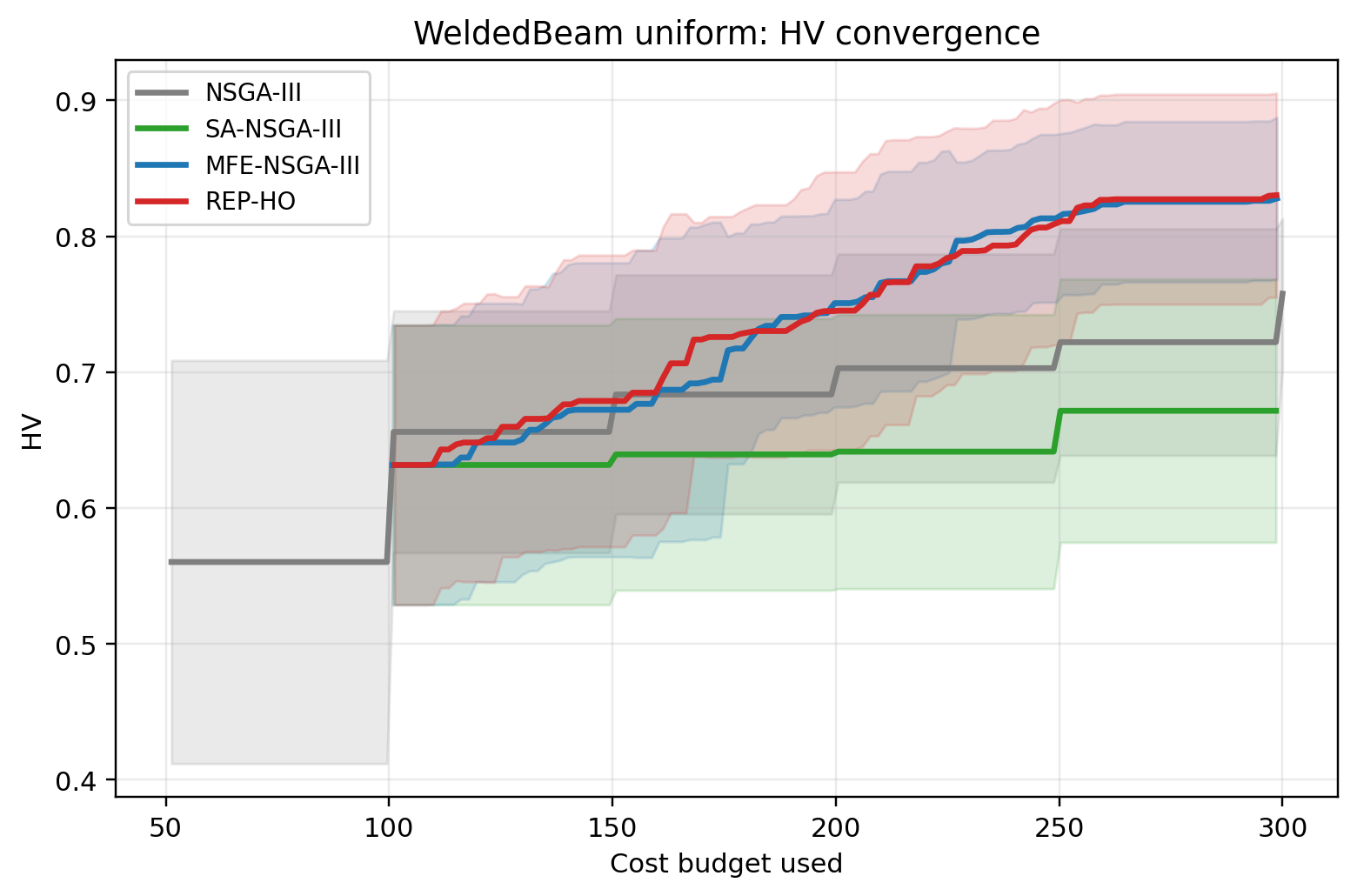}
    \end{minipage}\hfill
    \begin{minipage}{0.33\textwidth}
        \centering
        \includegraphics[width=\linewidth]{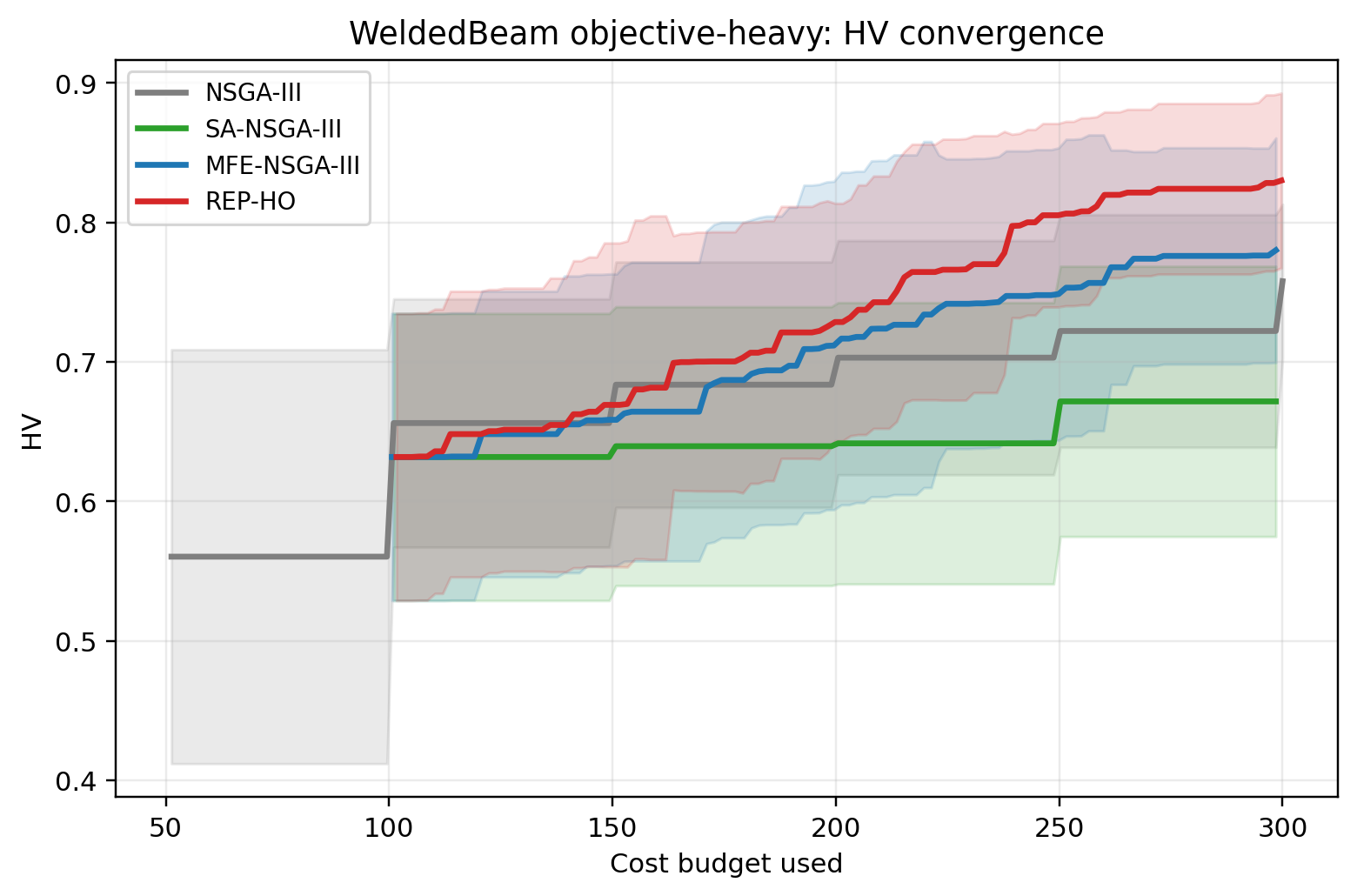}
    \end{minipage}\hfill
    \begin{minipage}{0.33\textwidth}
        \centering
        \includegraphics[width=\linewidth]{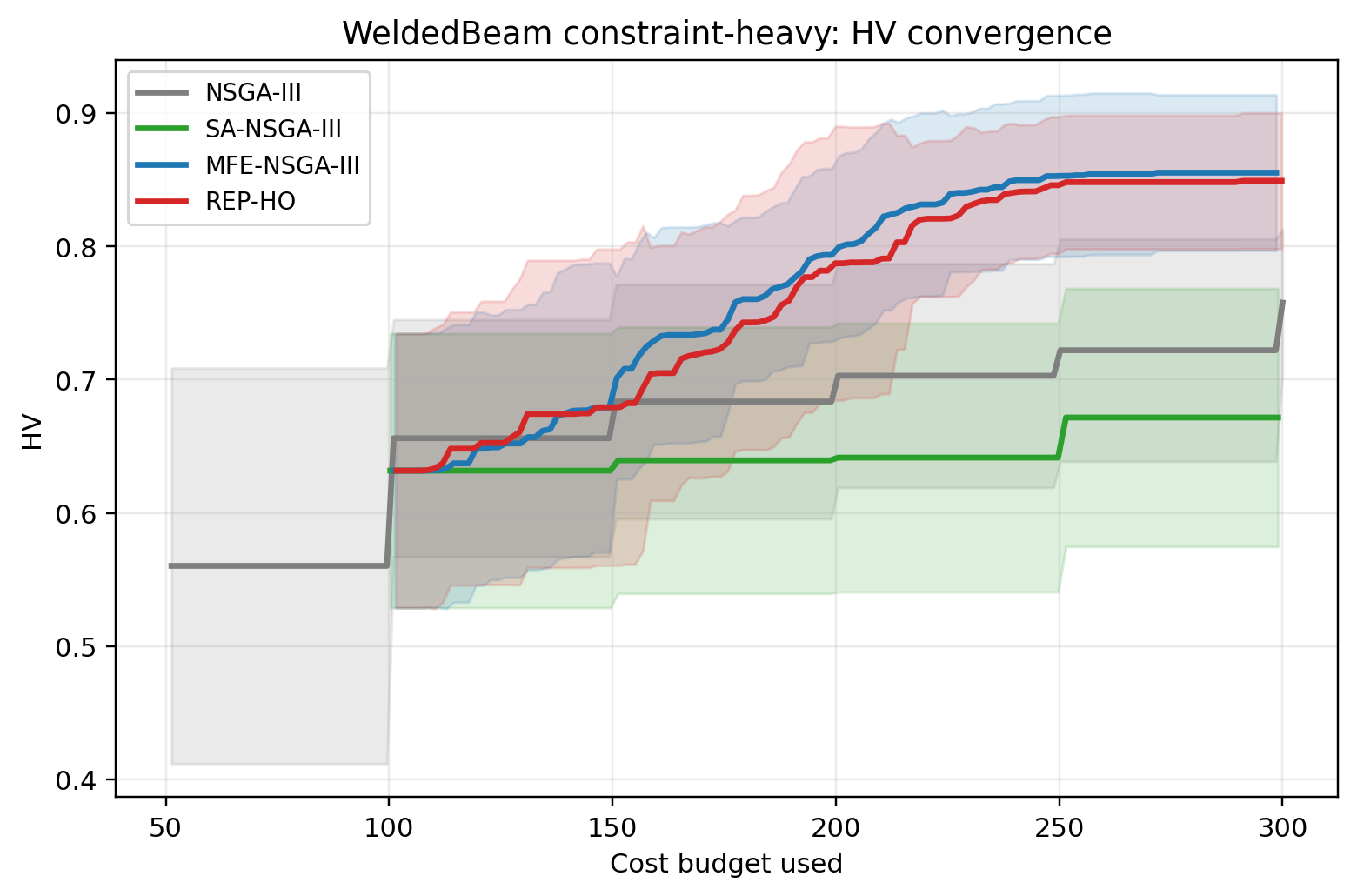}
    \end{minipage}
    \centering
    \begin{minipage}{0.33\textwidth}
        \centering
        \includegraphics[width=\linewidth]{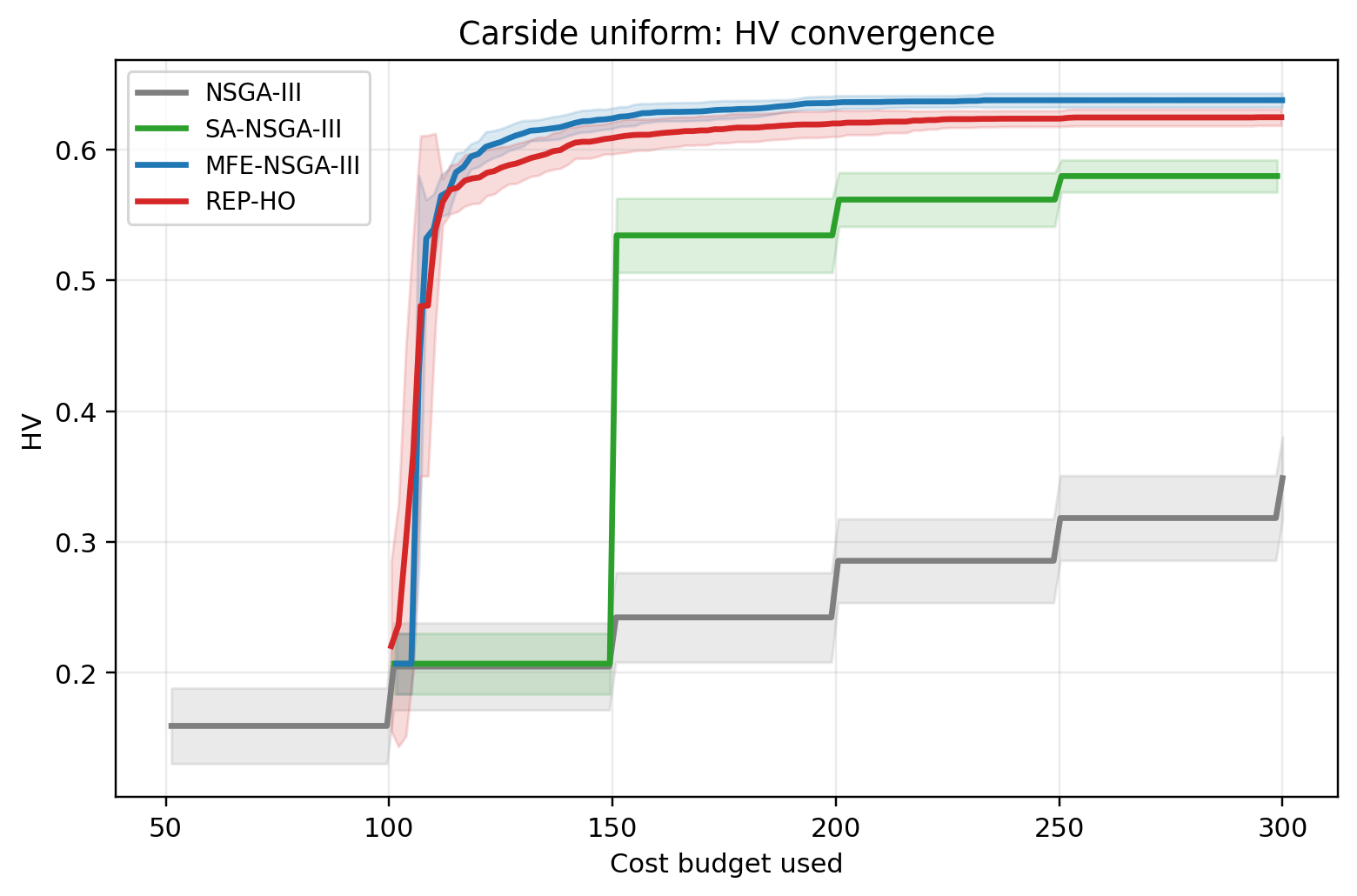}
    \end{minipage}\hfill
    \begin{minipage}{0.33\textwidth}
        \centering
        \includegraphics[width=\linewidth]{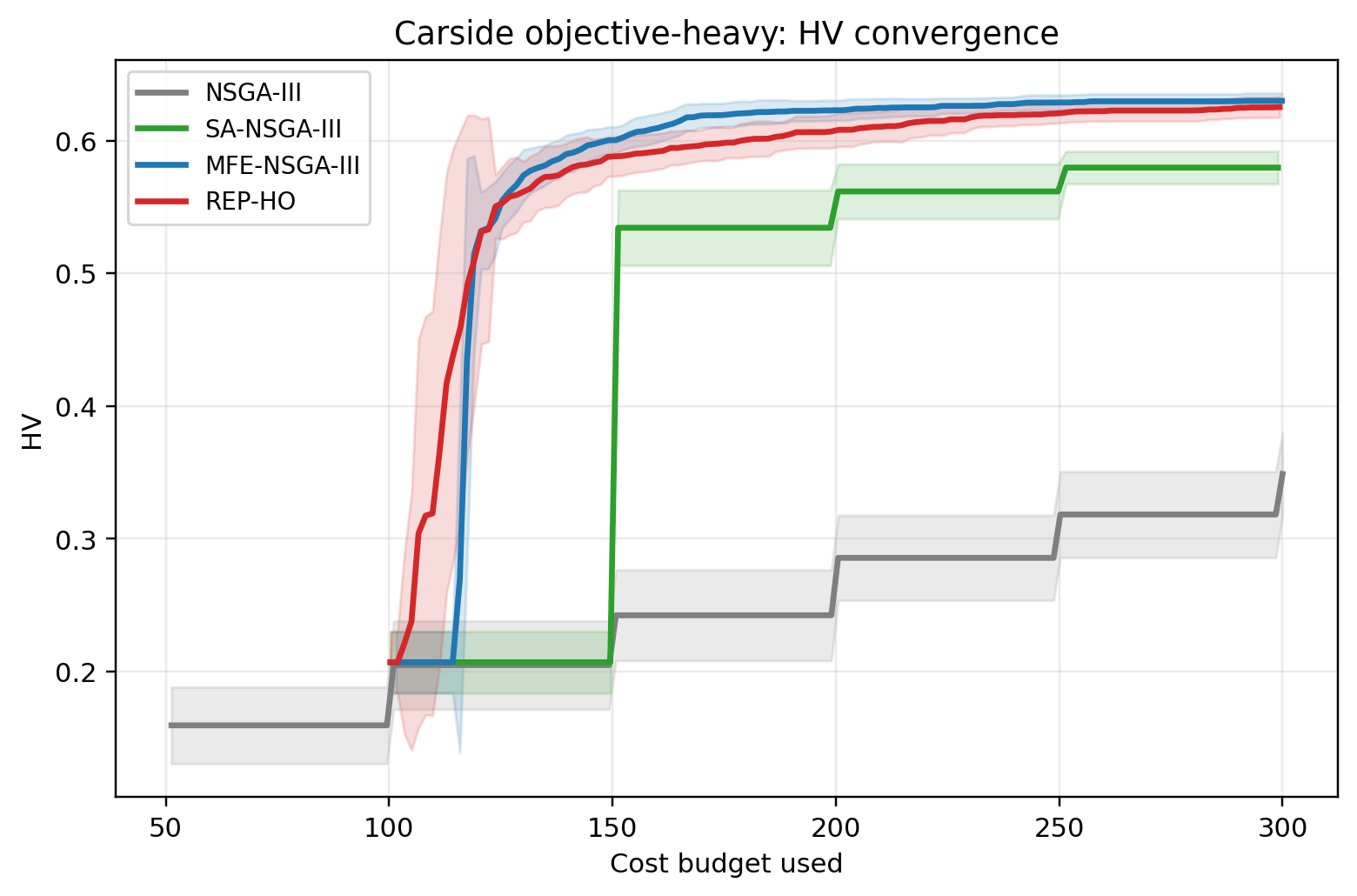}
    \end{minipage}\hfill
    \begin{minipage}{0.33\textwidth}
        \centering
        \includegraphics[width=\linewidth]{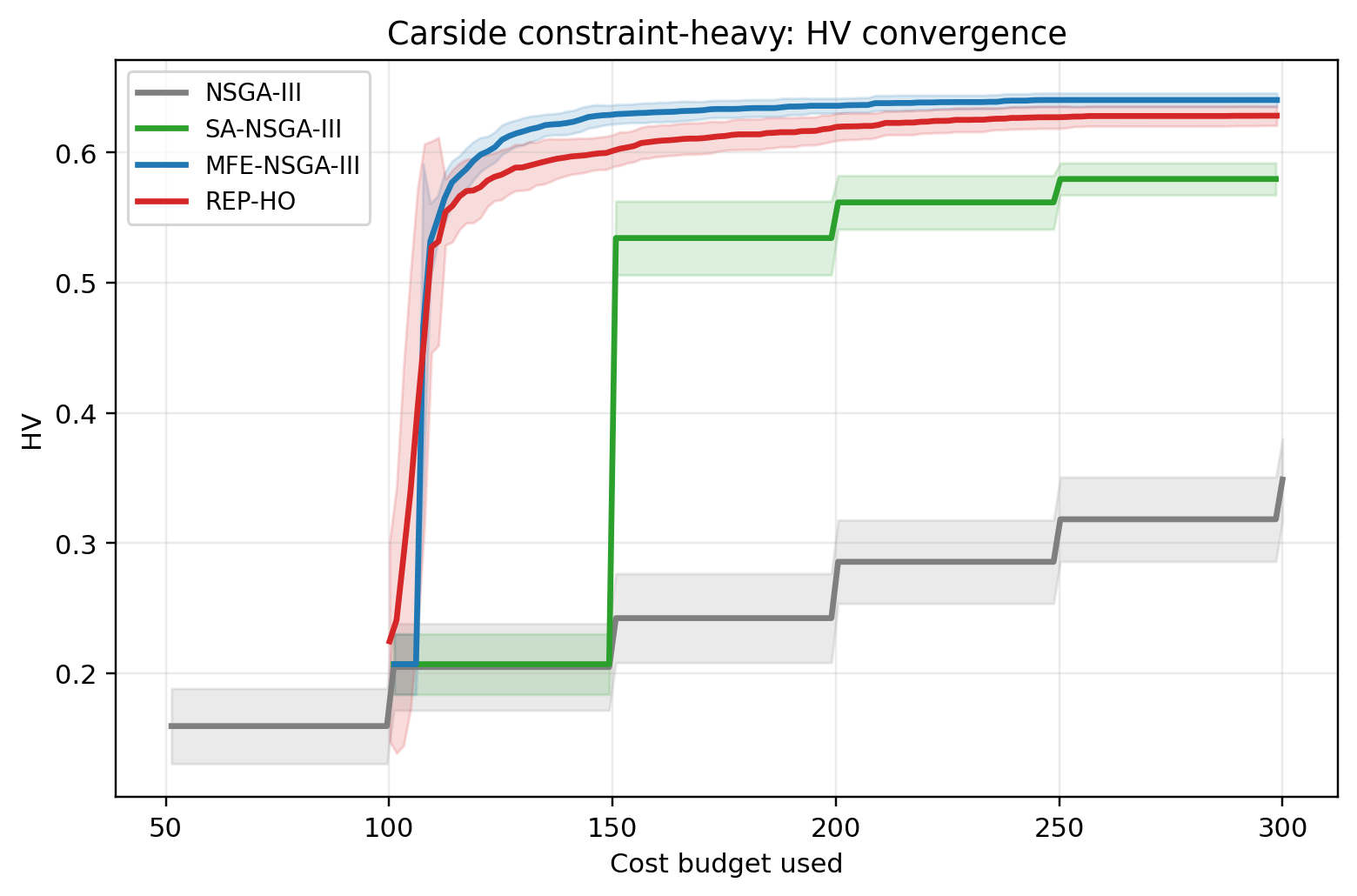}
    \end{minipage}
    \caption{HV convergence of experiments in regime 1.}
    \label{fig:regime1-convergence}
\end{figure}

\section{Prompt template example}
\label{app:prompt-example}

This appendix provides one concrete prompt--response example used by REP-HO.
The example is taken from Regime~1 on TNK with the constraint-heavy
cost setting \((0.15,0.05 \mid 0.70,0.10)\). The logged backbone is
DeepSeek-V4-Flash; Qwen3.5-35B-A3B receives the same template.
The problem name is not included.
Repeated candidates are abbreviated.
This first-generation log has an empty neighbor list for every candidate, so
\texttt{candidate\_A} below shows two illustrative neighbors rather than points from that log.

\begin{tcolorbox}[
    colback=gray!6,
    colframe=gray!35,
    title={Prompt example},
    fonttitle=\bfseries,
    sharp corners,
    boxrule=0.4pt,
    breakable,
    enhanced,
    left=4pt,
    right=4pt,
    top=4pt,
    bottom=4pt
]
\footnotesize
\begin{verbatim}
{
  "task_intro": "You are assisting a constrained bi-objective optimizer. Your goal is to choose the most informative high-fidelity evaluations under limited budget.",
  "problem_intro": "This is a two-objective constrained design problem. Lower objective values are preferred. A candidate is feasible only when all constraints are satisfied. The true formulas are hidden.",
  "task": {
    "obj_dirs": {"f1": "minimize", "f2": "minimize"},
    "constraint_rule": "For every constraint, predicted_value <= 0 means predicted feasible, and predicted_value > 0 means predicted violated.",
    "metric_def": "p_nd means the current surrogate-based probability that a candidate is non-dominated with respect to a specific objective view. hfe means high-fidelity evaluated. ocsp means overall constraint satisfaction probability."
  },
  "notes": [
    "Candidate labels are stable identifiers. Use them exactly as given.",
    "Some targets may already been evaluated with high-fidelity. Do not request them again.",
    "Only targets evaluated with high-fidelity consume budget, low-fidelity targets do not consume budget.",
    "nearest_history_points lists at most 3 later infill points in normalized design space. Initial DOE points are excluded; their landscape coverage is already reflected in the surrogate uncertainty of this candidate. Let d1 be the RMS distance to the nearest remaining historical infill. Keep a neighbor only if its RMS distance is <= 1.5 * d1. If fewer than 3 points qualify, list fewer; if there is no later infill history, use an empty list. Only high-fidelity evaluated values are shown there; value=null means that target has not been high-fidelity evaluated for that historical point.",
    "Overall feasibility is determined by satisfying all constraints at the same time. A candidate is truly feasible only if every constraint value is <= 0."
  ],
  "budget": {
    "current_cost_used": 100,
    "cost_used_since_initial_doe": 0,
    "total_budget": 250,
    "remaining_budget": 150,
    "fillup_cost_for_full_generation": 50,
    "budget_pressure_ratio": 0
  },
  "targets": {
    "objectives": [
      {"name": "f1", "evaluation_cost": 0.15},
      {"name": "f2", "evaluation_cost": 0.05}
    ],
    "constraints": [
      {"name": "g1", "evaluation_cost": 0.7},
      {"name": "g2", "evaluation_cost": 0.1}
    ]
  },
  "candidates": [
    {
      "label": "candidate_A",
      "x": [1.0404, 0.122],
      "objectives": {
        "f1": {"predicted_value": 1.0404, "uncertainty": 1.7211e-15, "p_nd": 1, "already_hfe": false},
        "f2": {"predicted_value": 0.122, "uncertainty": 1.9071e-15, "p_nd": 0, "already_hfe": false}
      },
      "constraints": {
        "g1": {"predicted_value": -0.03, "uncertainty": 0.0793, "satisfied_if_prediction_is_exact": true, "constraint_zscore": 0.3787, "already_hfe": false},
        "g2": {"predicted_value": -0.1302, "uncertainty": 9.421e-15, "satisfied_if_prediction_is_exact": true, "constraint_zscore": 1e6, "already_hfe": false}
      },
      "ocsp": 0.6475,
      "nearest_history_points": [
        {
          "distance": 0.041,
          "closer_than_fraction": 0.97,
          "x": [1.01, 0.16],
          "objectives": {
            "f1": {"high_fidelity_evaluated": true, "value": 1.01},
            "f2": {"high_fidelity_evaluated": true, "value": 0.16}
          },
          "constraints": {
            "g1": {"high_fidelity_evaluated": true, "value": -0.021},
            "g2": {"high_fidelity_evaluated": false, "value": null}
          }
        },
        {
          "distance": 0.058,
          "closer_than_fraction": 0.91,
          "x": [0.98, 0.19],
          "objectives": {
            "f1": {"high_fidelity_evaluated": true, "value": 0.98},
            "f2": {"high_fidelity_evaluated": false, "value": null}
          },
          "constraints": {
            "g1": {"high_fidelity_evaluated": true, "value": 0.014},
            "g2": {"high_fidelity_evaluated": true, "value": -0.44}
          }
        }
      ]
    },
    ...
    {
      "label": "candidate_J",
      "x": [0.7852, 0.5972],
      "objectives": { "...": "same fields as candidate_A" },
      "constraints": { "...": "same fields as candidate_A" },
      "ocsp": 0.9997,
      "nearest_history_points": []
    }
  ],
  "batch_condition": "There are 5 prompts in total for a generation, each prompt contains randomly selected 10 candidates.So try to make the best use of budget as there are 40 candidates in other prompts needed to be planned.",
  "request": {
    "scope": "Every candidate shown here has already been selected by the original optimizer. Do not decide candidate inclusion or exclusion. Only decide which objectives and constraints, if any, deserve high-fidelity evaluation for each shown candidate. Empty lists are allowed.",
    "n_candidates_in_batch": 10,
    "allowed_targets_per_candidate": "You may request zero or more objectives and zero or more constraints for each selected candidate."
  },
  "output": {
    "format": "Return JSON only. Do not use markdown.",
    "example schema": {
      "selections": [
        {"label": "candidate_A", "evaluate_objectives": ["f1", "f2"], "evaluate_constraints": []}
      ]
    },
    "rules": [
      "The objectives and constraints you put in the brackets will be evaluated with high-fidelity and consume budget, otherwise will be evaluated with low-fidelity and not consume budget,",
      "Use labels and target names exactly as provided.",
      "Do not output indices. Do not give any explanation.",
      "Return a decision for every candidate shown in this batch.",
      "Empty objective and constraint lists are allowed when no high-fidelity evaluation is needed.",
      "Return valid JSON that can be parsed directly."
    ]
  }
}
\end{verbatim}
\end{tcolorbox}

\begin{tcolorbox}[
    colback=gray!6,
    colframe=gray!35,
    title={LLM output example},
    fonttitle=\bfseries,
    sharp corners,
    boxrule=0.4pt,
    breakable,
    enhanced,
    left=4pt,
    right=4pt,
    top=4pt,
    bottom=4pt
]
\footnotesize
\begin{verbatim}
{
  "selections": [
    {"label": "candidate_A", "evaluate_objectives": ["f2"], "evaluate_constraints": ["g1"]},
    {"label": "candidate_B", "evaluate_objectives": ["f2"], "evaluate_constraints": ["g1"]},
    ...
    {"label": "candidate_J", "evaluate_objectives": ["f1", "f2"], "evaluate_constraints": ["g1"]}
  ]
}
\end{verbatim}
\end{tcolorbox}

\end{document}